\RequirePackage{fix-cm}
\documentclass[twocolumn]{svjour3}           % twocolumn
\smartqed  % flush right qed marks, e.g. at end of proof
\usepackage{graphicx}
\usepackage{balance}  % for  \balance command ON LAST PAGE  (only there!)
\usepackage{enumitem}
\usepackage{amsmath}
\usepackage{amsthm}
\usepackage{amssymb}
\usepackage{algorithm}
\usepackage{algorithmic}
\usepackage{subfigure}
\usepackage{multirow}
\usepackage{bm}
\usepackage{nicefrac}
\usepackage{subfigure}
\usepackage{caption}
\usepackage{enumitem}
\usepackage{url}
\usepackage{color}
\usepackage{cite}
\usepackage{times}

\usepackage{array}
\newcolumntype{L}[1]{>{\raggedright\let\newline\\\arraybackslash\hspace{0pt}}m{#1}}
\newcolumntype{C}[1]{>{\centering\let\newline  \\\arraybackslash\hspace{0pt}}m{#1}}
\newcolumntype{R}[1]{>{\raggedleft\let\newline \\\arraybackslash\hspace{0pt}}m{#1}}

 \usepackage{mathptmx}      % use Times fonts if available on your TeX system
\begin{document}

\title{Efficient, Simple and Automated Negative Sampling for 
	Knowledge Graph Embedding
}

\author{Yongqi~Zhang \and Quanming~Yao \and Lei~Chen}

\institute{Yongqi~Zhang \at
	4Paradigm Inc. \\
	\email{zhangyongqi@4paradigm.com}
	\and
	Quanming~Yao \at
	Department of Electronic Engineering,
	Tsinghua University and \\
	4Paradigm Inc. \\
	\email{ yaoquanming@4paradigm.com}
	\and
	Lei~Chen\at
    Hong Kong University of Science and Technology \\
    \email{leichen@cse.ust.hk} 
    \and
    Code: \url{https://github.com/AutoML-4Paradigm/NSCaching} \\
    Correspondance is to: Quanming Yao.
}

\date{Received: date / Accepted: date}

\maketitle
\begin{abstract}
Negative sampling,
which samples negative triplets from non-observed ones in knowledge graph (KG), 
is an essential step in KG embedding.
Recently,
generative adversarial network (GAN),
has been introduced in negative sampling.
By sampling negative triplets with large gradients,
these methods avoid the problem of vanishing gradient and thus obtain better performance.
However,
they make the original model more complex and harder to train.
In this paper,
motivated by the observation that negative triplets with large gradients are important but rare,
we propose to directly keep track of them with the cache.
In this way,
our method acts as a ``distilled'' version of previous GAN-based methods,
which does not waste training time on additional parameters to fit the full distribution of negative triplets.
However,
how to sample from and update the cache are two critical questions.
We propose to solve these issues by automated machine learning techniques.
The automated version also covers GAN-based methods as special cases.
Theoretical explanation of NSCaching is also provided,
justifying the superior over fixed sampling scheme.
Besides,
we further extend NSCaching with skip-gram model for graph embedding.
Finally,
extensive experiments show that 
our method can gain significant improvements on various KG embedding models and the skip-gram model, 
and outperforms the state-of-the-art negative sampling methods.

\end{abstract}

\section{Introduction}
\label{sec:introduction}

\begin{figure}[ht]
	\centering
	\includegraphics[width=0.45\textwidth]{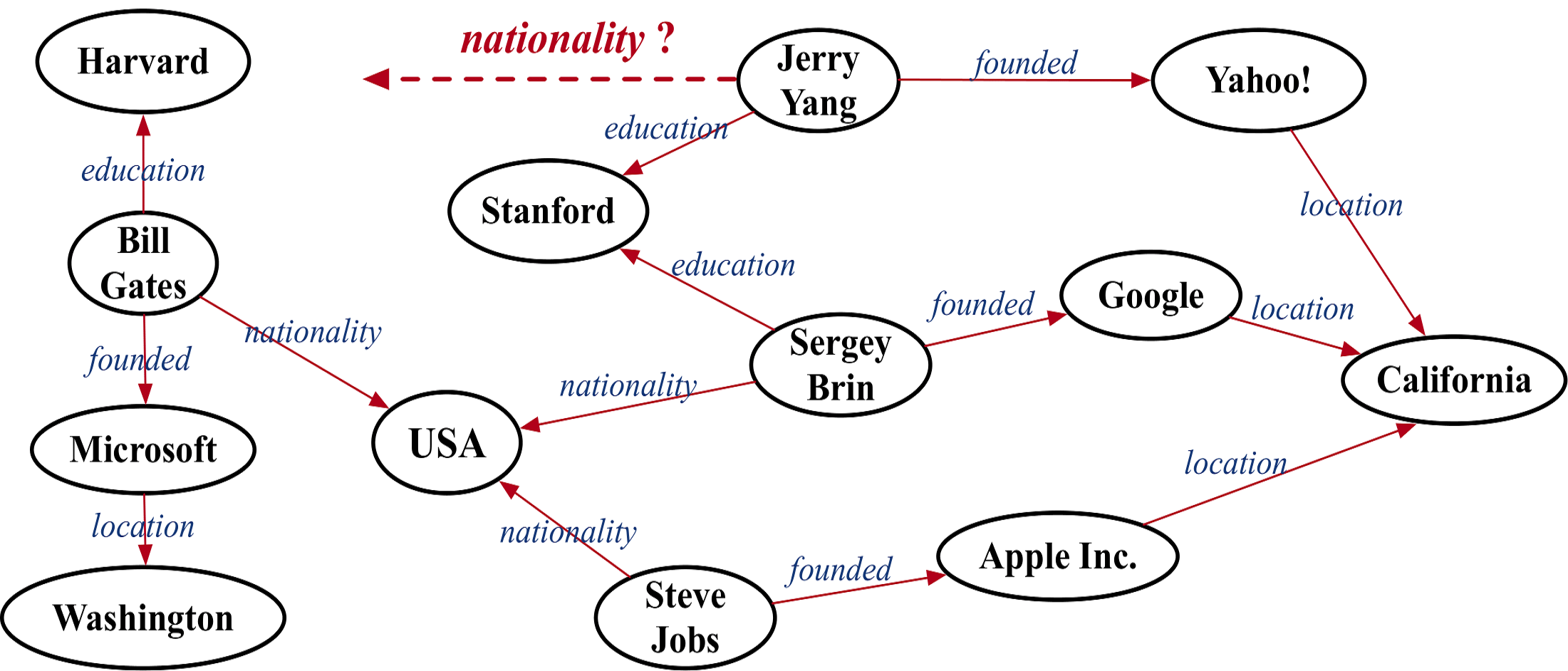}
	\caption{An example of knowledge graph.}
	\label{fig:kg}
\end{figure}

Knowledge graph (KG) is a special kind of graph structure, 
with entities as nodes
and relations as directed edges. 
Each edge (also called a fact) is represented as a triplet with the form \emph{(head entity, relation, tail entity)}, 
denoted as $(h, r, t)$, indicating that two entities are connected by a specific relation, 
e.g. \emph{(Steve Jobs, founded,  Apple Inc.)} in the example in Figure~\ref{fig:kg}.
These triplets are usually extracted manually or based on automatically constructed knowledge bases \cite{suchanek2007yago}.
KG is very general and useful.
It has been used as fundamental building blocks for many applications
like structured search \cite{dori2004visweb,mottin2016exemplar},
question answering \cite{bordes2014question},
recommendation \cite{zhang2016collaborative,noy2019industry}
and medical diagnosis \cite{zhang2018generative}.
This importance
has also inspired many famous KG projects,
such as
FreeBase \cite{bollacker2008freebase},
DBpedia \cite{auer2007dbpedia},
and YAGO \cite{suchanek2007yago}.

As these triplets are hard to manipulate,
how to find a good representation for entities and relations in the KG \cite{nickel2016review} is a fundamental problem.
Early works towards this goal lie in statistical relational learning by using the symbolic triplet data \cite{kok2007statistical, getoor2007introduction,lao2011random}. 
However, these methods neither lead to good generalization performance, 
nor can they be applied for large scale knowledge graphs. 
In comparison,
the embedding based methods have better generalization ability and inference efficiency
\cite{zou2012answering,perozzi2014deepwalk,bordes2013translating}. 
Recently, 
graph embedding techniques \cite{wang2017knowledge} have been introduced in KG learning.
These methods attempt to encode entities and relations in KG into a low-dimensional vector space
while capturing the original connection properties.
They are scalable and have also shown promising performance in basic KG tasks, 
such as link prediction and triplet classification \cite{bordes2013translating, wang2017knowledge}.

In recent years,
constructing new scoring functions that can better model the complex interactions between entities and relations 
has been the main focus for improving the performance of KG embedding approaches
\cite{wang2014knowledge,ji2015knowledge,yang2014embedding,trouillon2016complex}.
However,
another very important perspective of KG embedding,
i.e., negative sampling, is not sufficiently emphasized.
The need for negative sampling comes from the fact that there are usually only positive triplets in KG \cite{getoor2007introduction,wang2017knowledge}.
First,
to avoid trivial solutions of the embedding,
a set that contains all the possible negative samples
needs to be hand-made.
Then,
in consideration of both computation cost and memory space,
stochastic training is needed in each iteration.
Specifically,
once we have picked up a positive triplet,
we also need to sample some negative triplets 
from its corresponding negative sample set.
Besides,
the quality of these negative triplets does matter.

Due to its simplicity and efficiency,
uniform sampling is broadly used in KG embedding \cite{wang2017knowledge}.
However,
it is a fixed scheme and ignores changes in the distribution of negative triplets during the training process.
As a result,
it suffers seriously from the \textit{vanishing gradient} problem \cite{wang2018incorporating}
and biased estimation problem \cite{rawat2019sampled}.
As observed in \cite{wang2018incorporating},
most negative triplets in the sampling set can be easily classified.
Since the scoring functions tend to give observed (positive) triplets large values,
as training goes,
most of the non-observed (probably negative) triplets will have smaller values.
Thus, 
when negative triplets are uniformly sampled,
it is very likely that we pick up one with zero gradients.
As a result, 
the training process of KG embedding 
will be impeded by the vanishing gradients rather than by the optimization algorithm.
This problem prevents KG embedding from getting the desired performance.
A better sampling scheme,
the Bernoulli sampling, 
is introduced in \cite{wang2014knowledge}.
It improves uniform sampling
by considering one-to-many,
many-to-one,
and many-to-many mapping in the relations between entities.
For example, 
given a one-to-many relation,
replacing the tail entity will have a larger chance of getting a false negative triplet
compared with replacing the head.
However,
it is still a fixed and biased sampling scheme.

Therefore,
dynamically sampling from the negative triplets' distribution to help the training process
is important and non-trivial.
To efficiently capture them during training,
we have two main challenges for negative sampling:
(i). How to capture and model the negative triplets' dynamic distribution? 
and (ii). How can we effectively sample the negative triplets?
Recently,
there are two pioneering works, 
i.e.,
IGAN \cite{wang2018incorporating} and KBGAN \cite{cai2018kbgan},
attempting to address these challenges.
Their ideas are to replace the fixed sampling scheme with
a generative adversarial network (GAN) \cite{goodfellow2014generative}
based sampling scheme.
However,
GAN-based solutions still have many problems.
First, GAN increases the number of training parameters because an extra generator should be learned as sampler.
Second,
the training of 
the GAN model 
usually suffers from instability and degeneracy \cite{arjovsky2017wasserstein, gulrajani2017improved}. 
The REINFORCE gradient \cite{williams1992simple},
which is known to have high variance, 
should be used to train the generator.
Besides,
since only a few negative triplets can lead to large gradient,
IGAN and KBGAN take a lot of effort to model the distributions of all the negative ones.
These drawbacks lead to instable performance for different scoring functions, 
and hence pre-training becomes a must for both IGAN and KBGAN.
Self adversarial sampling
(Self-Adv) \cite{sun2018rotate} uses the self embedding model to replace the generator.
It solves the problem of training GAN model,
but it cannot guarantee to sample enough negative triplets with large gradient in each iteration.

In this paper, to address the challenges of capturing the dynamic and complex negative sampling distribution
while avoid the problems of using GANs, 
we propose a simple and efficient negative sampling method based on the cache, 
called NSCaching.
By empirically analyzing the gradient norm distribution of negative triplets, 
we find that the distribution is highly skewed.
In other words,
there are only a few pairs of training triplets
(i.e., a positive triplet and a negative triplet) have large gradient and the others are useless.
This observation motivates us to mainly maintain the negative triplets
that lead to large gradients
during the training, 
and dynamically update the maintained triplets. 
First,
we use the cache to store large-gradient negative triplets.
Then,
we carefully design the updating and sampling rules for the cache.
In detail,
the cache-based sampling problem is formed as a hyper-parameter optimization (HPO) problem,
and we use automated machine learning (AutoML) \cite{yao2018taking} to efficiently solve it.
In this way, 
we automatically take good care of ``exploration and exploitation'' (E\&E) \cite{march1991exploration},
which balances exploring all possible high-quality negative triplets
and sampling from a few of them in the cache. 
Contributions of our work are summarized as follows:
\begin{itemize}[leftmargin=*]
\item We propose a simple, 
efficient and automated negative sampling algorithm NSCaching, 
which is a general negative sampling scheme 
and can be easily injected into many popularly used KG embedding models. 
NSCaching has fewer parameters than both IGAN \cite{wang2018incorporating} and KBGAN \cite{cai2018kbgan}.

\item 
We	provide intuitions about how NSCaching helps the KG training under convex and non-convex cases.
in the convex case,
we show that the negative sampling scheme in NSCaching 
can lead to a smaller approximation error.
In the general non-convex case,
we show that NSCaching can benefit from \textit{self-paced learning} \cite{kumar2010self,bengio2009curriculum}
by learning easy samples first
and gradually switching to harder ones.

\item 
A critical issue in the NSCaching algorithm 
is how to balance ``exploration and exploitation''(E\&E) in updating 
and sampling from the cache.
Motivated by the success of automated machine learning \cite{yao2018taking},
we propose an automated version of NSCaching,
i.e., NSCaching (auto).
The AutoML-based method has a unified view of the 
hyper-parameters (related to E\&E) of NSCaching,
and also covers IGAN/KBGAN as special cases.
Thus, it enables us to automatically balance E\&E.

\item 
We conduct experiments on five popular data sets, 
i.e.,  WN18 and FB15K (and their variants WN18RR and FB15K237),
and YAGO3-10.
Experimental results demonstrate that the NSCaching algorithm is very efficient and is more effective than the baselines  as well.
The automated version further improves NSCaching by balancing between exploration and exploitation better.

\item We extend the negative sampling algorithm from KG embedding to graph embedding. 
Random walk based graph embedding \cite{Grover2016},
which is trained with the widely used skip-gram model \cite{mikolov2013efficient},
is chosen as the testbed.
The cache-based negative sampling is used to replace the frequency-based negative sampling in skip-gram models.
Experiments on the graph embedding show that NSCaching adapts well to the new task.
\end{itemize}

The preliminary version of this paper has been published in ICDE 2019 \cite{Zhang2018}.
Comparing with \cite{Zhang2018}, 
we have made the following important improvements:
\begin{enumerate}

\item We systematically extend the algorithm by introducing AutoML to automatically balance E\&E
by hyper-parameter optimization (Section~\ref{sec:autoML}).
The AutoML approach not only helps to improve the performance of NSCaching,
but also offers insight on how GAN-based methods work;

\item
We extend NSCaching to a new task, i.e., graph embedding based on the skip-gram model (Section~\ref{sec:graph}).
We show that NSCaching algorithm can be easily adopted into such a new application scenario 
and get good empirical performance (Section~\ref{sec:exp:graph}).

\item 
We add theoretical explanation of how NSCaching leads to a smaller approximation error when the objective is convex (Section~\ref{ssec:convex}).
The new result provides intuitions about how NSCaching helps train embeddings.

\item Based on the theoretical analysis, we reform the problem from gradient point of view in Section~3.1,
and added positive sampling into the algorithm (Section~\ref{ssub:algfra}).
To our knowledge,
the positive sampling problem has not been explored in KG embedding area.

\item 
Moreover, we have conducted more experiments with new data sets, scoring functions and tasks to show the effectiveness of our algorithm (Section~\ref{ssec:compstate}),
automated machine learning to further boost performance (Section~\ref{sec:ablation}),
ablation study to analyze the design components (Section~\ref{ssec:ablation}),
synthetic setting to illustrate the convergence properties  (Section~\ref{ssec:conv}),
and graph embedding to verify the extension to the skip-gram model (Section~\ref{sec:exp:graph}).
\end{enumerate}

\noindent
\textbf{Notations.}
The mostly used symbols and their descriptions are given in Table~\ref{tab:notation}.
Vectors are denoted by lowercase boldface,
and matrices by uppercase boldface.
${Re}(\cdot)$ takes the real part of complex numbers, 
${conj}(\mathbf t) = \mathbf t_{real} - i\mathbf t_{image}$ is the conjugate of complex vectors $\mathbf t=\mathbf t_{real} + i\mathbf t_{image}\in\mathbb C^d$. 
$\langle\mathbf a,\mathbf b,\mathbf c\rangle = \sum_{i=1}^d \mathbf a_i\cdot \mathbf b_i \cdot \mathbf c_i$ is the inner product.

\begin{table}[t]
	\centering
	\caption{Symbols and notations.}
	\label{tab:notation}
	\renewcommand{\arraystretch}{1.2}
	\begin{tabular}{c|c}
		\hline 
		Symbol & Description \\    	\hline
		$\mathcal E, \mathcal R$ & the set of entities and set of relations  \\ \hline
		$h, t\in \mathcal E, r\in\mathcal R$ & head and tail entity, relation  \\    \hline
		$\mathcal S\equiv \{(h, r, t) \}$ & the set of triplets \\ \hline
		$\bar{\mathcal S}_i =\{(\bar{h}_i, r_i, \bar{t}_i)\}$ & the set of negative triplets for $(h_i, r_i, t_i)$ \\ \hline
		$\mathbf h,  \mathbf t \in \mathbb R^{d_1}$ & embedding of head entity and tail entity \\ \hline
		$ \mathbf r \in \mathbb R^{d_2}$ & embedding of relation \\ \hline
		$\mathcal C$ & the cache	\\ \hline
		$f(h, r, t)$ & the scoring function of the triplet $(h,r,t)$ \\ \hline
		$\mathcal N_i\subset \bar{\mathcal S}_i$ & candidate subset of negative triplets \\ \hline
		$N_1, N_2$ & cache size $\left|\mathcal  C_i\right|$, candidate size $\left|\mathcal N_i\right|$ \\ \hline
		$\alpha_1, \alpha_2, \alpha_3>0$ & temperature values for softmax function \\ \hline
	\end{tabular}
\end{table}

\section{Preliminaries and Related Works}
\label{sec-prelim}

In this section,
we introduce the stochastic training algorithm
for KG embedding in Section~\ref{sec:uniframe},
current strategies for negative sampling 
in Section~\ref{sec:negasamp},
and AutoML techniques in Section~\ref{ssec:hpoautoml}.

\begin{table}[ht]
	\centering
	\caption{Definitions of some popular scoring functions.
		All model embeddings are real values,
		except that ComplEx has complex values.
		$\mathbf h_1, \mathbf r_1, \mathbf t_1$ and $\mathbf h_2, \mathbf r_2, \mathbf t_2$ 
		are indexed from two different sets of embeddings.}
	\label{tb-score-func}
	\renewcommand{\arraystretch}{1.4}
	\setlength\tabcolsep{3pt}
%	\begin{tabular}{c| C{50px} | c}
	\begin{tabular}{c| c | c}
		\hline
		 model   & scoring function                    & definition                                                                                                                                           \\ \hline
		transna- & TransE \cite{bordes2013translating} & $-\left\|\mathbf h+\mathbf r-\mathbf t \right\|_1$                                                                                                   \\ \cline{2-3}
		 tional  & TransH \cite{wang2014knowledge}     & $-\left\|\mathbf h  -  \mathbf w_r^\top \mathbf h\mathbf w_r  +  \mathbf r  \!-\!  (\mathbf t  -  \mathbf w_r^\top \mathbf t\mathbf w_r) \right\|_1$ \\ \cline{2-3}
		distance & TransD \cite{ji2015knowledge}       & $-\left\|\mathbf h  +  \mathbf w_r \mathbf w_h^\top\mathbf h +\mathbf r \!-\! (\mathbf t + \mathbf w_r \mathbf w_t^\top\mathbf t) \right\|_1$        \\ \hline
		semantic & DistMult~\cite{yang2014embedding}   & $\left\langle \mathbf h, \mathbf r, \mathbf t\right\rangle $                                                                                         \\ \cline{2-3}
		matching & ComplEx~\cite{trouillon2016complex} & $\text{Re}\left(\left\langle \mathbf h, \mathbf r, \text{conj}(\mathbf t)\right\rangle  \right) $                                                    \\ \cline{2-3}
		         & SimplE \cite{kazemi2018simple}      & $\left\langle \mathbf h_1, \mathbf r_1, \mathbf t_2\right\rangle + \left\langle \mathbf h_2, \mathbf r_2, \mathbf t_1\right\rangle$                  \\ \hline
	\end{tabular}
\end{table}

\subsection{Knowledge Graph (KG) Embedding}
\label{sec:uniframe}

To build a KG embedding model,
we first need to pick up a scoring function $f$,
which captures the similarities between two entities based on a relation \cite{wang2017knowledge}.
Different scoring functions have their own weaknesses and strengths in capturing the underneath interactions.
Some popularly used scoring functions 
are presented in 
Table~\ref{tb-score-func}.
Besides,
the observed facts in KG are supposed to have larger scores than the non-observed ones \cite{wang2017knowledge}.
With the factual information, 
the embeddings are learned by solving the optimization problem that maximizes 
the scoring function for observed triplets 
and minimizes it for non-observed triplets at the same time.
Based on the properties of scoring functions,
KG embedding models are generally 
divided into two categories.

\begin{itemize}
\item 
The \textit{translational distance model} exploits the distance-based scoring functions. 
Inspired by the word analogy results in word embeddings \cite{mikolov2013distributed}, 
the similarity is measured by the distance between two entities, 
after a translation carried out by the relation. 
TransE \cite{bordes2013translating},
as a representative translational model,
is defined by the (negative) distance between $\mathbf h+\mathbf r$
and $\mathbf t$, 
i.e.,
$f(h, r, t) = -\left|\left|\mathbf h+\mathbf r-\mathbf t \right|\right|_1$.
Other translational distance models like TransH \cite{wang2014knowledge}, TransD \cite{ji2015knowledge}
enhance over TransE by introducing extra mapping matrices.
The translational distance models are generally optimized by minimizing the ranking based loss function
\begin{equation}
\sum_{ (h_i, r_i, t_i) \in \mathcal{S} } 
\sum_{(\bar{h}_i, r_i, \bar{t}_i) \in \bar{\mathcal{S}}_{i}} \!\!\!\!
\left[\gamma - f(h_i, r_i, t_i) + f(\bar{h}_i, r_i, \bar{t}_i)\right]_+,
\label{eq:distance}
\end{equation}
where $\gamma>0$ is the margin value for the loss function.

\item 
Scoring functions in \textit{semantic matching models} exploit the 
similarity 
of a triplet by matching latent semantics of entities and relations embedied in their vector space representations.
Bilinear models are the state-of-the-art among the semantic matching models and
they share the form as 
$f({h}, {r}, {t})=\mathbf{h}^\top\mathbf{R}\mathbf{t}$,
where $\mathbf{R} \in \mathbb R^{d\times d}$ is a matrix referring to the embedding of relation $r$ \cite{wang2017knowledge}. 
DistMult \cite{yang2014embedding} measures the similarity by directly computing the 
element-wise
product of 
the embedding vectors,
i.e.,
$f({h}, {r}, {t})
= \left\langle \mathbf{h}, \mathbf{r}, \mathbf{t} \right\rangle$,
which restricts $\mathbf{R}$ to be a diagonal matrix.
However,
it can not model asymmetric triplets since $f(h, r, t)=f(t,r,h)$ is always satisfied.
ComplEx \cite{trouillon2016complex} and SimplE \cite{kazemi2018simple} improve over DistMult
by dealing with asymmetric triplets in different ways.
Another type of models conducts semantic matching using neural networks.
Multi-Layer Perceptron (MLP) is used in \cite{dong2014knowledge} to measure the similarities.
ConvE \cite{dettmers2018convolutional} takes advantage of convolutional neural network to increase the interactions among different dimensions.
Even though neural network models are more complex than the bilinear models,
they empirically perform worse than the bilinear models \cite{trouillon2016complex,kazemi2018simple}.
The semantic matching models are mainly optimized by minimizing the classification based loss function 
\begin{equation}
\!\!\!\!\!\!
\sum_{ (h_i, r_i, t_i) \in \mathcal{S}} 
\sum_{(\bar{h}_i, r_i, \bar{t}_i) \in \bar{\mathcal{S}}_{i}}
\!\!\!\!\!\!\!\!\!
\ell\left(1, \! f(h_i,r_i,t_i) \right) 
\!+\! \ell\left( -1, \! f(\bar{h}_i,r_i,\bar{t}_i) \right)
\!,\!\!\!
\label{eq:sematic}
\end{equation}
where
$(\bar{h},r,\bar{t}) \not\in \mathcal{S}$ is the hand-made negative triplet for $(h,r,t)$ and
$\ell(\alpha, \beta) = \log\left( 1 + \exp( - \alpha \beta ) \right)$.
\end{itemize}

\begin{algorithm}[ht]
	\caption{Stochastic gradient descent for knowledge graph embedding \cite{bordes2013translating,wang2017knowledge}.}
	\label{alg:basic}
	\begin{algorithmic}[1]
		\REQUIRE training set $\mathcal{S} = \{(h, r, t)\}$, embedding dimension $d$
		and scoring function $f$;
		
		\STATE initialize the embeddings for each $e \in \mathcal E$ and $r \in \mathcal R$.
		\FOR{$i = 1, \cdots, T$}
		
		\STATE sample an observed triplet $(h_i, r_i, t_i) \in \mathcal{S}$;
		
		\STATE sample the corresponding negative triplet $(\bar{h}_i, r_i, \bar{t}_i) \in \bar{\mathcal{S}}_i$;
		// \textit{negative sampling}
		
		\STATE update parameters of embeddings w.r.t. the gradients using 
		\\ (i). translational distance models:
		\begin{align}
		\nabla \left[\gamma -f\left( h_i, r_i, t_i \right) + f\left( \bar{h}_i, r_i, \bar{t}_i \right) \right]_+,
		\label{eq:graddist}
		\end{align}
		or (ii). semantic matching models:
		\begin{align}
		\nabla 
		\left[ \ell\left( +1, f(h_i, r_i, t_i) \right) + \ell\left( -1, f(\bar{h}_i, r_i, \bar{t}_i) \right) \right] ;
		\label{eq:gradsema}
		\end{align}
		
		\ENDFOR
	\end{algorithmic}
\end{algorithm}

The above two objectives,
i.e., \eqref{eq:distance} and \eqref{eq:sematic},
can be optimized by
using stochastic gradient descent
in an unified manner (Algorithm~\ref{alg:basic}).
In each iteration,
an observed (positive) triplet $(h_i, r_i, t_i)$ is firstly sampled from the training set $\mathcal{S}$ at step~3.  
Since there are no negative triplets in $\mathcal{S}$,
in step~4,
a negative triplet of $(h_i, r_i, t_i)$ is sampled from the corresponding negative triplets
set $\bar{\mathcal{S}}_i$ \cite{bordes2013translating}, i.e.,
\begin{align}
\bar{\mathcal{S}}_i = 
\left\lbrace  
(\bar{h}, r_i, t_i) \! \notin \! \mathcal{S}
\,|\, 
\bar{h}\in\mathcal E \right\rbrace  
\! \cup \!
\left\lbrace (h_i, r_i, \bar{t}) \! \notin \! \mathcal{S}
\,|\,
\bar{t} \in \mathcal E
\right\rbrace.
\label{eq:nega}
\end{align}
Finally,
embedding parameters are updated in step~5.
Since the quality of negative triplets in $\bar{\mathcal{S}}_i$ is diverse,
how to sample a proper $(\bar{h}_i, r_i, \bar{t}_i)$ has been developed as an important perspective
affecting the performance of knowledge graph embedding
\cite{wang2017knowledge,wang2018incorporating,cai2018kbgan,sun2018rotate}.

\subsection{Negative Sampling in KG Embedding}
\label{sec:negasamp}

Negative sampling is important for improving learning models
when there are only positive samples. 
The typical applications include
natural language processing \cite{mikolov2013linguistic,li2019sampling,chen2018improving},
computer vision \cite{wu2017sampling},
graph embedding \cite{perozzi2014deepwalk,tang2015line,Grover2016}, 
recommender system \cite{rendle2009bpr,ding2019reinforced,zhang2013optimizing,ying2018graph},
and KG embedding \cite{bordes2013translating,wang2017knowledge} here.
Existing works on negative sampling can be divided into two categories,
i.e., 
sampling from fixed distribution 
and sampling from dynamic distribution.

A uniform distribution over the negative samples in the candidate set is a simple yet efficient choice \cite{bordes2013translating,rendle2009bpr}. 
Whereas, many of the uniformly generated negative samples are not informative and too trivial to recognize \cite{mikolov2013linguistic,li2019sampling,wu2017sampling,ding2019reinforced}.
In order to generate more informative negative samples,
important statistics in the data set can be used to define the distribution,
such as the frequency of words \cite{mikolov2013distributed} 
and personalized PageRank score of items \cite{zhang2019deep}.
However,
the uniform sampling methods are biased-estimator of the full negative sampling distribution \cite{rawat2019sampled}.

As the training goes on,
the distributions of scores and gradients of negative samples keep changing.
In order to capture the dynamic distribution of negative samples,
several works are proposed to sample according to the scores \cite{wu2017sampling,li2019sampling,gao2018self,chen2018improving,bose2018adversarial,ding2019reinforced,wang2018incorporating}.
There are two approaches in general.
In one direction, the high-quality negative samples are selected 
based on the scores in a small pool sampled from all the candidates \cite{li2019sampling,sun2018rotate}.
This approach is efficient since only a small number of the scores need to be computed.
In another,
a distribution on all the candidates is modeled to generate the high-quality negative samples \cite{wang2018incorporating,gao2018self}.
Having an overall distribution of the negative samples makes these method more flexible.

In the following content, we discuss the sampling methods specifically used in KG embedding tasks.

\subsubsection{Sample from Fixed Distributions}
\label{ssec:fixed}
In the early work \cite{bordes2013translating}, 
negative triplets are \emph{uniformly} sampled from the set $\bar{\mathcal{S}}_i$.
This strategy is simple yet very efficient.
Later,
a better sampling scheme,
i.e.,
Bernoulli sampling, 
is introduced in \cite{wang2014knowledge}.
It improves uniform sampling
by reducing the appearance of false negative triplets
existing in one-to-many,
many-to-many,
and many-to-one relations between head and tail entities.
However,
as mentioned in the introduction,
the Bernoulli method still  samples from fixed distributions,
which can neither model the dynamic changes in distributions of negative triplets
nor can it sample triplets with large gradient.

As introduced in Section~\ref{sec:introduction},
the vanishing gradient (or  zero loss) problem~\cite{wang2018incorporating}
means a certain number of negative triplets will lead to zero gradient
and thus is not informative for training with gradient-based optimization algorithms.
Taking the ranking based loss \eqref{eq:graddist} for example,
the score of negative triplets $f(\bar{h}_i, r_i, \bar{t}_i)$ are gradually minimized as training goes on.
For most of the negative triplets,
the score will be very small and the loss will decrease to zero soon,
leading to zero gradients.
The gradient on classification loss \eqref{eq:gradsema} will go close to zero if $f(\bar{h}_i, r_i, \bar{t}_i)$ is small.
As a result,
those negative triplets cannot provide enough gradient value to update the embeddings.
Besides, the fixed sampling methods cannot capture the dynamic distribution of negative triplets,
leading to a biased estimator.

\subsubsection{Sample from Generative Adversarial Network (GAN)}
GAN \cite{goodfellow2014generative} is originally introduced as a powerful model for image generation. 
It contains two modules: 
a \emph{generator} that serves as the sampler, 
and a \emph{discriminator} that measures the quality of generated samples.
Under elaborate control on the training procedure of generator and discriminator, 
GAN has achieved significant success in many fields,
e.g., 
computer vision \cite{arjovsky2017wasserstein,gulrajani2017improved},
natural language processing \cite{fedus2018maskgan},
information retrieval \cite{wang2017irgan}
and graph mining \cite{wang2018graphgan}.
It has also been shown to generate negative samples with high-quality for knowledge graph embedding \cite{cai2018kbgan,wang2018incorporating,sun2018rotate}.

When GAN is applied to negative sampling, the jointly trained generator 
can dynamically adapt to the new distributions by confusing the discriminator and keeping training.
The discriminator, 
i.e.,
the KG embedding model,
learns to distinguish between the positive triplets and the negative triplets
sampled by the generator.
Under an alternating training process,
the generator dynamically approximates the negative sample distribution
and the KG embedding model is improved by the negative triplets 
with relatively large gradient  sampled by the generator.

Given a positive triplet $(h, r, t)$, 
IGAN \cite{wang2018incorporating} models the distribution $\bar{h}, \bar{t}\sim p(e|(h, r, t))$ over all the entities 
to sample a negative triplet $(\bar{h}, r, \bar{t})$.
The gradient of $(\bar{h}, r, \bar{t})$ is approximately measured by the discriminator,
i.e.,
the loss function
$ \ell = \left[\gamma - f(h,r,t) + f(\bar{h}, r, \bar{t})\right]_+ $
of the target KG embedding model.
By joint training, IGAN can dynamically capture the distribution of all negative triplets.
Instead of modeling a distribution over the whole entity set,
KBGAN \cite{cai2018kbgan} learns to sample from a subset of random entities.
A set of entities $\mathcal{N}= \{(\bar{h}, r, \bar{t}) \}$ is uniformly sampled first and then
the negative triplet is picked up from $\mathcal{N}$.
KBGAN is more efficient than IGAN,
but less effective 
since it is hard to guarantee the candidate set $\mathcal{N}$ to contain enough large-gradient negative triplets.

Even though GAN provides a solution to model the dynamic negative sample distribution,
it is famous for suffering from instability and degeneracy \cite{arjovsky2017wasserstein,gulrajani2017improved}.
Besides, REINFORCE gradient \cite{williams1992simple}, which is known to have high variance,
has to be used to optimize the generator.
Therefore,
pretraining is a must for both IGAN and KBGAN.
It increases the number of model's parameters and brings extra costs on training.
A concurrent method
Self-Adv \cite{sun2018rotate} adopts a similar approach as KBGAN.
The differences are that
(i) Self-Adv uses the self model embedding to measure the quality rather than training an extra generator;
(ii) Self-Adv treats the probability as weights rather than the sampling procedure.
However, it still cannot 
guarantee the candidate set $\mathcal{N}$ to contain large-gradient negative triplets.

\begin{figure*}[ht]	
\centering
\subfigure[Grad norm v.s epoch.]
{\includegraphics[height = 4.6cm]{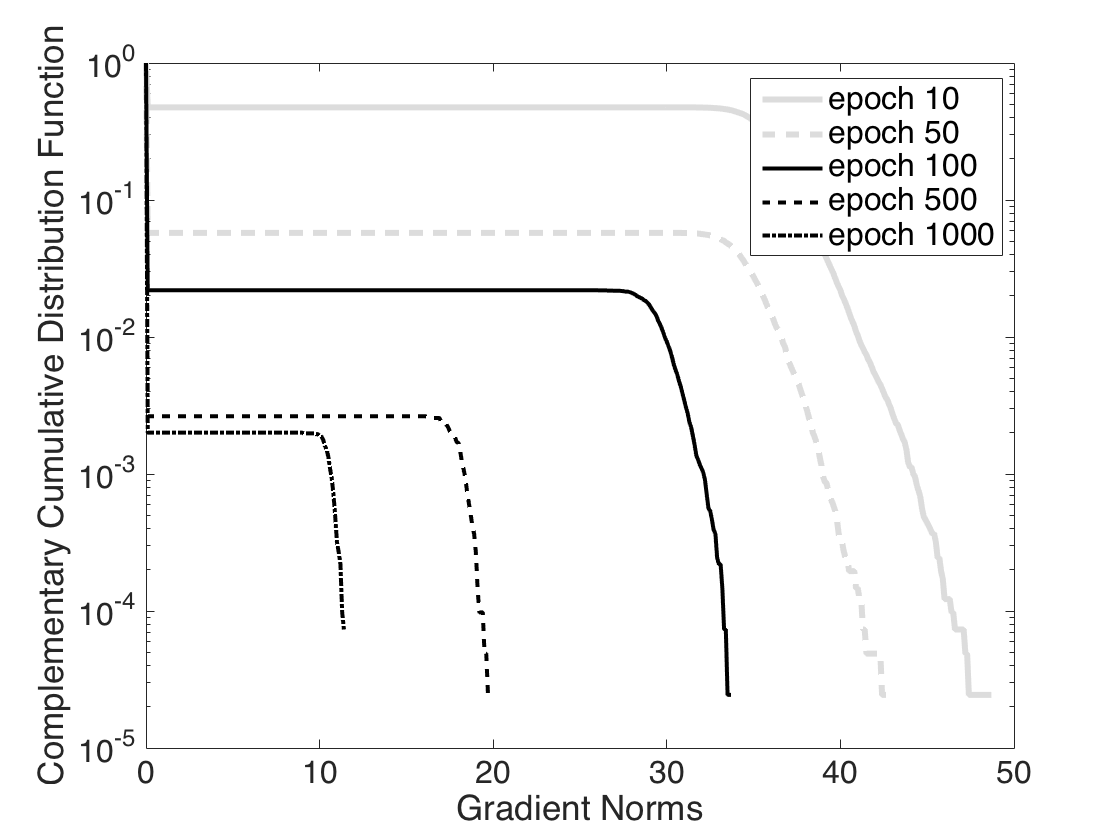}}
\qquad \qquad
\subfigure[Grad norm v.s pos-triplet.]
{\includegraphics[height = 4.6cm]{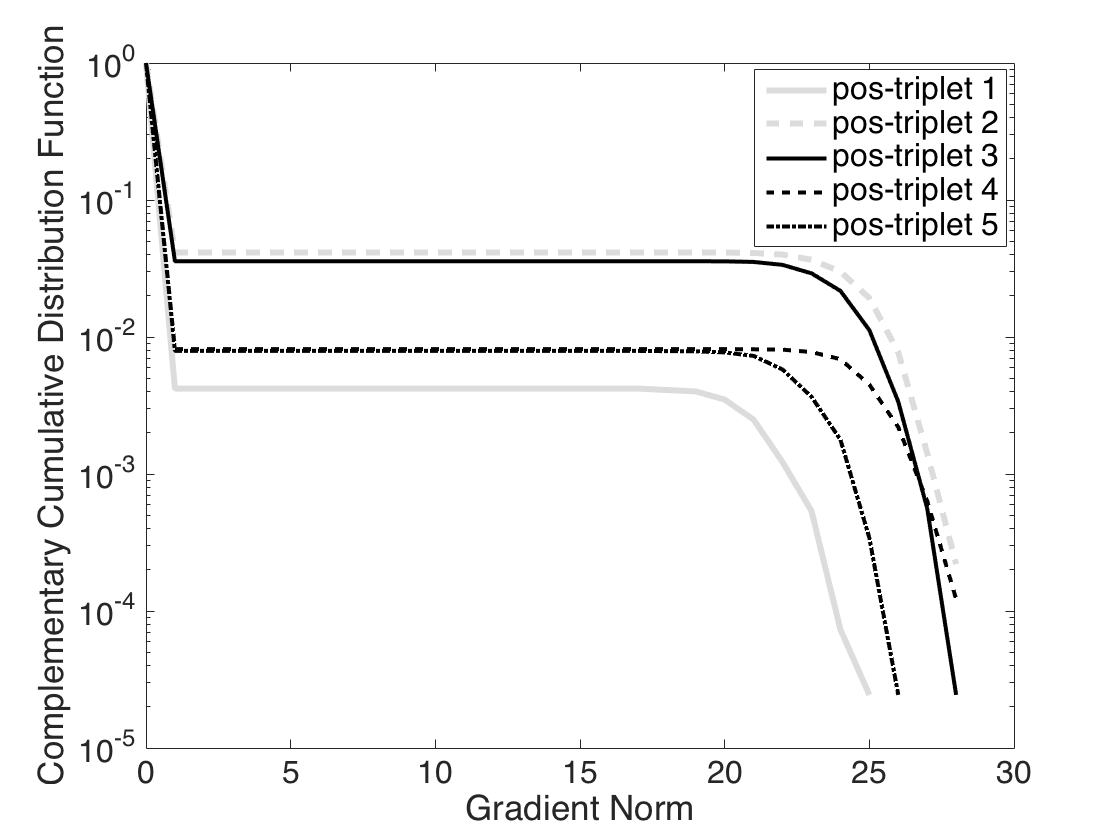}}
\caption{Distribution of training pairs' gradients on WN18 trained by Bernoulli-TransE
	(see Section~\ref{sssec:methods}). 
	For a given triplet $(h_i, r_i, t_i)$, we fix the head entity $h$ and relation $r$, and compute the $\ell_2$-norm of gradient $\|\mathbf g\|_2$ in \eqref{eq:graddist} for all $\bar{t}\in\mathcal E$.
	We measure the \emph{complementary cumulative distribution function} (CCDF) $F_{\|\mathbf g\|_2}(x) = P(\|\mathbf g\|_2\geq x)$ to show the proportion of negative triplets that satisfy  $\|\mathbf g\|_2\geq x$. 
	(a) is the distribution of training pairs in 5 timestamp of a certain triplet $(h_i, r_i, t_i)$. 
	(b) is the distribution of 5 different triplets $(h_i, r_i, t_i)$ after the pretraining stage.}
\label{fig-distance}
\end{figure*}

\subsection{Automated Machine Learning (AutoML)}
\label{ssec:hpoautoml}

Automated machine learning (AutoML) 
\cite{yao2018taking,automl_book}
has recently shown its power in easing the usage of and in designing better machine learning models.
It can be regarded as a black-box optimization problem 
where we target at efficiently searching for better hyper-parameters or model structures.
Regarding the success of AutoML,
there are two important perspectives
\begin{itemize}[leftmargin=*]
	\item Search space: This helps us to figure out important properties of the underlying learning model.
	The search space cannot be too general,
	otherwise the searching cost in such a space will be too expensive.
	
	\item Search algorithm: Since the computation cost of evaluating the settings in search space is high,
	efficient algorithms should be designed to search efficiently.
	Taking hyper-parameter optimization (HPO) as an example,
	gird search or random search \cite{bergstra2012random} are the mostly used method due to their simplicity.
	However, the searching is usually inefficient and 
	Bayesian optimization \cite{bergstra2011algorithms,hutter2011sequential} is a well-known method to improve the efficiency in HPO.
\end{itemize}

Considering that the distribution of negative triplets is highly skewed,
as will be discussed in Section~\ref{sec:proposed},
we should take the sampling distribution seriously.
As will be shown in Section~\ref{sec:autoML},
the proposed NSCaching method naturally allows 
a search space to automatically balance the
exploration and exploitation (E\&E) problem in negative sampling,
which can further improve the quality of embeddings.

\section{Proposed Model}
\label{sec:proposed}

In this section,
we first describe our key observations regarding the negative sampling in Section~\ref{sec:closer},
which are ignored by existing works but are the main motivations of our work.
The proposed method is described in Section~\ref{sec:NSCaching},
where we show how to address the challenges in negative sampling by using cache.
Then,
we analyze the proposed method from theoretical perspectives in Section~\ref{sec:theory}. 
In Section~\ref{sec:autoML},
we balance exploration and exploitation through AutoML techniques
for the proposed method.
Finally,
we discuss the problem of false negative triplets
in Section~\ref{ssec:falseneg}.

\subsection{Revisiting Distribution of Training Pairs}
\label{sec:closer}

Before introducing the proposed method,
we analyze the distribution of gradients for training pairs here.
This motivates us to use cache to efficiently approximate an unbiased distribution of the training pairs.
%since the main problem of negative sampling is about 
%\footnote{$\surd$ see footnotes in Section~2.2.1,
%	the problem is not sufficiently analyzed.}
%vanishing gradient \cite{wang2018incorporating},
%which results from frequently selecting low-quality negative samples.
%\footnote{+++ the concept of training pair is used here.}
Recall that 
the gradient at stochastic training of KG embedding is
determined by a pair of triplets,
i.e., a positive one $(h_i, r_i, t_i)$ from the training set 
and a negative one $(\bar{h}_i, r_i, \bar{t}_i)$ from negative set $\bar{\mathcal{S}}_i$ (step~3-4 in Algorithm~\ref{alg:basic}).
We show the distribution with the $\ell_2$-norm of all the training pairs' gradients for a positive $(h_i, r_i, t_i)$.
\begin{itemize}
\item Figure~\ref{fig-distance}(a) shows changes of the training pair distribution for a fixed positive triplet $(h_i, r_i, t_i)$ in different epochs; and 

\item Figure~\ref{fig-distance}(b)
shows the training pair distributions for five different positive triplets $(h_i, r_i, t_i)\in\mathcal S$.
\end{itemize}
First,
we can see that the distribution of training pairs' gradient are dynamic and highly skewed.
Second,
the training pairs with large gradients become rare along the iterating (epoch gets more),
which is consistent with the observations in \cite{wang2018incorporating,cai2018kbgan}.
Besides, the distributions of training pairs'  gradient for different positive triplets are various.

Even though GAN has strong ability in monitoring the full distribution of negative triplets,
the GAN-based methods still have a lot of limitations.
First,
they waste a lot of parameters and computational costs on learning
how negative triplets with small gradient norms are distributed.
Second,
reinforcement learning,
which provides gradient to the generator but increases the training difficulties,
should be applied in the GAN-based algorithms \cite{cai2018kbgan,wang2018incorporating}.

Besides,
the GAN-based methods ignore the distribution of positive triplets.
Since each training pair is composed of a positive triplet and a negative triplet,
the differences among positive triplets should also be considered.
Some positive triplets
can have more large gradient negative samples 
(see 
positive triplet~2 v.s. positive triplet~1 in Figure~\ref{fig-distance}(b)).
Thus,
we want to more frequently pick up 
the positive triplets with more large gradient negative samples over the others during the stochastic training,
i.e., in step~3 of Algorithm~\ref{alg:basic}. 
However,
all existing works
including  \cite{wang2018incorporating,cai2018kbgan,Zhang2018,sun2018rotate}
use uniform sampling over positive triplets.
For methods sampling from fixed distributions \cite{bordes2013translating,wang2014knowledge},
they cannot model the difference of both positive and negative triplets.
GAN-based ones are already too complex \cite{wang2018incorporating,cai2018kbgan}
to model the positive sampling.
Capturing the difference of positive triplets
will further increase the model's parameters and make the training even harder.

\subsection{NSCaching: the Proposed Method}
\label{sec:NSCaching}

From observations in Section~\ref{sec:closer},
we have three questions on how to sample the training pairs
(i.e., a positive and a negative triplet).
\begin{itemize}
\item[1).] 
Is it possible to directly keep track of negative triplets
which can give large gradient  for a given positive triplet,
rather than the whole negative triplets' distribution?
\item[2).] 
How can we adaptively sample positive triplets having more large-gradient training pairs?
\end{itemize}
Besides,
considering the distribution is dynamic and hard to estimate,
\begin{itemize}
\item[3).] 
how to balance exploring all the training pairs leading to large gradient 
and exploiting those that have the largest gradient norms?
\end{itemize}

In this section,
we describe the proposed method
to address these three questions.

\subsubsection{Core Idea: Caching training pairs}

As  in Section~\ref{sec:closer},
the total number of large-gradient negative triplets associated with a positive one is small.
Therefore, 
we are motivated to use a small amount of extra memory,
which caches negative samples with large gradient norms for each triplet $(h_i, r_i, t_i) \in \mathcal{S}$.
The designed cache acts as a truncated representation
of triplets' distribution 
$(\bar{h}_i, r_i, \bar{t}_i) \in \bar{\mathcal{S}}_{i}$.
Such an idea is previously explored in Word2Vec \cite{mikolov2013distributed},
where the estimated distribution of negative samples is also truncated.
This improves both the efficiency and quality of negative sampling.

Note that,
as in \eqref{eq:nega}, the negative triplet $(\bar{h}_i, r_i, \bar{t}_i) \not\in \mathcal{S}$ is formed 
by either $(\bar{h}_i, r_i, t_i)$ or $(h_i, r_i, \bar{t}_i)$. 
Thus, we associate each $(h_i, r_i, t_i)$ with 
\begin{itemize}
	\item a cache $\mathcal{C}_{i}$,
	which keeps large-gradient triplets from $\bar{\mathcal{S}}_{i}$,
	to store a set of $(\bar{h}_i, r_i, t_i)$ or $(h_i, r_i, \bar{t}_i)$ and the corresponding gradient norms $\|\mathbf{g}_i\|$
	(given by \eqref{eq:graddist} or \eqref{eq:gradsema}).
\end{itemize}

However, since the size of $\bar{\mathcal{S}}_{i}$ is very large,
evaluating all of them in $\bar{\mathcal{S}}_{i}$ to pick up the large-gradient triplets is intractable.
The proposed method
will adaptively sample a pair of positive and negative triplets directly through the cache. 
In the sequel,
we show how the cache is updated and sampled.

\subsubsection{Algorithm Framework} 
\label{ssub:algfra}

Algorithm \ref{alg-cache} shows the KG embedding framework based on our cache-based negative sampling scheme.
Note that the proposed algorithm does not depend on the choice of scoring functions,
all those in Table~\ref{tb-score-func} can be used here.
In Algorithm~\ref{alg-cache}:
first,
a pair of positive triplet and negative triplet is sampled
in step~\ref{step:sample};
then, the cache is updated in step~\ref{step:updatecache};
finally, 
in step~\ref{step:updateemb},
the embeddings are updated based on the choice of scoring functions and loss functions.

\begin{algorithm}[ht]
\caption{NSCaching: Cache-based KG embedding.}
\label{alg-cache}
\begin{algorithmic}[1]
	\REQUIRE training set $\mathcal{S} = \{(h, r, t)\}$, embedding dimension $d$, scoring function $f$.
	\STATE initialize embeddings for each $e \in \mathcal E$ and $r \in \mathcal R$, and cache $\mathcal{C}$;
	\FOR{$i = 1, \cdots, T$}
	\STATE sample a pair of positive and negate triplet, 
	i.e., $(h,r,t)$ and $(\bar{h}, r, \bar{t})$,
	using Algorithm~\ref{alg-cache-sample}; \label{step:sample}
	\IF {$i\% (n+1) == 0$}
	\STATE update the cache $\mathcal{C}_{i}$  using Algorithm~\ref{alg-cache-update}; \label{step:updatecache}
	\ENDIF
	\STATE update embeddings 
	using \eqref{eq:graddist} or \eqref{eq:gradsema}; \label{step:updateemb}
	\ENDFOR
\end{algorithmic}
\end{algorithm}

\begin{table*}[ht]
	\caption{Comparison of the proposed approach with state-of-the-arts,
		which address the negative sample.
		Model parameters are based on TransE,
		$m$ is the size of mini-batch,
		$n$ is the epoch of lazy-update.}
	\centering
	\label{tab:compare}
	\renewcommand{\arraystretch}{1.3}
%	\begin{tabular}{c | C{30px} | C{30px} | c | c | c | c }
	\begin{tabular}{c | c | c | c | c | c | c }
		\hline
		&  \multicolumn{2}{c|}{triplet sampling}   & \multirow{2}{*}{training}          &     \multicolumn{2}{c|}{mini-batch computation}      & model                             \\
		& positive & negative                      &                                    & time                         & space                 & parameters                        \\ \hline
		baseline              & \multicolumn{2}{c|}{uniform random (UR)} & gradient descent (from scratch)    & $O(m d)$                     & $O(m d)$              & $ (|\mathcal E|+|\mathcal R|) d$  \\ \hline
		IGAN \cite{wang2018incorporating} & UR       & GAN                           & reinforce learning (with pretraining) & $O(m |\mathcal E| d)$        & $O(m |\mathcal E| d)$ & $3(|\mathcal E|+|\mathcal R|) d $ \\ \hline
		KBGAN \cite{cai2018kbgan}     & UR       & GAN                           & reinforce learning (with pretraining)   & $O(m N_1 d)$                 & $O(m N_1  d)$         & $2(|\mathcal E|+|\mathcal R|) d $ \\ \hline
		NSCaching             &   \multicolumn{2}{c|}{using the cache}   &  gradient descent (from scratch)  & $O( \frac{m}{n+1} (N_1+N_2)d)$ & $O(m (N_1+N_2)d)$     & $ (|\mathcal E|+|\mathcal R|) d$  \\ \hline
	\end{tabular}
\end{table*}

An overview comparison of the proposed method
with state-of-the-art negative sampling method is in Table~\ref{tab:compare}.
The main difference with general KG embedding framework in Algorithm~\ref{alg:basic} is step~\ref{step:sample}
in Algorithm~\ref{alg-cache},
where the sampling scheme is based on the cache rather than a uniform Bernoulli sampling.
Besides,
compared with 
the complex GAN-based works \cite{wang2018incorporating,cai2018kbgan},
our method in Algorithm~\ref{alg-cache} acts like a discriminative and distilled model of GAN,
and it only cares about negative triplets leading to large gradient norms during the training.
Thus,
the proposed
NSCaching algorithm
not only has fewer parameters,
but also can be easily trained from randomly initialized models (from the scratch).
Moreover,
experimental results in Section~\ref{sec:exp} show that NSCaching achieves the best performance.

However,
in order to achieve a good performance,
we need to carefully design how to sample from the cache (step~\ref{step:sample}) and how to update the cache (step~\ref{step:updatecache}).
In these steps,
``\textit{exploration and exploitation}'' (E\&E) \cite{march1991exploration} is the main concern.
Specifically,
how to keep the balance between \textit{exploration} 
(explore all the possible large-gradient negative triplets in $\bar{\mathcal S}$)
and \textit{exploitation} (sample the training pair leading to the largest gradient norm in cache $\mathcal C$).

\noindent
\textit{1). Sampling from the cache (step~\ref{step:sample}).}
Before describing the sampling scheme,
we introduce some notations for subsequent usage.
Let $\mathbf{c}^{(i)}$ be a $N_1$-dimensional vector containing
gradient norms of $(\bar{h}_{i_j}, r_i, \bar{t}_{i_j}) \in \mathcal{C}_i, j=1\dots N_1$.
Thus,
if the $j$-th element  $\mathbf{c}^{(i)}_j$
is larger,
it means that the $j$-th negative triplet $(\bar{h}_{i_j}, r_i, \bar{t}_{i_j})$ in cache is of higher quality.
Finally,
we further define a vector $\mathbf{p}$,
of which the length is the number of positive triplets $\mathcal S$
and each element $\mathbf{p}_i = \| \mathbf{c}^{(i)} \|_2$.
Intuitively,
if $\mathbf{p}_i$ is larger,
then $(h_i, r_i, t_i)$ is more likely to have more large-gradient negative triplets. 

As in Algorithm~\ref{alg-cache}, we need to sample a pair of 
positive and negative triplets.
Based on above notations, 
we can do it as follows.
First,
we can pick up a positive triplet 
$(h_i, r_i, t_i) \in \mathcal{S}$
following 
a probability distribution given by 
\begin{align}
p((h_i, r_i, t_i)) 
= \sigma_1(\mathbf{p}_i; \mathbf{p}),
\label{eq-samp-pos}
\end{align}
\begin{algorithm}[ht]
	\caption{Sampling from the cache (step~3).}
	\begin{algorithmic}[1]
		\REQUIRE Training set $\mathcal S$ and cache $\mathcal C$.
		\STATE sample a positive triplet $(h_i, r_i, t_i)\in\mathcal S$ according to $p\left((h_i, r_i, t_i)\right)$ in \eqref{eq-samp-pos};
		\STATE index the specific cache $\mathcal{C}_{i}$ of $(h_i, r_i, t_i)$;
		\STATE sample a negative triplet $(\bar{h}_j, r_j, \bar{t}_j)$ from $\mathcal{C}_i$ according to $p\left((\bar{h}_{i_j}, r_i, \bar{t}_{i_j}) \right)$ in \eqref{eq-samp-neg}.
	\end{algorithmic}
	\label{alg-cache-sample}
\end{algorithm}
where the distribution $\sigma_1(\mathbf{p}_i; \mathbf{p})$ satisfies that
\begin{itemize}[leftmargin=*]
	\item  $\sum_{a} \sigma_1(\mathbf{p}_a; \mathbf{p}) = 1$; and $\sigma_1(\mathbf{p}_a; \mathbf{p}) \ge \sigma_1(\mathbf{p}_b; \mathbf{p})$ if $\mathbf{p}_a \ge \mathbf{p}_b$.
\end{itemize}
In this way, $(h_i, r_i, t_i)$ will be more frequently sampled
if $\mathbf{p}_i$ is larger.
Then,
after picking up the positive triplet $(h_i, r_i, t_i)$,
we sample the negative triplet $(\bar{h}_{i_j}, r_i, \bar{t}_{i_j}) \in {\mathcal{C}}_i$ following
\begin{align}
p\left((\bar{h}_{i_j}, r_i, \bar{t}_{i_j}) \right)
= \sigma_2( \mathbf{c}^{(i)}_j; \mathbf{c}^{(i)} ),
\label{eq-samp-neg}
\end{align}
where $\sigma_2(c)$ is  defined in the same way as $\sigma_1(p)$.
The full procedures are shown in
Algorithm~\ref{alg-cache-sample}.

\begin{remark} \label{rmk:sample}
	The choice of $\sigma$ is important,
	as it greatly affects E\&E and how we can adapt to the sampling distributions.
	Let us consider two extreme examples.
	First,
	if we pick $\sigma_1$ as an indicator function (as in Figure~\ref{fig:alphainf})
	where $\sigma_1(\mathbf{p}_i; \mathbf{p}) = 1$ if $\mathbf{p}_i$ is the largest and $\sigma_1(\mathbf{p}_j; \mathbf{p}) = 0$ for $j \not = i$.
	Then,
	it is equal to deterministically select the negative triplet $(h_i, r_i, t_i)$ with the highest-quality.
	However,
	as the distribution can change during iterations of the algorithm,
	both of the embedding quality and the negative triplets in the cache may not be accurate enough for the sampling in the latest iteration.
	Besides,
	consistently sampling the largest one may make the algorithm only focus on a small amount of triplets, 
	failing to capture the distribution well.
	Thus, 
	we also need to consider the other candidates except the one with the largest $\mathbf p_i$.
	Second,
	if we take $\sigma_1(\mathbf{p}_i; \mathbf{p}) = 1/N$ for any $i \in \{1, \dots, N \}$ (as in Figure~\ref{fig:alpha0})
	where $N$ is the total number of training triplets
	(i.e., uniform sampling is used),
	then all triplets have equal possibilities to be sampled.
	However,
	this ignores the difference of candidates. 
	The two cases also adapt to $\sigma_2$ when negative triplets are sampled from $\mathcal{C}_{i}$.
	In Section~\ref{sec:autoML},
	we will propose a novel method to balance E\&E automatically.
\end{remark}

\begin{algorithm}[ht]
	\caption{Updating the cache (step~4).}
	\label{alg-cache-update}
	\begin{algorithmic}[1]
		\REQUIRE  cache $\mathcal{C}_{i}$ of size $N_1$.
		\STATE initialize $\hat{\mathcal{C}}_{i} \leftarrow \emptyset$, $\hat{\mathbf c}^{(i)} = \mathbf 0$.
		\STATE sample a subset $\mathcal N_i\subset  \bar{\mathcal S}_{i}$ with $N_2$ triplets;
		\STATE compute $\| \mathbf g_k \|_2$ for all $(\bar{h}, r, \bar{t}) \in \mathcal N_i\cup\mathcal{C}_{i}$;
		\FOR{$j = 1, \cdots, N_1$}
		\STATE sample $(\bar{h}_{i_k},r_i,\bar{t}_{i_k})$  with probability in \eqref{eq:prupc};
		\STATE remove $(\bar{h}_{i_k},r_i,\bar{t}_{i_k})$ from $\mathcal N_i\cup\mathcal{C}_{i}$;
		\STATE $\hat{\mathcal{C}}_{i} \leftarrow \hat{\mathcal{C}}_{i} \cup \{(\bar{h}_{i_k},r_i,\bar{t}_{i_k})\} $;
		\STATE $\hat{\mathbf c}_k^{(i)} = \|\mathbf g_k\|_2$;
		\ENDFOR
		\STATE update by $\mathcal{C}_{i} \leftarrow \hat{\mathcal{C}}_{i}$.
		\RETURN $\mathcal{C}_{i} \leftarrow \hat{\mathcal{C}}_{i}$ 
		and $\mathbf{c}^{(i)} \leftarrow \hat{\mathbf c}^{(i)}$.
	\end{algorithmic}
\end{algorithm}

\noindent
\textit{2). Updating the cache (step~\ref{step:updatecache}).}
\label{sssec:update}
As mentioned in Section~\ref{sec:closer},
the cache needs to be dynamically 
changed during iterations of the algorithm.
Otherwise,
the content in cache will not be changed
and the sampling will be highly biased 
since most of the negative triplets will not be visited.
Thus,
we need to refresh the cache periodically.
Moreover,
the cache needs to be updated in an efficient way.

As in \eqref{eq:nega}, the number of negative triplets in $\bar{\mathcal S}_i$ is quite large for a given positive triplet $(h_i, r_i ,t_i)$.
However, it is impossible for us to evaluate all the candidates in $\bar{\mathcal S}_{i}$. 
Since we want to efficiently capture the large-gradient negative triplets in $\mathcal{C}_{i}$, 
we sample a small subset $\mathcal N_i \subset \bar{\mathcal S}_{i}$ of size $N_2$, 
with $N_2 \ll \left|\bar{\mathcal S}_i\right|$.
Then
for each $(\bar{h}_{i_k},r_i,\bar{t}_{i_k})\in\mathcal N_i \, \cup \, \mathcal{C}_{i}$,
we evaluate the gradient norm $\|\mathbf g_k\|_2$ by \eqref{eq:graddist} or \eqref{eq:gradsema}.
Then we construct a new set $\hat{\mathcal{C}}_{i} \subset \mathcal N_i\cup\mathcal{C}_{i}$,
whose components are sampled from $\mathcal N_i\cup\mathcal{C}_{i}$ without replacement $N_1$ times
following the probability distribution
\begin{equation}
p\left((\bar{h}_{i_k},r_i,\bar{t}_{i_k}) \right) = \sigma_3\left(\mathbf g_k; \mathbf{g} \right).
\label{eq:prupc}
\end{equation}
Finally, $\hat{\mathcal{C}}_{i}$, which contains $N_1$ negative triplets 
and their corresponding gradient norms $\hat{\mathbf{c}}^{(i)}$,
are returned.

\begin{remark} \label{rmk:update}
	Exploration and exploitation also need to be carefully balanced in Algorithm~\ref{alg-cache-update}.
	As the cache needs to be updated,
	we have to sample from $\bar{\mathcal{S}}_{i}$.
	The subset $\mathcal N_i$ is chosen as a substitute of  $\bar{\mathcal{S}}_{i}$ in consideration of efficiency.
	Therefore,
	a bigger $N_1$ implies more exploitation,
	while a larger $N_2$ leads to more exploration.
	In step~5,
	indeed, the choice of $\sigma_3$ is important under the same consideration as $\sigma_1$ and $\sigma_2$.
	The balance of E\&E on $N_1$, $N_2$ and $\sigma_3$ is further discussed in Section \ref{sec:autoML}.
\end{remark}

\subsubsection{Space and Time Complexities}
\label{sssec:complexity}
In this part,
we analyze the space and time complexities of NSCaching (Algorithm~\ref{alg-cache}).
Comparing with basic training framework in Algorithm~\ref{alg:basic},
the main additional cost by introducing cache comes from step~\ref{step:updatecache} in Algorithm~\ref{alg-cache}, 
i.e. updating the cache using Algorithm~\ref{alg-cache-update}.
In Algorithm \ref{alg-cache-update},
the main time cost comes from 
computing the gradients $\|\mathbf g_k\|_2$ for $N_1+N_2$ training pairs, 
whose complexity is $O((N_1+N_2)d)$.
The cost of step~\ref{step:sample} in Algorithm~\ref{alg-cache} is rather small,
which comes from importance sampling according to the gradient norms.
This part takes $O(N_1)$ time.
Hence, the total cost by introducing the cache is $O((N_1+N_2)d)$ for a single training pair.
In practice,
we can lazily update the cache every $n$ epochs rather than do immediate updating,
which can further reduce the updating complexity to $O\left(\nicefrac{(N_1+N_2)d}{(n+1)}\right)$.

As for the space complexity, 
evaluating the gradients for $N_1+N_2$ training pairs takes $O((N_1+N_2)d)$ space. 
Since we only store indexes in the cache, 
it takes $ O(|\mathcal S|N_1)$ space to store these indexes for negative triplets.
Note that,
$N_1$ is small
since large-gradient negative triplets are rare.
This is also verified in our experiments in Section~\ref{ssec:exp:size}.

In comparison, to generate a training pair, the generator in IGAN \cite{wang2018incorporating} takes $O(|\mathcal E|d)$ time since it needs to compute the distribution over all entities. 
KBGAN \cite{cai2018kbgan} needs $O(N_1 d)$ time to measure a candidate set of $N_1$ triplets. The additional space cost for IGAN and KBGAN is also $O(|\mathcal E|d)$ and $O(N_1 d)$ respectively.
Finally,
the comparison of space and time complexities is summarized in Table~\ref{tab:compare} with TransE as the scoring function. 

\subsection{Theoretical Analysis}
\label{sec:theory}

In this part,
we theoretically analyze
the convergence
and learning performance of the proposed method.

\subsubsection{Convex Case: Faster convergence}
\label{ssec:convex}

Before presenting our analysis,
we first simplify and take a uniform treatment over
\eqref{eq:distance} and \eqref{eq:sematic}.
Let $\mathbf{w} = \{ \mathbf{h}, \mathbf{r}, \mathbf{t} \}$,
then we can take the loss $\phi_i(\mathbf{w})$ on a training pair
 for translational distance model as 
\begin{align}
\phi_i(\mathbf{w})
= \left[\gamma - f(h_i, r_i, t_i) + f(\bar{h}_i, r_i, \bar{t}_i)\right]_+,
\label{eq:phitrans}
\end{align}
and for semantic matching model as
\begin{align}
\phi_i(\mathbf{w})
= \ell\left(1, f(h_i, r_i, t_i) \right) 
+ \ell\left(-1, f(\bar{h}_i, r_i, \bar{t}_i) \right).
\label{eq:phismena}
\end{align}
Thus,
we can express \eqref{eq:distance} and \eqref{eq:sematic} as 
\begin{align}
\min_{\mathbf{w}} F(\mathbf{w})
\equiv
\frac{1}{n} \sum\nolimits_{i = 1}^n \phi_i(\mathbf{w}),
\label{eq:genobj}
\end{align}
where $n$ is the number of
all the training pair of positive and negative triplets.
Using above notation,
we can abstract NSCaching (Algorithm~\ref{alg-cache})
as in Algorithm~\ref{alg:prototype}.
Basically,
the cache scheme is used to
generate a probability distribution $\mathbf{p}^t$ over all $\phi_i$,
which changes over iterations.

\begin{algorithm}[H]
	\caption{Abstraction of NSCaching.}
	\begin{algorithmic}[1]
		\FOR{$t = 1, \cdots, T$}
		\STATE sample $\phi_{i_t}$ from $\{ \phi_i \}_{i = 1}^n$ based on $\mathbf{p}^t$;
		\STATE 
		update $\mathbf{w}^{t + 1}$ by 
		$\mathbf w^{t+1} = \mathbf w^t - \eta(n\mathbf p^t_i)^{-1}\nabla \phi_{i_t}(\mathbf w^t)$;
		\ENDFOR
	\end{algorithmic}
	\label{alg:prototype}
\end{algorithm}

The convergence of Algorithm~\ref{alg:prototype} is in Theorem~\ref{thm:conv},
which is inspired by some recent works in stochastic optimization \cite{zhao2015stochastic,needell2014stochastic}.

\begin{theorem} \label{thm:conv}
	If $F$ is smooth and convex, then
	\begin{align}
	&
	\frac{1}{T}\sum\nolimits_{t =1}^T \mathbb{E} 
	\left[ F(\mathbf{w}^t) \right]  - 
	\mathbb{E} 
	\left[ F(\mathbf{w}^*) \right] 
	\notag
	\\
	& 
	\le
	\frac{2}{\eta T} \| \mathbf{w}^* \! - \! \mathbf{w}^t \|^2
	\! + \! \frac{\eta}{2\sigma T} \sum\nolimits_{t = 1}^T 
	\mathbb{E}
	\left[
	\| \nicefrac{\nabla \phi_{i_t} (\mathbf{w}^t)}{n \mathbf{p}^t_{i_t}} \|^2
	\right], 
	\label{eq:bound}
	\end{align}
	where 
	$\mathbf{w}^* \!\! = \! \arg\min_{\mathbf{w}} F$,
	$\eta$ is the step-size for stochastic optimization,
	and expectation is taken w.r.t. distribution $\mathbf{p}^t$.
\end{theorem}

The proof is in Appendix~\ref{prf:thm:conv}.
As we can see,
how fast and well 
$\mathbf{w}^t$ converges to the optimal solution depends on $\mathbf{p}^t$
via the second term in \eqref{eq:bound}.
The solution which minimizes this term is offered in Proposition~\ref{pr:dist}.

\begin{proposition}[\cite{zhao2015stochastic}] \label{pr:dist}
	$\mathbb{E}
	\left[
	\| \nicefrac{\nabla \phi_{i_t} (\mathbf{w}^t)}{n \mathbf{p}^t_{i_t}} \|^2
	\right]$
	is minimized when the possibility $\mathbf{p}^t$
	follows
	$\mathbf{p}^t_i
	= \nicefrac{\| \nabla \phi_i (\mathbf{w}^t) \|}{\sum_{j = 1}^n \| \nabla \phi_j (\mathbf{w}^t) \|}$.
\end{proposition}

Since the cache scheme is used to avoid vanishing gradient problem 
and distill the full sampling distribution,
the samples with 
larger $\| \nabla \phi_i(\mathbf{w}^t) \|$
should have higher possibility to be sampled.
If the dynamic distribution $\mathbf p^t$ is captured,
we can then adaptively sample from it.
In other words, the sample $i_t$  with larger $\mathbf{p}_{i_t}^t$ has
larger possibility to be sampled.
As a result,
the last term in \eqref{eq:bound} in NSCaching can have smaller value compared with the uniform sampling.
This indicates that
NSCaching has both faster convergence speed and smaller approximation error.
In practice,
most of the existing embedding models are non-convex
and stochastic optimization \cite{kingma2014adam} is used to update the parameters.
However,
the above bound still offers insights on 
how the proposed method works.

\subsubsection{Nonconvex Case: Self-paced learning}
\label{sec:self-pace}

The main idea 
of self-paced  learning (or curriculum learning) \cite{bengio2009curriculum,kumar2010self}
is to pick up easy samples first,
and then gradually switch to harder ones.
In this way,
the classifier can firstly identify the rough position where the decision boundary should locate.
Then the boundary can be further refined by the hard examples.
It is effective for complex and noncovex models.
Recently,
self-paced learning is also introduced into graph embedding 
and the improvement on the quality of embeddings has been reported \cite{gao2018self}.
Besides,
GAN is also used to monitor the distribution of edges in the network,
and negative edges with scores above a given threshold are sampled from the generator in GAN. 
Self-paced learning is achieved by increasing the threshold during the training of embedding \cite{gao2018self}.
As a comparison,
the GAN models used in KBGAN and IGAN are not benefited from self-paced learning.

In contrast,
our caching scheme can explicitly benefit from it.
The reason is that 
the embedding model only has weak discriminative ability in the beginning of the training. 
Thus,
while there exist a lot of negative triplets leading to large gradient norms,
it is more likely that they are easy ones as most of the negative samples are easily classified.
As training process continuous,
those easy samples will gradually have small gradients and are removed from the cache.
These mean NSCaching will learn from easy samples first,
but then gradually focus on hard ones,
which is exactly the principle of self-paced learning.
The above explanations are also verified by
experiments,
where we can see 
the negative triplets in the cache change from easy to hard ones (Section~\ref{sec:visself})
and NSCaching (training from scratch) can already achieve better performance
than IGAN and KBGAN with pretraining (Section~\ref{ssec:compstate}).

\begin{figure*}[ht]
\centering
\subfigure[uniform function $\alpha=0$]
{\includegraphics[width=0.3\textwidth]{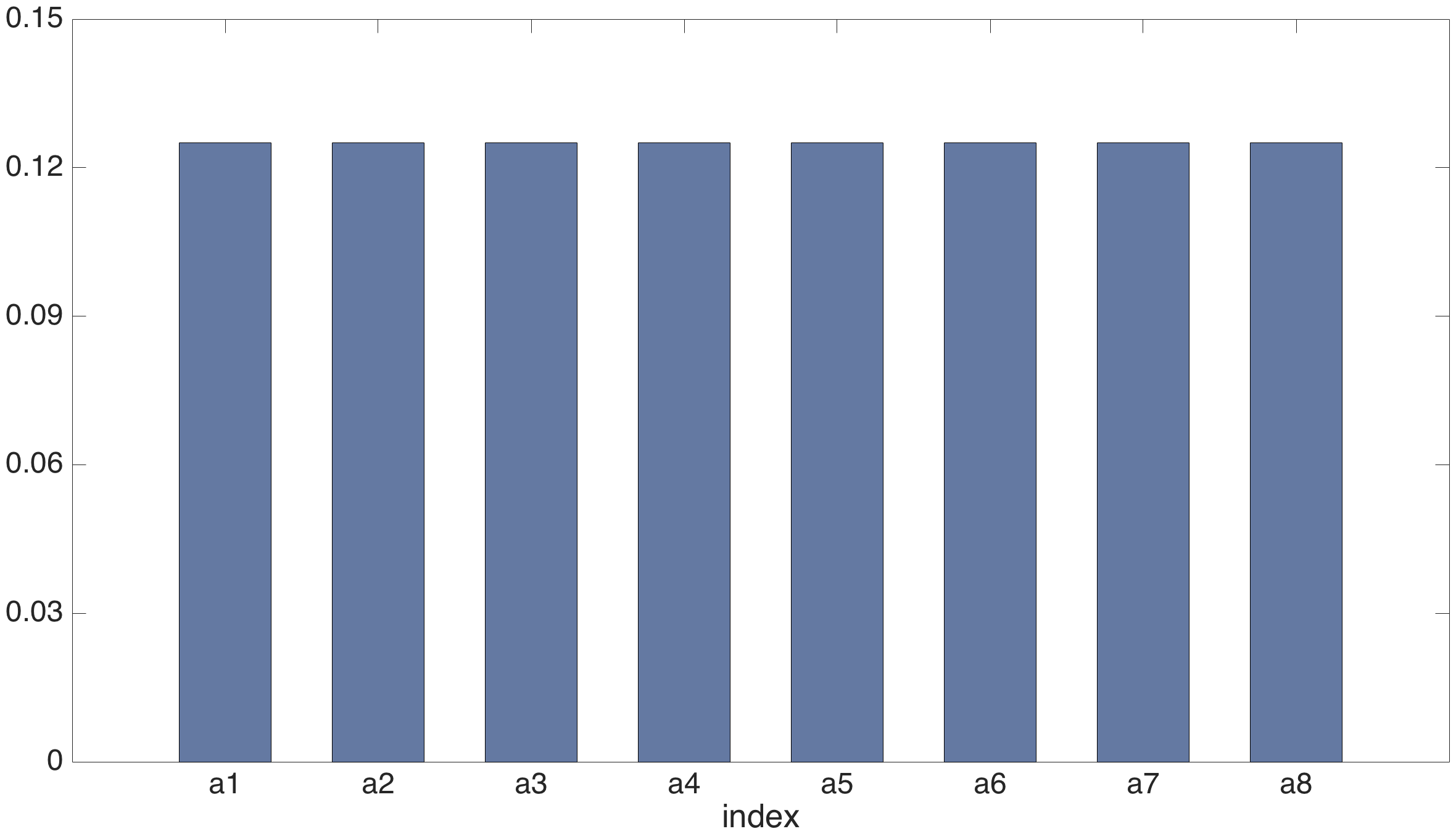} \quad
\label{fig:alpha0}}
\subfigure[general softmax $\alpha=1$]{
	\includegraphics[width=0.3\textwidth]{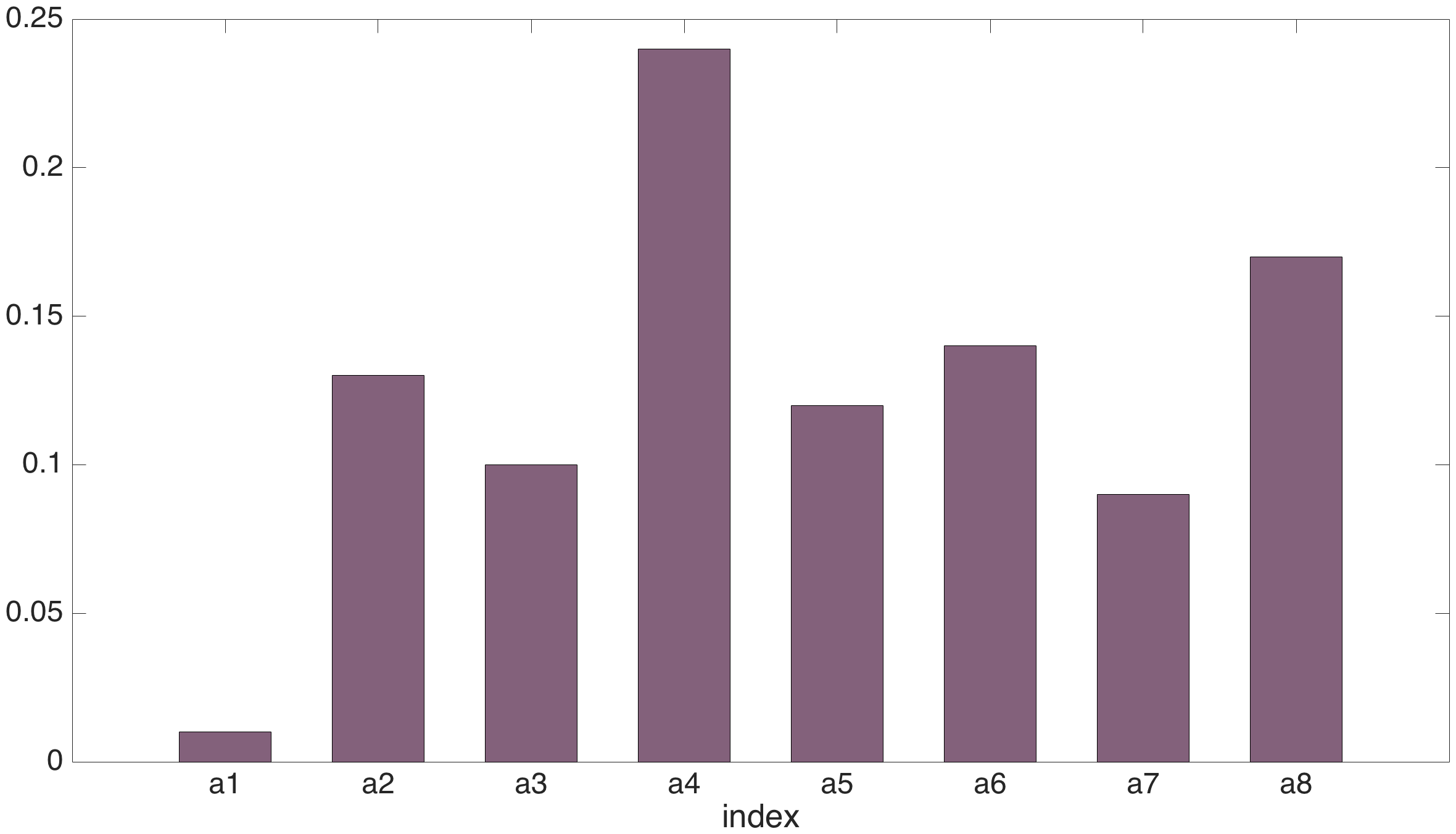}\label{fig:alpha1}}\quad
\subfigure[indicator function $\alpha=\inf$]{
	\includegraphics[width=0.3\textwidth]{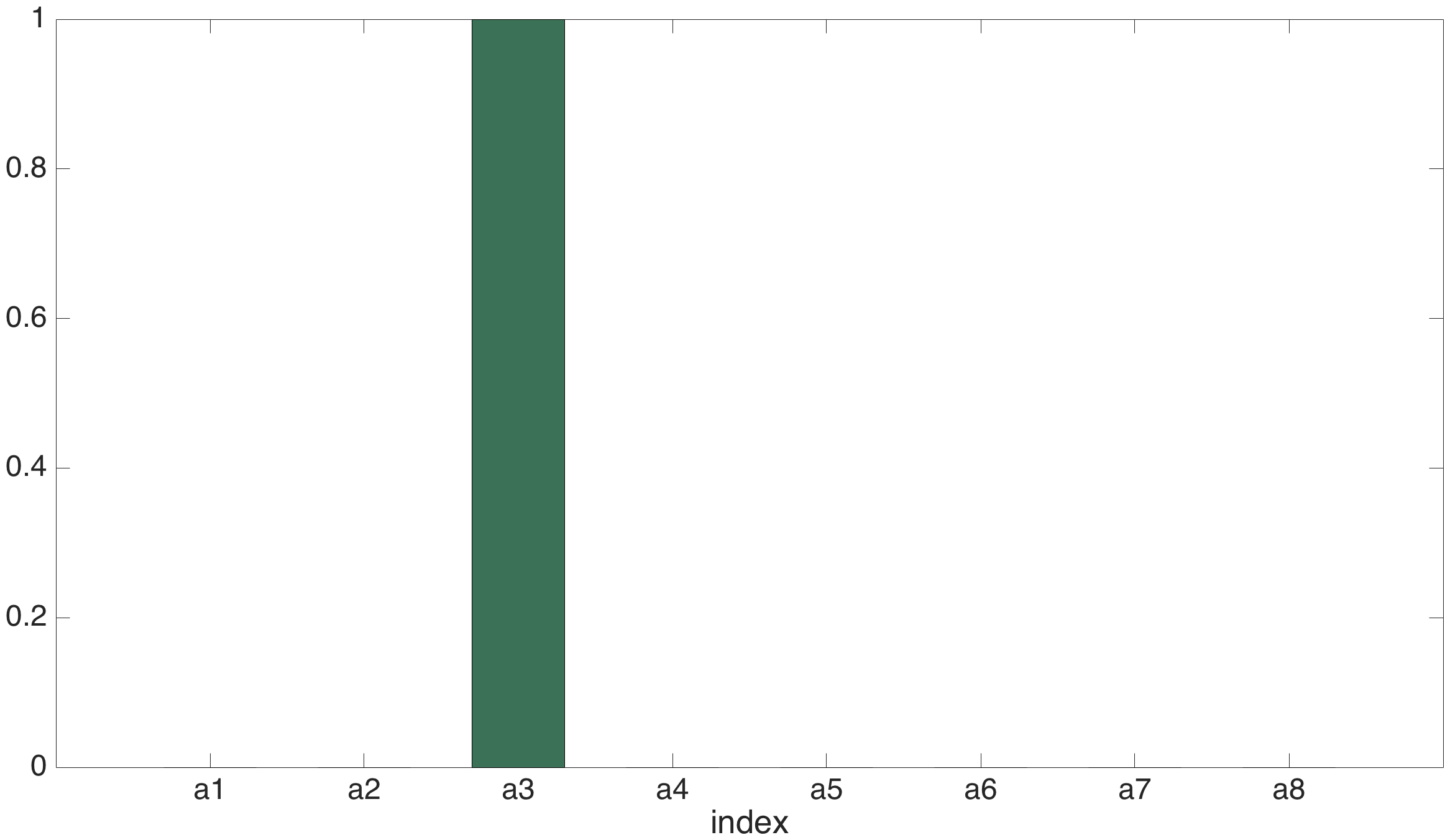}\label{fig:alphainf}}
\caption{Example of the different distributions of $\sigma$,
	where $x$-axis indicates each dimension of $\mathbf{a}$ in \eqref{eq:sigma} and $y$-axis is the sampling probability.}
\label{fig:sigma}
\end{figure*}

\subsection{Automatic Balancing Exploration and Exploitation}
\label{sec:autoML}

In previous parts,
we have described the proposed framework (Section~\ref{sec:NSCaching})
and analyzed why it works (Section~\ref{sec:theory}).
Based on Proposition~\ref{pr:dist},
we aim to capture the dynamic sampling distribution $\mathbf{p}^t_i
= \nicefrac{\| \nabla \phi_i (\mathbf{w}^t) \|}{\sum_{j = 1}^n \| \nabla \phi_j (\mathbf{w}^t) \|}$.
To guarantee efficiency,
we design the sampling scheme (Algorithm~\ref{alg-cache-sample}) 
and updating scheme (Algorithm~\ref{alg-cache-update})
to distill such a distribution.
However,
the distributions for different tasks and for different training status are different in practice.
Therefore, we should carefully adjust the sampling and updating schemes.
As mentioned in Section~\ref{sec:NSCaching},
to achieve better performance for different scenarios,
the remained question is
how can we carefully balance E\&E?
Here,
we show 
how AutoML techniques can be combined with the proposed framework
to automatically balance E\&E.

\subsubsection{Search Space from NSCaching}

From Remarks~\ref{rmk:sample}, \ref{rmk:update} and Proposition~\ref{pr:dist},
we can see that $\sigma(\mathbf{a}_i;\mathbf{a})$ needs to cover three special cases,
i.e.,
(i). uniformly sampling on all elements,
(ii). deterministically sampling the max,
and (iii). importance sampling as Proposition~\ref{pr:dist}.
Thus,
we are motivated to choose the weighted softmax distribution as the probability function
\begin{align}
\sigma(\mathbf{a}_i; \mathbf a) = \nicefrac{\exp(\alpha\cdot\mathbf a_i)}{\sum_j\exp(\alpha\cdot \mathbf a_j)},
\label{eq:sigma}
\end{align}
where $\alpha \ge 0$ is a hyper-parameter to be tuned.
We can see $\alpha = 0$ covers (i) as in
Figure~\ref{fig:alpha0},
$\alpha = \infty$ covers (ii) as in Figure~\ref{fig:alphainf},
and other values of $\alpha$ cover (iii) as in Figure~\ref{fig:alpha1}.
Specifically,
we use three different $\alpha$'s for \eqref{eq-samp-pos},
\eqref{eq-samp-neg} and \eqref{eq:prupc} respectively.

\subsubsection{Search by Bayesian Optimization }
\begin{table}[ht]
	\centering
	\caption{How exploration and exploitation (E\&E) in Algorithm~\ref{alg-cache}
		are affected by hyper-parameters. }
	\renewcommand*{\arraystretch}{1.25}
	\begin{tabular}{c| c|c}
		\hline
		hyper-parameters & functionality                & larger leads to \\ \hline
		$\alpha_1$    & positive sample $(h_i,r_i,t_i)$             & exploitation    \\ \hline
		$\alpha_2$    & negative sample $(\bar{h}_i,r_i,\bar{t}_i)$ & exploitation    \\ \hline
		$\alpha_3$    & update $\mathcal{C}_{i}$ & exploitation    \\ \hline
		$N_1$      &  size of cache $\mathcal C_i$                  & exploitation    \\ \hline
		$N_2$      &  size of subset $\mathcal N_i$                  & exploration     \\ \hline
	\end{tabular}
	\label{tab:eepara}
\end{table}
All hyper-parameters balancing E\&E are summarized in Table~\ref{tab:eepara}.
Manually tuning these hyper-parameters is time consuming.
Simple search approaches such as grid search and random search are usually inefficient.
Inspired by the choice of \eqref{eq:sigma}, and the recent success of 
automated machine learning (AutoML) \cite{yao2018taking}, 
especially hyper-parameter optimization,
we use a versatile
hyper-parameter optimization method,
i.e., Sequential Model-based optimization for general Algorithm Configuration (SMAC) \cite{hutter2011sequential}.
SMAC allows efficiently and automatically
tuning of both discrete ($N_1$ and $N_2$) and continuous 
($\alpha_1$, $\alpha_2$ and $\alpha_3$) hyper-parameters.

\subsubsection{Discussion: connection with existing methods}
\label{ssec:relation:exist}

The hyper-parameters in Table~\ref{tab:eepara} can not only help us balance E\&E,
but also give a unified view of the baselines,
namely covering Bernoulli, KBGAN and IGAN as special cases.
In this part, $\alpha_1$ is always 0 since none of the three methods
consider non-uniform positive sampling.

\begin{itemize}

\item \textit{Bernoulli} \cite{wang2014knowledge}:
In Bernoulli sampling, negative samples are generated uniformly from the whole candidate space.
In this case, we can set both 
$\alpha_2$ and $\alpha_3$ to be 0.
%the sampling hyper-parameter $\alpha_2$ and updating one $\alpha_3$ to be 0.
Then, the cache updating is never dependent on scores as well as the sampling schemes.
Therefore, Bernoulli sampling is a special case of NSCaching when  $\alpha_2 = \alpha_3=0$;

%		\begin{table}[ht]
%		\centering
%		\caption{A unified view of the proposed method and the existing methods. ``-'' means not care. }
%		\label{tab:unified-view}
%		\renewcommand*{\arraystretch}{1.2}
%%		\begin{tabular}{C{35px}|C{33px}|C{60px}|C{33px}}
%	\begin{tabular}{c|c|c|c}
%			\hline
%			{hyper-} & \multirow{2}{*}{Bernoulli \cite{wang2014knowledge}} &KBGAN \cite{cai2018kbgan} / & \multirow{2}{*}{IGAN \cite{wang2018incorporating}} \\
%			parameters & & {Self-Adv \cite{sun2018rotate}} & \\ \hline
%			$\alpha_1$    &                  0                   &              0              &                  0               \\ \hline
%			$\alpha_2$    &                  0                   &            $>0$             &                $>0$             \\ \hline
%			$\alpha_3$    &                  0                   &              0              &                  -               \\ \hline
%			$N_1$       &                  -                   &              -              &           $\max_i\left(|\bar{\mathcal S}_i|\right)$        \\ \hline
%			$N_2$       &                  -                   &             $\max_i\left(|\bar{\mathcal S}_i|\right)$             &                  0         \\ \hline
%		\end{tabular}
%	\end{table}

	\begin{table}[ht]
	\centering
	\caption{A unified view of the proposed method and the existing methods. ``-'' means not care. }
	\label{tab:unified-view}
	\renewcommand*{\arraystretch}{1.2}
%	\begin{tabular}{C{35px}|C{33px}|C{60px}|C{33px}}
			\begin{tabular}{c|c|c|c}
		\hline
		{hyper-} & \multirow{2}{*}{Bernoulli \cite{wang2014knowledge}} &KBGAN \cite{cai2018kbgan} / & \multirow{2}{*}{IGAN \cite{wang2018incorporating}} \\
		parameters & & {Self-Adv \cite{sun2018rotate}} & \\ \hline
		$\alpha_1$    &                  0                   &              0              &                  0               \\ \hline
		$\alpha_2$    &                  0                   &            $>0$             &                $>0$             \\ \hline
		$\alpha_3$    &                  0                   &              0              &                  -               \\ \hline
		$N_1$       &                  -                   &              -              &           $\max_i\left(|\bar{\mathcal S}_i|\right)$        \\ \hline
		$N_2$       &                  -                   &             $\max_i\left(|\bar{\mathcal S}_i|\right)$             &                  0         \\ \hline
	\end{tabular}
\end{table}

\item \textit{KBGAN} \cite{cai2018kbgan}:
The key thought in KBGAN is to use a generator to pick up large-gradient negative samples in a random subset	$\mathcal N_i\in\bar{\mathcal{S}}_i$.
Given a positive triplet $(h_i, r_i, t_i)$, 
the scores $\mathbf c_j^{(i)}$ stored in cache work as an alternative of the generator in KBGAN 
to measure the quality of $(\bar{h}_j, r_j, \bar{t}_j)$.
Different from standard NSCaching, 
we use $\alpha_3=0, \alpha_2>0$ and $N_2=\max_i\left(|\bar{\mathcal S}_i|\right)$
so that the content in cache $\mathcal C_i$ is similar as that in $\mathcal N_i$ of KBGAN.
Self-Adv \cite{sun2018rotate} uses in the similar approach as KBGAN.
Differently, 
Self-Adv uses the model's embedding itself to measure the quality of negative samples in $\mathcal N_i$.
Compared with KBGAN and Self-Adv,
NSCaching improves upon them by controlling the quality of negative triplets in $\mathcal N_i$.

\item \textit{IGAN} \cite{wang2018incorporating}:
When $N_1=\max_i\left(|\bar{\mathcal S}_i|\right)$ and $N_2=0$, NSCaching resembles IGAN. 
In IGAN, the generator chooses large-gradient negative samples from the entire candidate set $\bar{\mathcal S}_i$.
Thus, we can set the cache size to be $\max_i\left(|\bar{\mathcal S}_i|\right)$, and mask the positive positions.
Besides, the cache does not need to be updated by setting $N_2=0$.
We use $\alpha_2> 0$ to select large-gradient negative samples from cache to replace the generator.
In this way, NSCaching can also approximate the sampling procedure in IGAN.
	
\end{itemize}
We show the values of $\alpha$'s and $N$'s in Table~\ref{tab:unified-view}
about how to cover the baselines by NSCaching.
This finding also explains why NSCaching (auto) is better than the baselines.
Besides,
comparing with Bernoulli, KBGAN, IGAN and Self-Adv,
NSCaching (auto) adapts the sampling distribution to approximate a relatively unbiased estimator for specific tasks.

\subsection{Understanding the false negative triplets.}
\label{ssec:falseneg}
Since KG is incomplete \cite{wang2017knowledge},
there exist triplets that do not appear in the training set but are not necessarily false.
For example, the triplets in the valid or test set can be viewed as false negative triplets during training.
From the literature \cite{wu2017sampling,wang2017knowledge,li2019sampling},
the false negative samples will have larger scores but lower variance than the large-gradient negative samples.
We also have such an observation in 
Section~\ref{ssec:falseexp}.
Thus, 
the false negative triplets may be detected by tracing the variance during the training.
%While this observation on false negatives can be true,
However,
since the number of false negative triplets is small and evaluating the variance is expensive,
we show in 
Section~\ref{ssec:falseexp} that false negatives are empirically not a concern.

\section{Extension to Skip-gram Model}
\label{sec:graph}

The skip-gram model \cite{mikolov2013efficient}
is a popular method originating from word embedding,
which can explore surrounding words given the current embedded one.
Due to its superior performance in natural language processing,
two representative methods
Deepwalk \cite{perozzi2014deepwalk}
and Node2vec \cite{Grover2016} 
adopt the skip-gram model for graph embedding.
And
they have achieved significant improvements over previous works on the embedding quality \cite{cai2018comprehensive}.
Moreover,
motivated by the success of 
the skip-gram model on the graph and word embedding \cite{perozzi2014deepwalk,Grover2016,mikolov2013efficient,mikolov2013distributed},
we extend the NSCaching algorithm to the skip-gram model here.
We first introduce the skip-gram model \cite{mikolov2013efficient} in Section~\ref{ssec:skipgram}.
Then,
we adapt the 
cache-based negative sampling algorithm to the skip-gram model for graph embedding
in Section~\ref{ssec:graphembed}.
In this way,
we show that the NSCaching algorithm (Algorithm~\ref{alg-cache}) is not limited to KG embeddings.

\subsection{Negative Sampling in Skip-gram Model}
\label{ssec:skipgram}

Skip-gram model is originally used to learn word embeddings  \cite{mikolov2013efficient}.
It aims at maximizing the co-occurrence probability among the words that appear within a window $\mathcal W$.
Given a positive word $u$,
the training objective in skip-gram model is to learn word embeddings 
that are good at predicting the words $v\in\mathcal W_u$,
where $\mathcal W_u$ is a set of nearby or context words of $u$.
More formally,
for a sequence of training words $u_1, u_2, \dots, u_L$ with length $L$,
the objective is to maximize the average log probability
\begin{equation}
\frac{1}{L}\sum\nolimits_{i=1}^L \sum\nolimits_{v_j \in \mathcal W_{u_i}} \log p\left(v_j|u_i\right),
\end{equation}
where 
$\mathcal W_{u_i}= \{v_j|-c\leq j\leq c, j\neq 0 \} $, $c$ is a pre-defined window size.
Basically,
the probability $p\left(v_j|u_i\right)$ is defined as the softmax function
\begin{equation}
p\left(v_j|u_i\right) = \nicefrac{\exp\left(\bm v_j^\top \bm u_i\right)}{\sum_{k=1}^{|V|}\exp\left(\bm v_k^\top \bm u_i \right)  },
\label{eq:softmax}
\end{equation}
where the boldface represents the embedding and 
$|V|$ is the vocabulary size. 
However, since $|V|$ is usually large,
negative sampling is used to avoid computing the dot product similarity among all the words \cite{mikolov2013distributed}.
In this way, the log probability $\log p\left(v_j|u_i\right)$ is computed based on 
\begin{equation}
\log \sigma (\bm v_j^\top \bm u_i ) 
+ \sum\nolimits_{n=1}^N \log\sigma(-\bm v_n^\top \bm u_i),
\label{eq:skip:loss}
\end{equation}
where $\sigma(x) = \nicefrac{1}{1+\exp (-x)}$ is the sigmoid function and
$v_n$'s are the negative samples drawn from the noise distribution $p(u_i)$.
Generally,
the noise distribution $p(u_i)$ is a unigram distribution,
or a weighted distribution proportional to the word frequency \cite{mikolov2013distributed}.
Similar as the methods in Section~\ref{ssec:fixed},
the negative sampling used for \eqref{eq:skip:loss} is also fixed.
Hence, the quality of negative samples cannot be dynamically captured.

\subsection{Graph Embedding with Skip-gram Model}
\label{ssec:graphembed}

In order to preserve the graph structure
while make it easy for a machine learning model to process,
random walks are widely used to learn  graph embeddings \cite{perozzi2014deepwalk,Grover2016,cai2018comprehensive}.
A graph is firstly represented as a set of random walk paths sampled from it.
Then skip-gram model \cite{mikolov2013efficient} is applied to preserve graph properties carried by the paths \cite{cai2018comprehensive}.

DeepWalk \cite{perozzi2014deepwalk},
as a representative random walk based graph embedding model,
first samples a set of paths from the input graph.
Then, the sampled paths are regarded as sentences that describe the graph,
and the nodes are regarded as words.
Skip-gram model is applied on the paths to maximize
the probability of observing a node's neighborhood conditioned on its embedding.
In this way,
nodes with similar neighborhoods will have larger co-occurrence and more similar embeddings.
Node2vec \cite{Grover2016} improves upon DeepWalk \cite{perozzi2014deepwalk}
by using a biased random walk.
Two parameters $p$ and $q$ are used to control breadth-first sampling or depth-first sampling,
which is shown to better capture the local topologies \cite{Grover2016}.

Different from KG, skip-gram model does not have exact positive samples
since the context in the window $\mathcal W_{u_i}=\{v_j|-c\leq j \leq c,j\neq 0\}$ 
does not necessarily to have strong connection with $u_i$.
Therefore, we build a cache $\mathcal C_i$ for each word rather than each training sample.
Then,
the negative part in \eqref{eq:skip:loss} is sampled from the cache.
We treat the number of negative samples $N$ as a hyper-parameter and optimize it together with the cache-related hyper-parameters.
In addition,
we use all the nodes that are not in $\mathcal W_{u_i}$ as the negative samples.
Therefore, 
balancing between E\&E is also important in this setting.

We use the Node2vec \cite{Grover2016} model as the testbed for skip-gram model.
In Node2vec, a biased random walk method is used to generate a sequence of walks from the graph.
In this way, embeddings are updated through the cache-based skip-gram algorithm.
The cache-based Node2vec method is given in Algorithm~\ref{alg:graphemb}.

\begin{algorithm}[ht]
	\caption{Node2vec (NSCaching).}
	\label{alg:graphemb}
	\begin{algorithmic}[1]
		\REQUIRE Graph $G=(V, E)$, embedding dimension $d$, walks per node $r$, walk length $l$, window size $w$, $p$, $q$.
		\STATE Initialize embeddings for each node $v\in V$.
		\STATE computing the neighborhood sampling probability of each node based on $p$ and $q$;
		\STATE Initialize \textit{walks} to empty
		\FOR{$i = 1, \cdots, r$}
		\FOR{all nodes $v\in V$}
		\STATE sample a \textit{walk} starting from $u$ with length $l$;
		\STATE append \textit{walk} to \textit{walks}.
		\ENDFOR		
		\ENDFOR
		\REPEAT
		\STATE sample a node $u_i$ and its context $v_j$ in the window $\mathcal W_{u_i}$.
		\STATE sample a set of negative nodes $\bar{v}_n$'s from the cache $\mathcal C_i$.
		\STATE update embedding using the gradient of \eqref{eq:skip:loss}.
		\UNTIL{converge}
	\end{algorithmic}
\end{algorithm}

\section{Experiments}
\label{sec:exp}

In this section,
we conduct empirical study of our method. 
All algorithms are written in Python with PyTorch framework \cite{paszke2017automatic} 
and run on a TITAN Xp GPU with 12GB memory.
Our code is public available in xxx.

\subsection{Implementation Details}
\label{ssec:imp-details}
Since a lot of triplets share the same \textit{(head, relation)} or \textit{(relation, tail)} pairs,
we use two caches, namely a head cache $\mathcal H_{(r, t)}$ 
and a tail cache $\mathcal T_{(h, r)}$,
to separately store negative triplets in $\{(\bar{h}, r, t)\notin \mathcal S|\bar{h}\in\mathcal E\}$
and $\{(h, r, \bar{t})\notin \mathcal S|\bar{t}\in\mathcal E\}$.
Using two caches instead of one can help us to reduce the time and space cost.
The value we stored in cache is the score of negative triplets 
according to the pre-defined scoring function $f$,
instead of gradient norms.
The main consideration is that
gradient norms for each training pair can not be efficiently obtained through mini-batches,
especially for complex scoring functions like TransD \cite{ji2015knowledge}.
Given a positive triplet $(h_i, r_i, t_i)$,
the value $\mathbf p_i$ is computed by the sum of scores stored in $\mathcal H_{(r, t)}$ and $\mathcal T_{(h, r)}$.

In general, 
the training of KG embedding model is under the open world assumption,
which means that KGs contain only positive triplets 
and non-observed triplets can be either false or just missing \cite{wang2017knowledge}.
To reduce sampling the negative examples that are just missing,
we use the same scheme proposed in Bernoulli sampling \cite{wang2014knowledge}
to get the subset $\mathcal N_i$.
Specifically,
different probabilities are given when replacing the head or the tail
for different relations.
For each relation $r$,
we compute and denote the average number of tail entities per head as $tph$,
and the average number of head entities per tail as $hpt$.
Then the probabilities of replacing the head and the tail
are $\nicefrac{tph}{tph+hpt}$ and $\nicefrac{hpt}{tph+hpt}$ respectively.

To constrain values of $\alpha$'s in a certain range, 
we rescale the value of $\mathbf p_i$, $\mathbf c_j^{(i)}$ and $\mathbf g_i$ to lie in the interval $[0,1]$
before computing the sampling probability $\sigma(\mathbf a_i;\mathbf a)$. 
Specifically,
given the vector $\mathbf a$ and
let $q^{\text{low}}$ and $q^{\text{high}}$ be the quantiles of $\mathbf a$, 
we choose $q^{\text{low}}$ to be $20^{th}$ and $q^{\text{high}}$ to be $80^{th}$ percentiles, respectively. 
Then the rescaling function $r(\mathbf a_i)$ is formed as:
\begin{equation}
r(\mathbf a_i) = 
\begin{cases}
1 & \mathbf a_i>q^{\text{high}} 
\\
0 & \mathbf a_i<q^{\text{low}}
\\
\frac{\mathbf a_i-q^{\text{low}}}{q^{\text{high}} - q^{\text{low}}} & \text{otherwise}
\end{cases}.
\end{equation}
The rescaling function can also help us to avoid the case that some samples have extremely large score.
In this case, these samples will be selected for too many times.

\subsection{Experiment Setup}
\label{secc:exp-setup}

Five datasets are used here,
i.e., WN18, FB15K and their variants WN18RR, FB15K237, and YAGO3-10.
WN18 and FB15K are firstly introduced in \cite{bordes2013translating}.
They are widely tested in the literature
\cite{bordes2013translating,ji2015knowledge,trouillon2016complex,wang2018incorporating,cai2018kbgan,Zhang2018,kazemi2018simple}. 
WN18RR  \cite{dettmers2018convolutional} and FB15K237 \cite{toutanova2015observed} are variants that remove near-duplicate or inverse-duplicate relations from WN18 and FB15K.
The two variants are harder and more realistic.
YAGO3-10 is much larger than the others and is a subset of YAGO \cite{suchanek2007yago}.
Their statistics are shown in Table \ref{tb-dataset}.

\begin{table}[ht]
	\centering
	\caption{Detailed information of the datasets used in KG embedding experiments.}
	\label{tb-dataset}
	\renewcommand{\arraystretch}{1.2}
	\setlength\tabcolsep{3pt}
	\begin{tabular}{c|ccccc}
		\hline
		Dataset  & \#entity & \#relation &  \#train  & \#valid & \#test \\ \hline
		  WN18   &  40,943  &     18     &  141,442  &  5,000  & 5,000  \\
		 WN18RR  &  40,943  &     11     &  86,835   &  3,034  & 3,134  \\
		 FB15K   &  14,951  &   1,345    &  484,142  & 50,000  & 59,071 \\
		FB15K237 &  14,541  &    237     &  272,115  & 17,535  & 20,466 \\
		YAGO3-10 & 123,188  &     37     & 1,079,040 &  5,000  & 5,000  \\ \hline
	\end{tabular}
\end{table}

Following previous KG embedding works \cite{bordes2013translating,wang2014knowledge,ji2015knowledge,trouillon2016complex}
and the GAN-based works \cite{wang2018incorporating,cai2018kbgan}, 
we mainly test the performance on \emph{link prediction} task. 
This is also the testbed to measure KG embedding models.
Link prediction aims to predict the missing entity $h$ or $t$ for a positive triplet $(h, r, t)$. 
In this task, we measure the rank of head entity $h$ and tail entity $t$ among all the entities in $\mathcal E$.
Thus, link prediction emphasizes the rank of the correct entities
rather than their concrete scores.
Besides, to further verify the quality of the learned embedding,
we test the learned embeddings on \textit{triplet classification} task.
This task is to confirm whether a given triplet $(h,r,t)$ is correct or not, 
i.e., binary classification of triplet \cite{wang2014knowledge}.
In practice, it can help us to quickly answer the truth-or-false questions.

As in previous works \cite{bordes2013translating,trouillon2016complex,wang2018incorporating,cai2018kbgan,kazemi2018simple}, 
we evaluate the link prediction performance based on the following two metrics
\footnote{The mean rank (MR) metric, which is given in the conference version \cite{Zhang2018}, is removed in the journal version since 
	(i) space is limited. and (ii) the mean rank is easily influenced by low ranking samples.}:
\begin{itemize}
	\item Mean reciprocal ranking (MRR):
	It is computed by the average of the reciprocal ranks $1/|\mathcal S|\sum_{i=1}^{|\mathcal S|}\frac{1}{\text{rank}_i}$
	where $\text{rank}_i, i\in \{ 1,\dots, |\mathcal S| \} $ is a set of ranking results;
	\item Hit@10: 
	The percentage of appearance in the top-$10$ ranking: $1/|\mathcal S| \sum_{i=1}^{|\mathcal S|}\mathbb I(\text{rank}_i\!\leq\!\!10)$, where $\mathbb I(\cdot)$ is the indicator function;
\end{itemize}

MRR and Hit@$10$ measure the top rankings of positive entity in different levels. 
Hit@10 cares about general top rankings while the top~1 samples contribute most to MRR.
The larger value of MRR and Hit@$10$ indicates better performance.
To avoid underestimating the performance of different models, 
we report the performance in a ``filtered'' setting, 
i.e., all the corrupted triplets that exist in train, valid and test set are filtered out \cite{wang2018incorporating,cai2018kbgan}. 
A large amount of scoring functions have been proposed in the literature, 
please see a recent survey \cite{wang2017knowledge} for a review.
In this part,
following \cite{cai2018kbgan,wang2018incorporating},
TransE \cite{bordes2013translating}, 
TransH \cite{wang2014knowledge}, 
TransD \cite{ji2015knowledge}, 
DistMult \cite{yang2014embedding} 
and 
ComplEx \cite{trouillon2016complex} 
will be used as scoring functions for comparison;
besides,
the recently developed scoring function
SimplE \cite{kazemi2018simple}
is also included (see Table~\ref{tb-score-func}).

\subsection{Comparison with State-of-the-arts}
\label{ssec:compstate}

In this section,
we focus on the comparison 
between our proposed cache-based negative sampling
with the other baseline sampling methods.

\begin{table*}[ht]
	\centering
	\caption{Comparison of various algorithms on the five data sets. 
		Performance of the pretrained model is included as reference.
		As code of IGAN is not available,                
		its performance is directly copied from \cite{wang2018incorporating}.
		Note that MRR, and those on WN18RR, FB15K237, YAGO3-10 data sets are not reported as they are not shown in IGAN.
		Besides, KBGAN on YAGO3-10 is not reported since it easily runs out of memory.
		Bold number means the best performance, and underline means the second best under each setting.
	}
	\label{tb-perf}
	\renewcommand{\arraystretch}{1}
		\begin{tabular}{c|cc|cc|cc|cc|cc|cc}
			\hline
			scoring          &  \multicolumn{2}{c|}{Dataset}   &       \multicolumn{2}{c|}{WN18}        &      \multicolumn{2}{c|}{WN18RR}       &       \multicolumn{2}{c|}{FB15K}       &     \multicolumn{2}{c|}{FB15K237}      &      \multicolumn{2}{c}{YAGO3-10}      \\ \cline{2-13}
			functions         &  \multicolumn{2}{c|}{Metrics}   &        MRR         &      $\!\!$Hit@10$\!\!$       &        MRR         &      $\!\!$Hit@10$\!\!$       &        MRR         &      $\!\!$Hit@10$\!\!$       &        MRR         &      $\!\!$Hit@10$\!\!$       &        MRR         &      $\!\!$Hit@10$\!\!$       \\ \hline
			\multirow{8}{*}{TransE}  & \multicolumn{2}{c|}{pretrained} &       0.4213       &       91.50       &       0.1753       &       44.48       &       0.4679       &       74.70       &       0.2262       &       38.64       &       0.1723       &       32.24       \\
			& \multicolumn{2}{c|}{Bernoulli}  &       0.5001       &       94.13       &       0.1784       &       45.09       &       0.4951       &      {77.37}      &       0.2556       &       41.89       &       0.2053       &       37.87       \\
			&   KBGAN   & pretrain            &       0.6880       &  \textbf{94.92}   &       0.1864       &       45.39       &       0.4858       &       77.02       &       0.2938       &       46.69       &       ------       &      ------       \\
			&           & scratch             &       0.6606       &       94.80       &       0.1808       &       43.24       &       0.3771       &       72.67       &       0.2926       &       46.59       &       ------       &      ------       \\
			& NSCaching & pretrain            &  \textbf{0.7867}   &  \textbf{94.92}   &  \textbf{0.2048}   & \underline{47.38} &  \textbf{0.6475}   &  \textbf{81.54}   &  \textbf{0.3004}   & \underline{47.36} & \underline{0.3065} &  \textbf{51.36}   \\
			&           & scratch             & \underline{0.7818} &       94.63       & \underline{0.2002} &  \textbf{47.83}   & \underline{0.6391} & \underline{80.95} & \underline{0.2993} &  \textbf{47.64}   &  \textbf{0.3074}   & \underline{50.65} \\
			&   IGAN*   & pretrain            &       ------       &       91.3        &       ------       &      ------       &       ------       &       74.0        &       ------       &      ------       &       ------       &      ------       \\
			&           & scratch             &       ------       &       92.7        &       ------       &      ------       &       ------       &       73.1        &       ------       &      ------       &       ------       &      ------       \\ \hline
			\multirow{8}{*}{TransH}  & \multicolumn{2}{c|}{pretrained} &       0.4527       &       92.71       &       0.1755       &       43.30       &       0.4316       &       73.98       &       0.2222       &       38.80       &       0.1444       &       29.70       \\
			& \multicolumn{2}{c|}{Bernoulli}  &       0.5206       &       94.52       &       0.1862       &       45.09       &       0.4518       &       76.55       &       0.2329       &       40.10       &       0.1783       &       35.35       \\
			&   KBGAN   & pretrain            &       0.6168       &       94.84       &       0.1923       &       45.31       &       0.4262       &       75.91       &       0.2807       &      {46.39}      &       ------       &      ------       \\
			&           & scratch             &       0.6018       &       94.60       &       0.1869       &       44.81       &       0.3364       &       72.53       &       0.2779       &       46.19       &       ------       &      ------       \\
			& NSCaching & pretrain            &  \textbf{0.8063}   &  \textbf{95.32}   & \underline{0.2038} &  \textbf{48.04}   & {\textbf{0.6520}}  &  \textbf{81.96}   & \underline{0.2812} & \underline{46.48} & \underline{0.2988} & \underline{50.82} \\
			&           & scratch             & \underline{0.8038} & \underline{95.29} &  \textbf{0.2041}   &  \textbf{48.04}   & \underline{0.6391} & \underline{81.05} &  \textbf{0.2832}   &  \textbf{46.59}   &  \textbf{0.3013}   &  \textbf{51.07}   \\
			&   IGAN*   & pretrain            &       ------       &       94.0        &       ------       &      ------       &       ------       &       77.0        &       ------       &      ------       &       ------       &      ------       \\
			&           & scratch             &       ------       &       86.9        &       ------       &      ------       &       ------       &       73.3        &       ------       &      ------       &       ------       &      ------       \\ \hline
			\multirow{8}{*}{TransD}  & \multicolumn{2}{c|}{pretrained} &       0.4426       &       92.69       &       0.1782       &       42.18       &       0.4320       &       73.98       &       0.2244       &       39.53       &       0.1571       &       31.95       \\
			& \multicolumn{2}{c|}{Bernoulli}  &       0.5093       &       94.61       &       0.1901       &       46.41       &       0.4529       &       76.55       &       0.2451       &       42.89       &       0.2014       &       38.61       \\
			&   KBGAN   & pretrain            &       0.6130       &       94.92       &       0.1917       &       46.49       &       0.4069       &       74.27       &       0.2487       &       44.33       &       ------       &      ------       \\
			&           & scratch             &       0.5950       &       94.68       &       0.1875       &       46.41       &       0.3151       &       69.77       &       0.2465       &       44.40       &       ------       &      ------       \\
			& NSCaching & pretrain            &  \textbf{0.8022}   & \underline{94.99} &  \textbf{0.2013}   & \underline{48.36} &  \textbf{0.6567}   &  \textbf{82.02}   &  \textbf{0.2883}   &  \textbf{48.33}   &  \textbf{0.3180}   &  \textbf{53.05}   \\
			&           & scratch             & \underline{0.7994} &  \textbf{95.16}   &  \textbf{0.2013}   &  \textbf{48.39}   & \underline{0.6415} & \underline{81.32} & \underline{0.2863} & \underline{47.85} & \underline{0.3146} & \underline{52.65} \\
			&   IGAN*   & pretrain            &       ------       &       93.3        &       ------       &      ------       &       ------       &      {77.6}       &       ------       &      ------       &       ------       &      ------       \\
			&           & scratch             &       ------       &       93.0        &       ------       &      ------       &       ------       &       74.0        &       ------       &      ------       &       ------       &      ------       \\ \hline\hline
			\multirow{6}{*}{DistMult} & \multicolumn{2}{c|}{pretrained} &       0.6340       &       92.28       &       0.3765       &       44.85       &       0.4985       &       78.28       &       0.2247       &       36.03       &       0.2805       &       48.81       \\
			& \multicolumn{2}{c|}{Bernoulli}  &       0.7918       &  {93.38}   &       0.3964       &       45.25       &       0.5376       &      {78.69}      &       0.2491       &       42.03       &       ------       &      ------       \\
			&   KBGAN   & pretrain            &       0.6955       &       93.11       &       0.3849       &       44.32       &       0.5568       &       75.57       &       0.2670       &       45.34       &       0.3262       &       54.58       \\
			&           & scratch             &       0.7275       &       93.08       &       0.2039       &       29.52       &       0.4227       &       64.35       &       0.2272       &       39.91       &       ------       &      ------       \\
			& NSCaching & pretrain            & \underline{0.8297} & \textbf{93.83} &  \textbf{0.4148}   &  \textbf{45.80}   & \underline{0.7447} &  \textbf{84.16}   &  \textbf{0.2882}   &  \textbf{45.79}   &       \textbf{0.4112}      &  \textbf{57.24}   \\
			&           & scratch             &  \textbf{0.8306}   &       \underline{93.74}       & \underline{0.4128} & \underline{45.45} & {\textbf{0.7448}}  & \underline{83.91} & \underline{0.2834} & \underline{45.56} &       \underline{0.4032}       & \underline{56.58} \\ \hline
			\multirow{6}{*}{ComplEx}  & \multicolumn{2}{c|}{pretrained} &       0.8046       &       93.75       &       0.3934       &       41.63       &       0.5191       &       78.02       &       0.2201       &       35.55       &       0.3008       &       49.94       \\
			& \multicolumn{2}{c|}{Bernoulli}  &       0.9115       &  \textbf{94.39}   &       0.4431       &  \textbf{51.77}   &       0.6253       &      {80.72}      &       0.2596       &       43.54       &       0.3568       &       54.67       \\
			&   KBGAN   & pretrain            &       0.8976       &       93.73       &       0.4287       &       47.03       &       0.6254       &       80.95       &       0.2818       &       45.37       &       ------       &      ------       \\
			&           & scratch             &       0.7233       &       85.81       &       0.3180       &       35.51       &       0.5002       &       76.10       &       0.1910       &       32.07       &       ------       &      ------       \\
			& NSCaching & pretrain            & \underline{0.9326} & \underline{94.03} &  \textbf{0.4487}   & \underline{51.76} & \underline{0.7994} &  \textbf{86.32}   & \underline{0.3017} & \underline{47.75} & \underline{0.4020} & \underline{56.50} \\
			&           & scratch             &  \textbf{0.9355}   &       93.98       &       0.4463       &       50.89       &  \textbf{0.7995}   & \underline{86.28} &  \textbf{0.3021}   &  \textbf{48.05}   &  \textbf{0.4045}   &  \textbf{57.79}   \\ \hline
			\multirow{7}{*}{SimplE}  & \multicolumn{2}{c|}{pretrained} &       0.9056       &       94.34       &       0.3989       &       46.09       &       0.5684       &       79.62       &       0.2355       &       35.71       &       0.3405       &       55.46       \\
			& \multicolumn{2}{c|}{Bernoulli}  &       0.9300       &       94.40       &       0.4256       &       46.90       &       0.6704       &       81.47       &       0.2388       &       36.80       &       0.3614       &       56.90       \\
			&   KBGAN   & pretrain            &       0.9335       &       94.69       &       0.4331       &       47.64       &       0.7692       &       85.76       &       0.2454       &       40.32       &       ------       &      ------       \\
			&           & scratch             &       0.9258       & \underline{94.76} &       0.4126       &       46.43       &       0.4737       &       66.44       &       0.2278       &       35.94       &       ------       &      ------       \\
			& NSCaching & pretrain            &       0.9412       &       94.71       &       0.4352       & \underline{48.01} &  \underline{0.8033}   & \underline{86.90} &       0.2711       & \underline{43.88} & \underline{0.4211} & \underline{59.77} \\
			&           & scratch             & \underline{0.9415} & \underline{94.76} & \underline{0.4361} &       47.67       & {0.8026} &       86.86       & \underline{0.2718} & \underline{43.88} &       0.4193       &       56.47       \\ \cline{3-13}
			&           & auto                &  \textbf{0.9446}   &  \textbf{94.98}   &  \textbf{0.4388}   &  \textbf{48.36}   & \textbf{0.8148} &  \textbf{88.47}   &  \textbf{0.2966}   &  \textbf{46.22}   &  \textbf{0.4532}   &  \textbf{61.84}   \\ \hline
		\end{tabular}
\end{table*}

\subsubsection{Compared Methods}
\label{sssec:methods}

Following methods for negative sampling in KG embedding are compared:
\begin{itemize}
	\item \textit{Bernoulli} \cite{wang2014knowledge}:
	As an extension of the uniform sampling scheme used in TransE, 
	Bernoulli sampling controls the probability for sampling $(\bar{h}, r, t)$  or $(h, r, \bar{t})$
	in the  one-to-many, many-to-one and many-to-many relations. 
	Specifically, it samples $(\bar{h}, r, t)$ or $(h, r, \bar{t})$ under a fixed Bernoulli distribution for each $r$.
	
	\item \textit{KBGAN} \cite{cai2018kbgan}\footnote{\url{https://github.com/cai-lw/KBGAN}}:
	This method firstly samples a set $\mathcal{N}$ uniformly from the whole entity set $\mathcal E$. 
	Then the head or tail entity is replaced with the entities in  $\mathcal{N}$ to form a set of candidate $(\bar{h}, r, t)$  and $(h, r, \bar{t})$. 
	The generator in KBGAN tries to pick up one triplet among them. As proposed in \cite{cai2018kbgan}, we choose the simplest model TransE as the generator. 
	For a fair comparison, the size of set $\mathcal{N}$ is the same as our cache size $N_1$.
	We use the published code and change the configure same as ours in the comparison.		
		\textit{Self-Adv} \cite{sun2018rotate} works similarly as KBGAN.
		The main difference is that Self-Adv
		uses the target embedding model itself as the generator.
	
	\item \textit{NSCaching}
	(Algorithm~\ref{alg-cache}):
	As in Section \ref{sec:proposed} and \ref{ssec:imp-details}, 
	the negative triplets are formed by replacing the head entity $h$ or tail entity $t$ with the one sampled from
	head-cache $\mathcal{H}$ or tail-cache $\mathcal{T}$.
	The cache is updated as in Algorithm \ref{alg-cache-update}. 
	Note that we can also lazily update the cache several iterations later to save time.
	We use $n=0$ without lazy-update unless otherwise specified.
	Besides, we use \textit{AutoML} to denote the improved version which tunes the hyper-parameters to balance E\&E.
\end{itemize}

As the source code of \textit{IGAN} \cite{wang2018incorporating} is not available, 
we do not compare with it here.
Instead,
we directly use the reported performance in the sequel.
Finally,
we also use Bernoulli sampling to choose between $(\bar{h}, r, t)$ and $(h, r, \bar{t})$ for \textit{KBGAN} and \textit{NSCaching}.
Besides,
as in \cite{cai2018kbgan,wang2018incorporating},
two strategies are used for \textit{KBGAN} and 
\textit{NSCaching}:
\begin{itemize}
	\item \textit{Scratch}: The embedding of relations and entities are initialized by the Xavier uniform initializer \cite{glorot2010understanding},
	and the models
	(denoted as \textit{KBGAN + scratch} and \textit{NSCaching + scratch})
	are directly applied to train the given KG;

	\item \textit{Pretrain}:
	Same as \cite{cai2018kbgan,wang2018incorporating},
	we firstly pretrain each scoring function with \textit{Bernoulli} sampling, 
	several epochs on the data sets. 
	We denote it as \textit{pretrained}. 
	Then the obtained parameters are used to warm-start the given KG.
	We keep training the warm-started KG embedding 
	and evaluate the performance under different sampling methods,
	i.e., \textit{Bernoulli}, \textit{KBGAN + pretrain} and \textit{NSCaching + pretrain}.
	Besides, the generator in \textit{KBGAN} is warm-started with corresponding TransE model.
\end{itemize}                               

\begin{figure*}[ht]
	\centering
	\subfigure[WN18.]
	{\includegraphics[width = 0.244\textwidth]{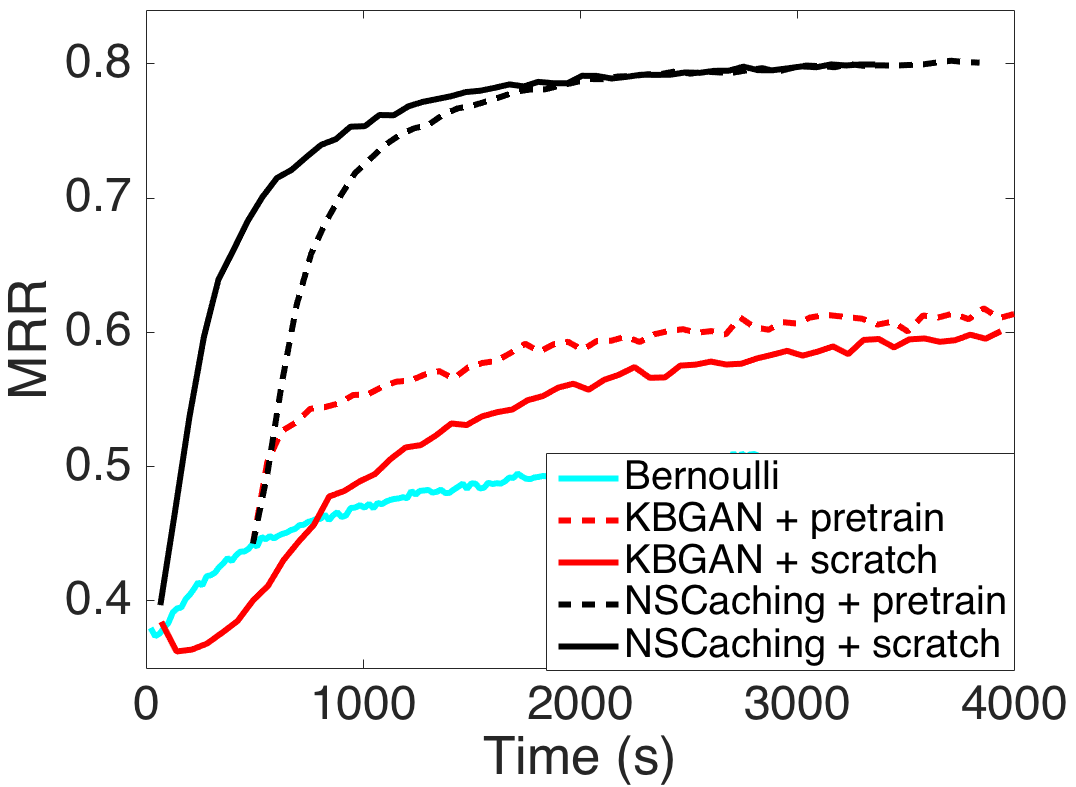}}
	\subfigure[WN18RR.]
	{\includegraphics[width = 0.244\textwidth]{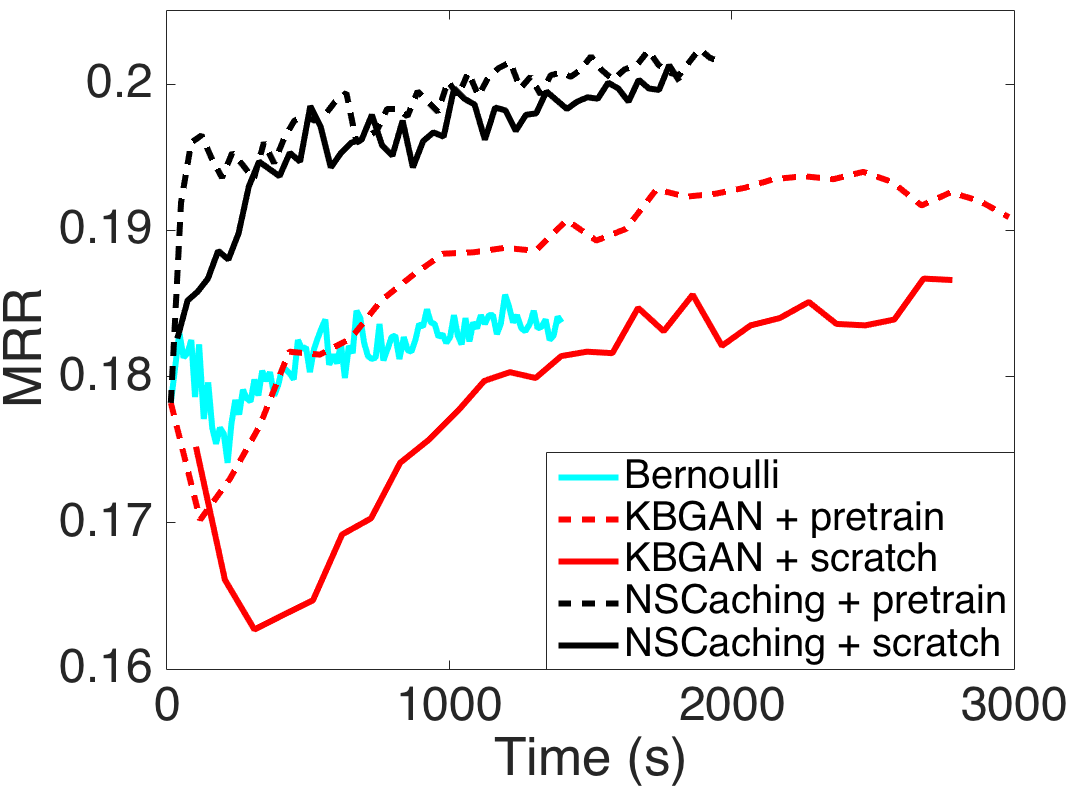}}
	\subfigure[FB15K.]
	{\includegraphics[width = 0.244\textwidth]{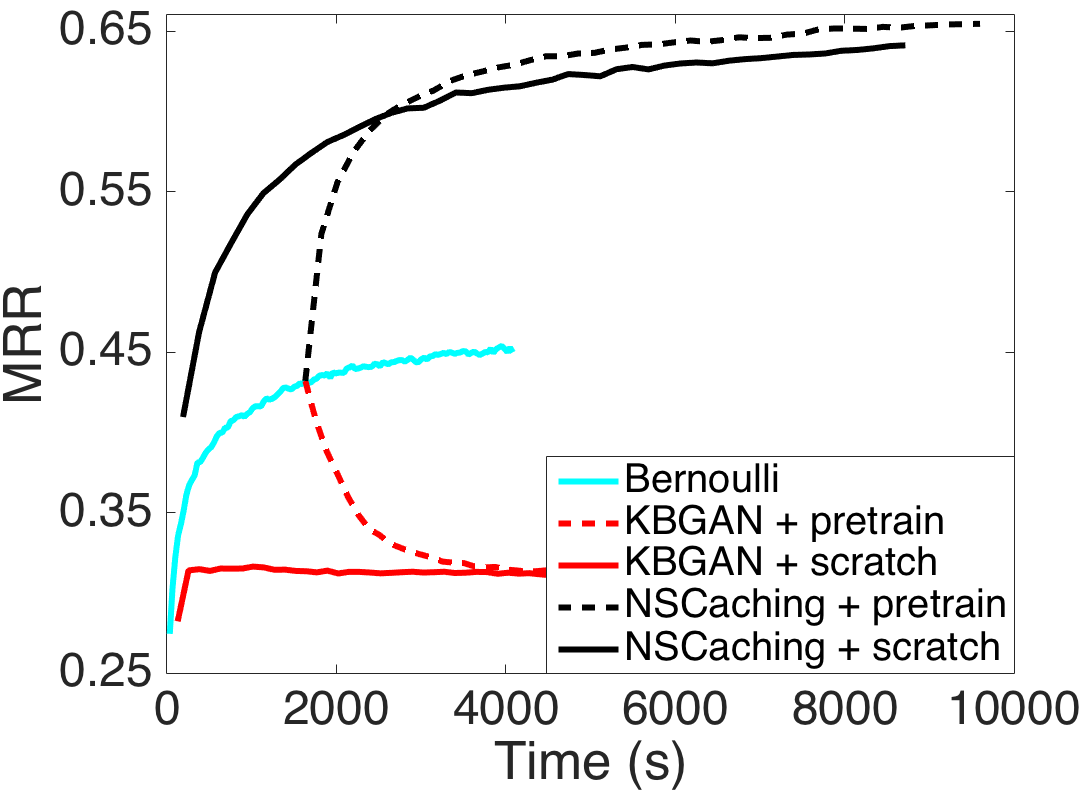}}
	\subfigure[FB15K237.]
	{\includegraphics[width = 0.244\textwidth]{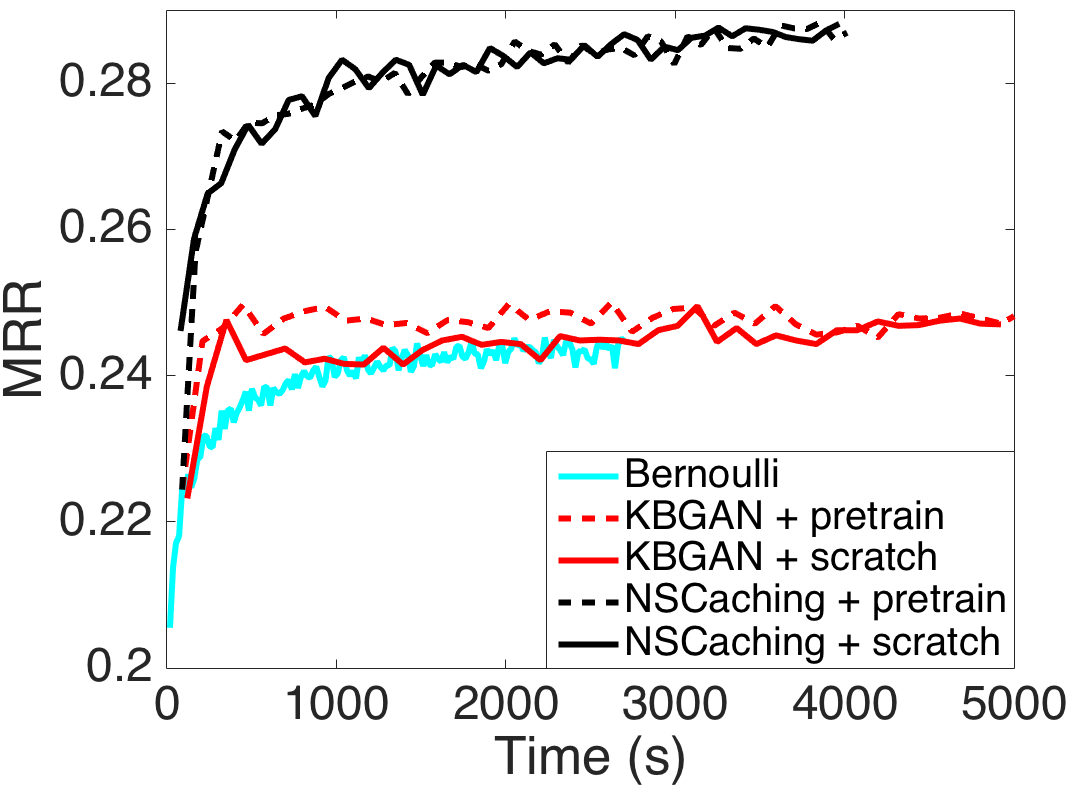}}
	
		\subfigure[WN18.]
	{\includegraphics[width = 0.244\textwidth]{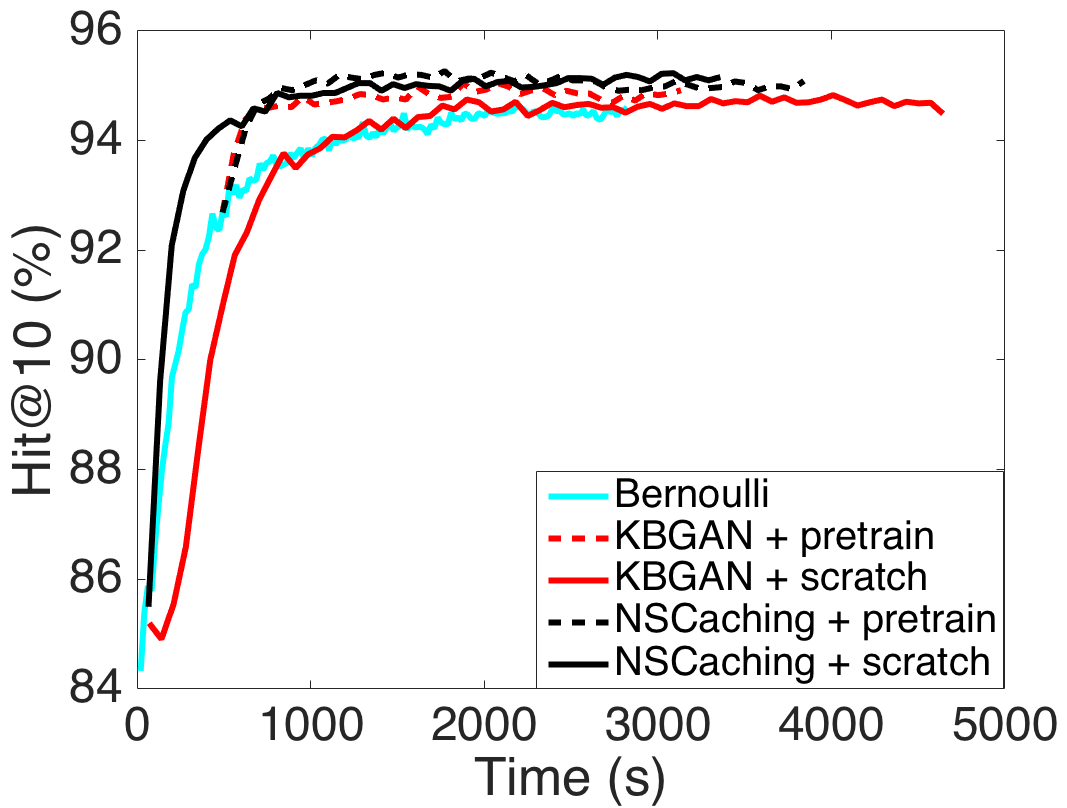}}
	\subfigure[WN18RR.]
	{\includegraphics[width = 0.244\textwidth]{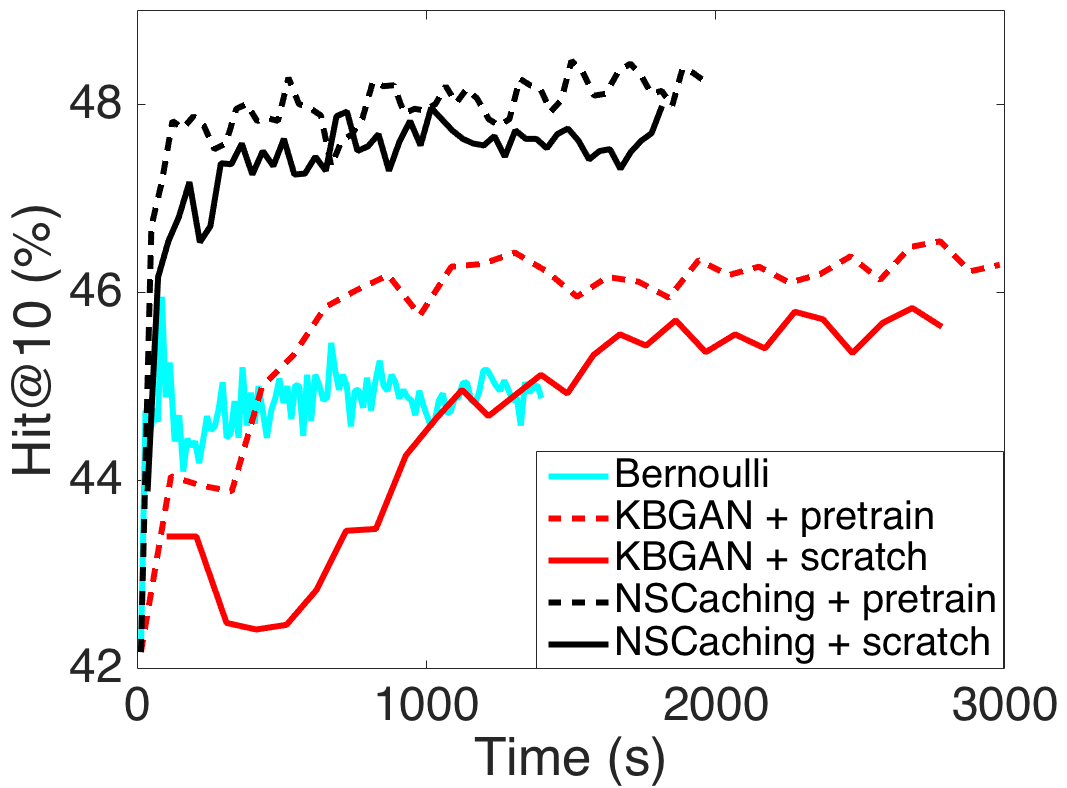}}
	\subfigure[FB15K.]
	{\includegraphics[width = 0.244\textwidth]{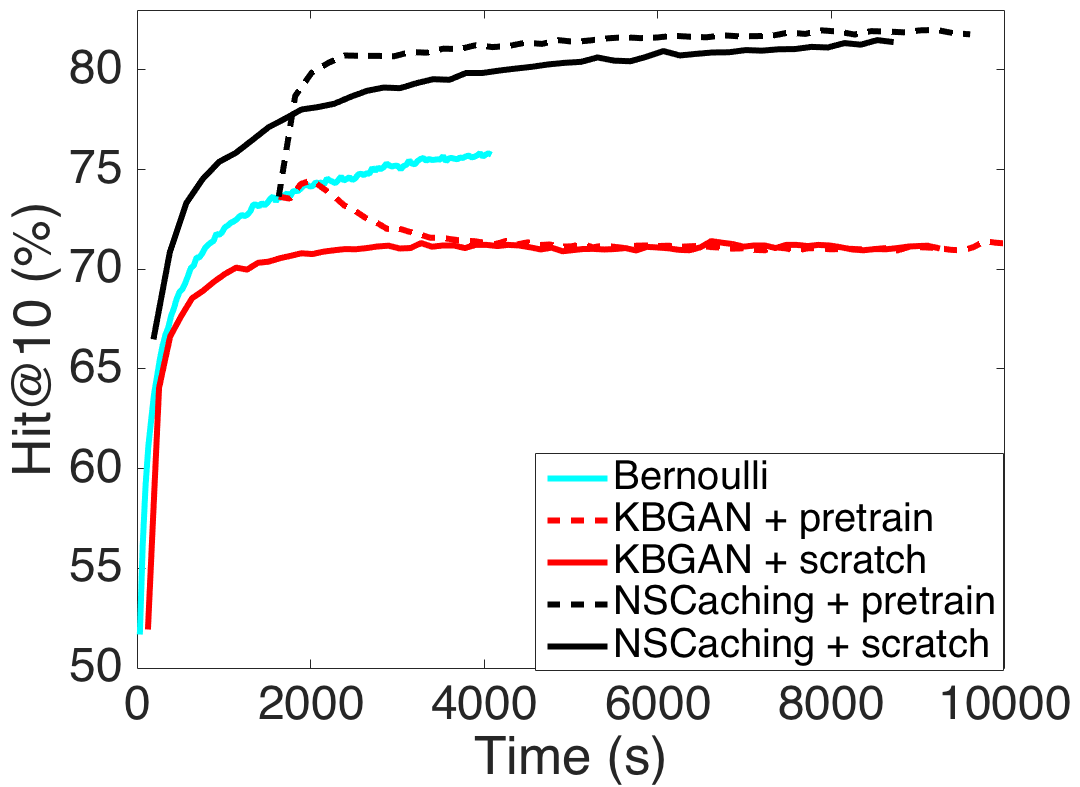}}
	\subfigure[FB15K237.]
	{\includegraphics[width = 0.244\textwidth]{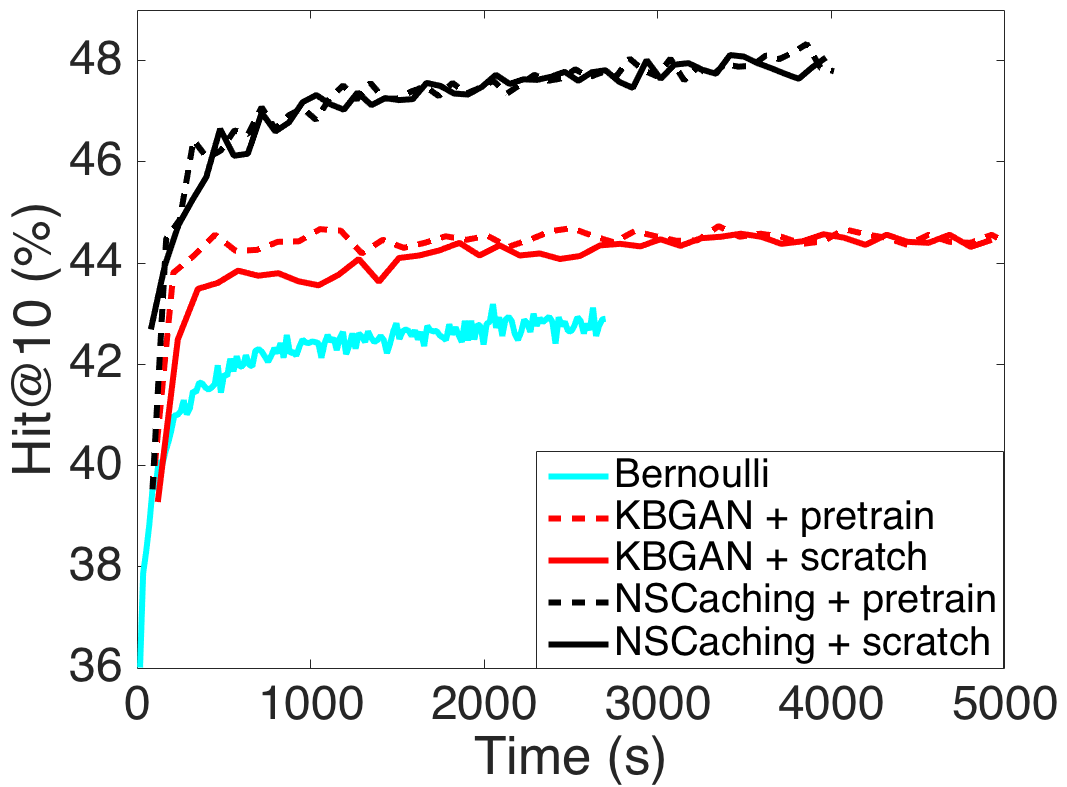}}
	
	\caption{Testing performance v.s. clock time (in seconds) based on TransD (best viewed in color).}
	\label{fig-perf-transd}
\end{figure*}

\begin{figure*}[ht]
	\centering
	\subfigure[WN18.]
	{\includegraphics[width = 0.244\textwidth]{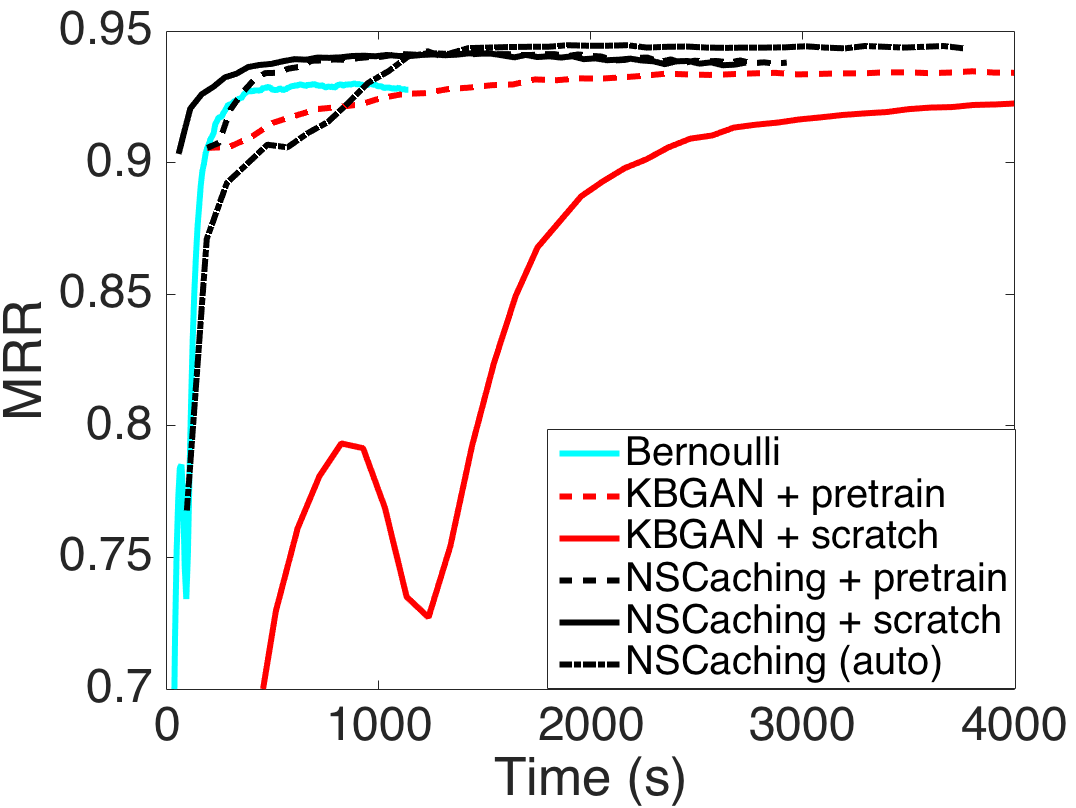}}
	\subfigure[WN18RR.]
	{\includegraphics[width = 0.244\textwidth]{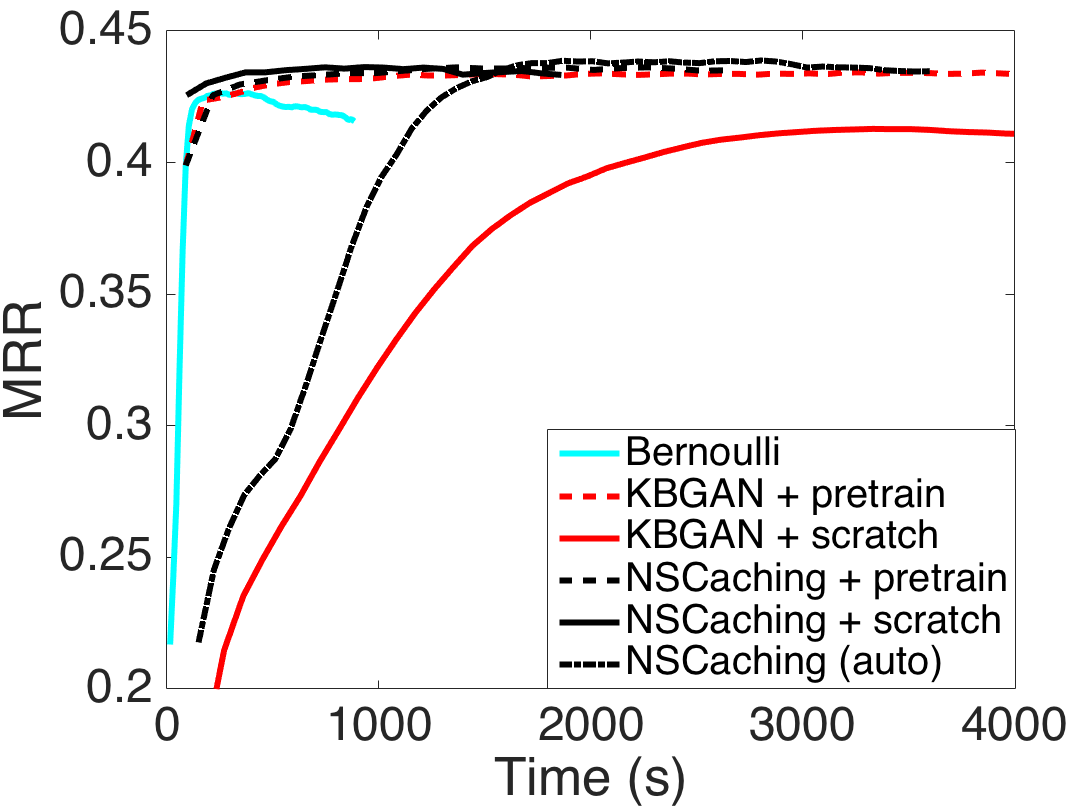}}
	\subfigure[FB15K.]
	{\includegraphics[width = 0.244\textwidth]{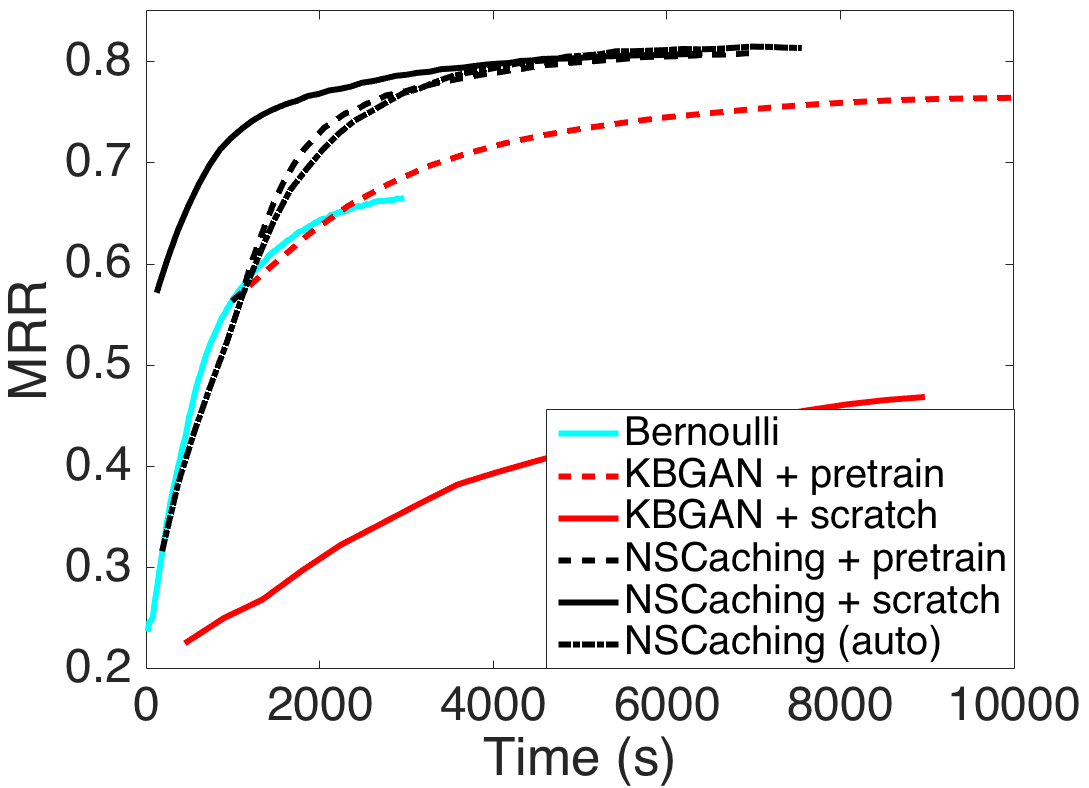}}
	\subfigure[FB15K237.]
	{\includegraphics[width = 0.244\textwidth]{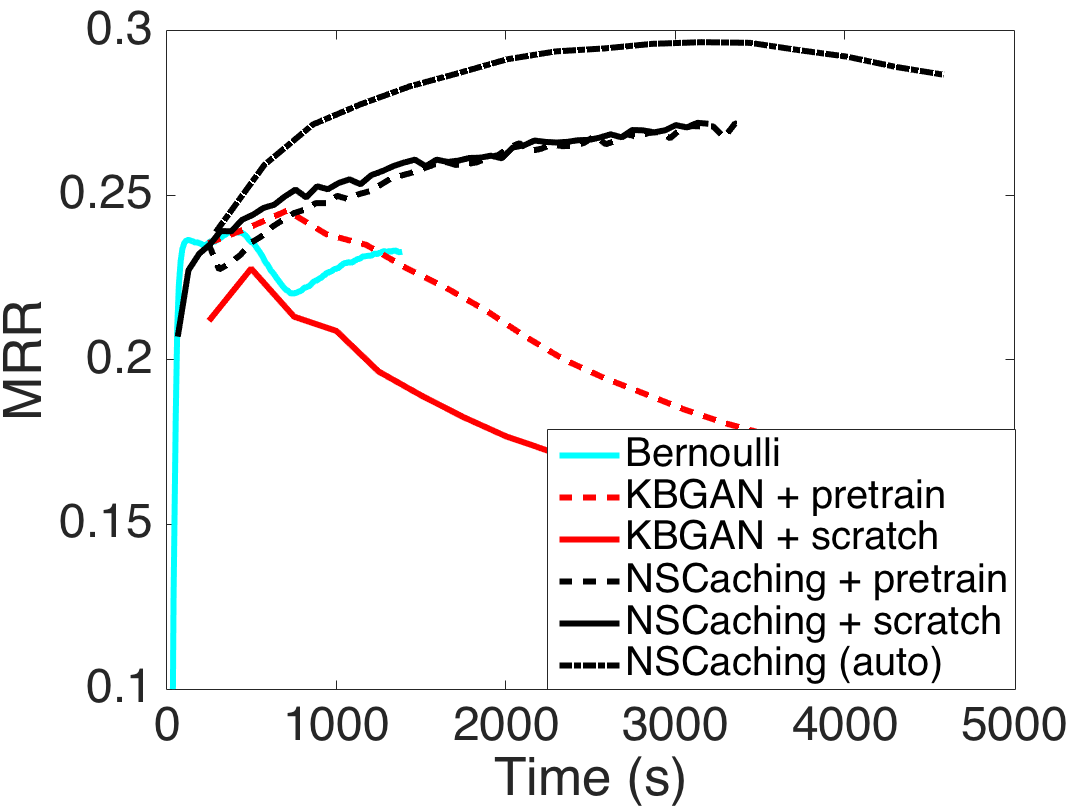}}
	
	\subfigure[WN18.]
	{\includegraphics[width = 0.244\textwidth]{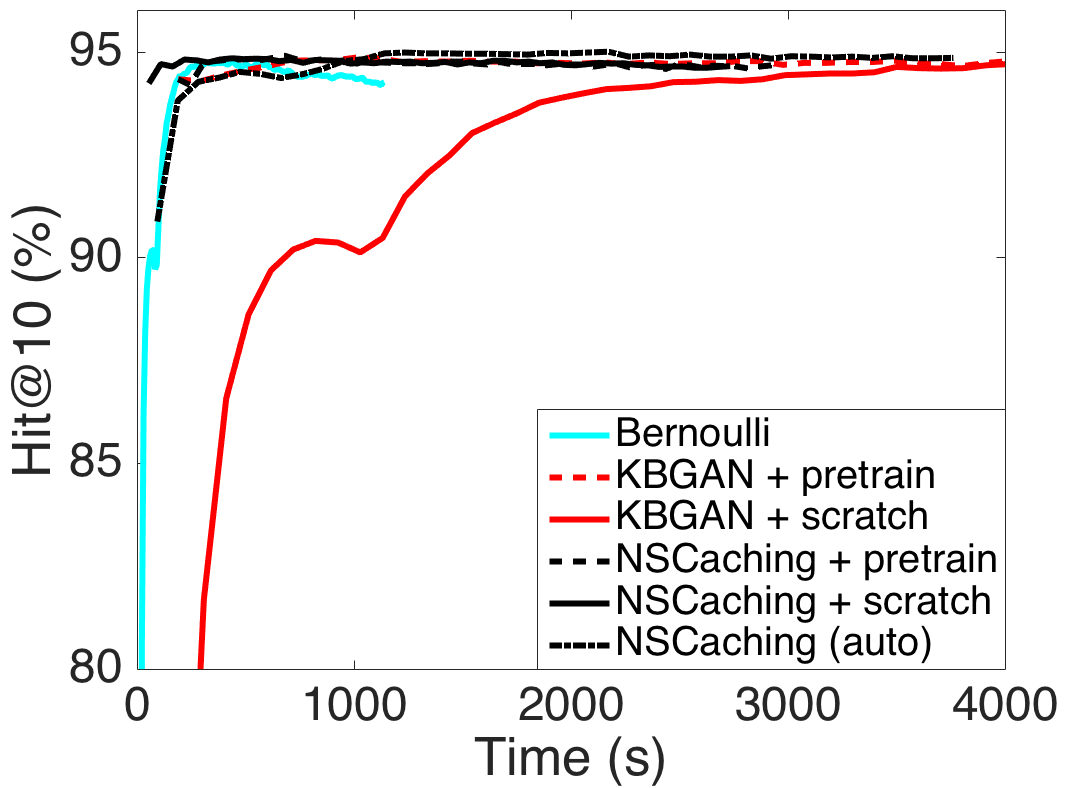}}
	\subfigure[WN18RR.]
	{\includegraphics[width = 0.244\textwidth]{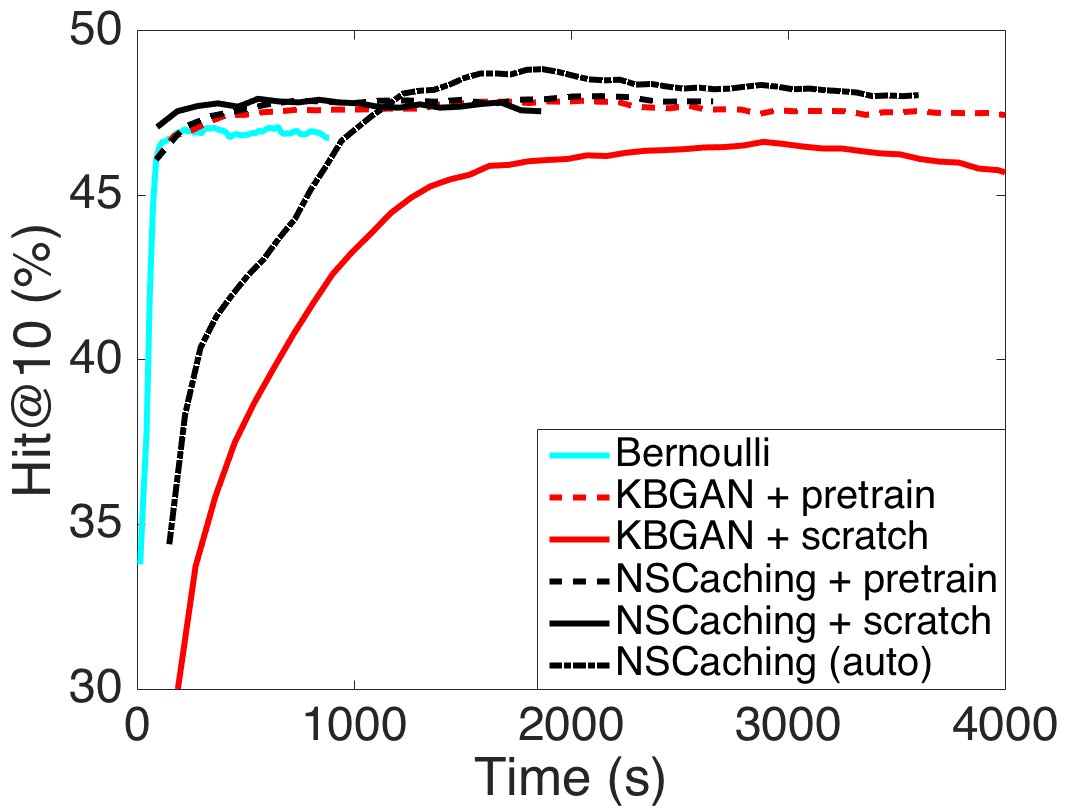}}
	\subfigure[FB15K.]
	{\includegraphics[width = 0.244\textwidth]{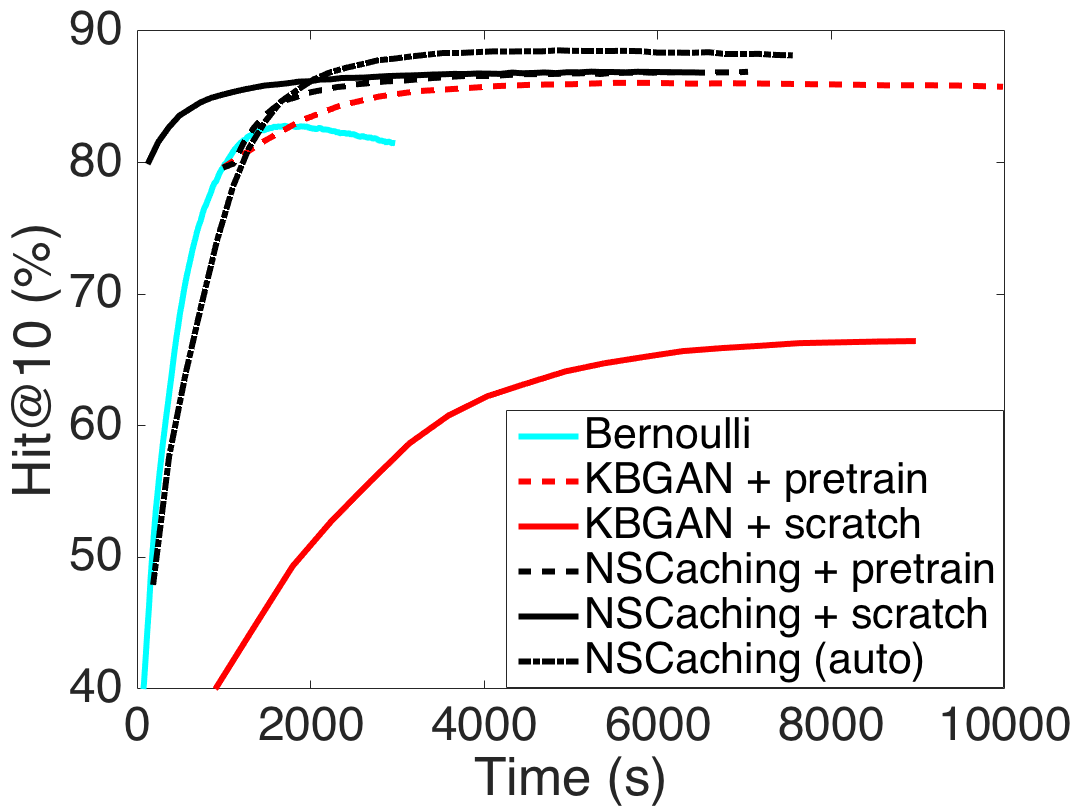}}
	\subfigure[FB15K237.]
	{\includegraphics[width = 0.244\textwidth]{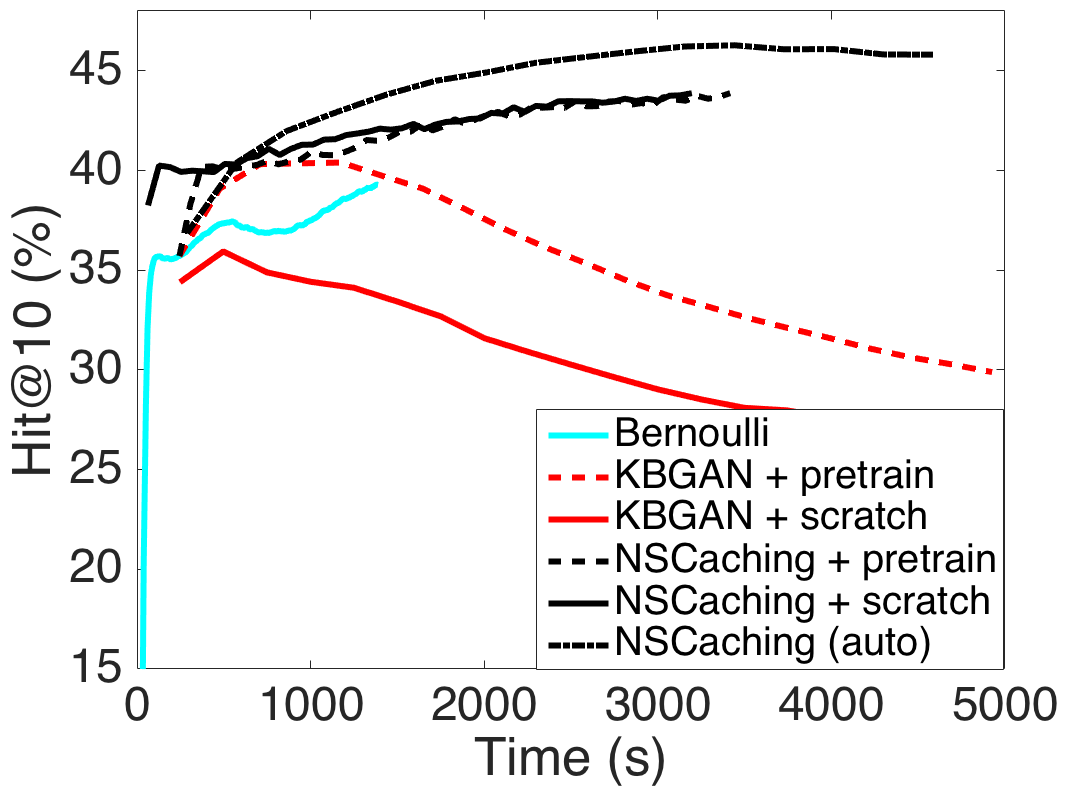}}
	\caption{Testing performance v.s. clock time (in seconds) based on SimplE (best viewed in color).}
	\label{fig-perf-complex}
\end{figure*}

Same as the KG embedding works in the literature \cite{kadlec2017knowledge,bordes2013translating,kazemi2018simple},
we use grid search to select the KG embedding related hyper-parameters:
hidden dimension $d\in \{50$, $100$, $200\}$,
batch size $m\in\{1024$, $2048$, $4096\}$,
learning rate $\eta \in$ $\{0.0001$, $0.0003$, $0.001$, $0.003$, $0.01$, $0.03$, $0.1\}$.
For translational distance models, we tune the margin value $\gamma \in \{1$, $2$, $3$, $4\}$.
And for semantic matching models, we tune the penalty value $\lambda\in$ $\{0.001$, $0.01$, $0.1 \}$ \cite{trouillon2016complex}.
We use Adam \cite{kingma2014adam}, 
which is a popular variant of SGD algorithm for the training,
and adopt its default settings except for the learning rate.
The best hyper-parameters are tuned under Bernoulli sampling scheme and evaluated by the MRR metric on the valid set. 
We keep them fixed for the baselines \textit{Bernoulli}, \textit{KBGAN} and \textit{NSCaching} here.
Following \cite{cai2018kbgan}, we save and record 
the \textit{pretrained} model after initial training epochs.
Then, \textit{Bernoulli} method keeps training until 3000 epochs; and 
the results of \textit{KBGAN} and \textit{NSCaching} algorithm are evaluated within 1000 epochs,
either from scratch or with pretraining.
All the recorded results are tested based on the best hyper-parameters chosen by the MRR value on valid set.
For cache related hyper-parameters, we choose $\alpha_1=\alpha_2=0, \alpha_3=1$ and $N_1=N_2=50$ 
for \textit{NSCaching}.

\subsubsection{Results on Translational Distance Models}
The performance on \textit{link prediction task} is compared in 
Table~\ref{tb-perf}.
First, we can see that,
for the translational distance models (TransE, TransH, TransD), 
\textit{KBGAN}, \textit{NSCaching} and \textit{IGAN} (both \textit{pretrain} and \textit{scratch}) gain significant improvements upon the baseline scheme \textit{Bernoulli}, 
especially for the performance gaining on the MRR metric,
which is mainly influenced by the top rankings.
This verifies the advantages of using large-gradient negative triplets during negative sampling and these methods can effectively pick up these negative triplets.

Then, \textit{IGAN} and \textit{KBGAN} with 
pretraining can perform better,
indicated by MRR and Hit@10,
than from scratch.
This shows that pretraining is helpful for the GAN-based methods.
In comparison, \textit{NSCaching} trained from either state (\textit{pretrain} or \textit{scratch})
can outperform \textit{IGAN} and \textit{KBGAN} on all the scoring functions.

The learning curve of the testing performance for various algorithms is shown in 
Figure~\ref{fig-perf-transd}. 
We use TransD here as the testbed.
As can be seen, 
all algorithms will converge to some points with stable testing performance, 
which empirically verifies the convergence of Adam optimizer \cite{kingma2014adam}.
Then, pretrain is a must for \textit{KBGAN} to achieve good performance.
When the generator is trained from scratch,
the whole model will suffer from instability,
especially at the initial training stages.
As a result,
it prevents the GAN-based models converging to some good local points.
\textit{NSCaching} can obtain good performance either from \textit{scratch} or with \textit{pretrain}.
Note that,
even through the updating and sampling scheme introduces extra training cost,
we can achieve better performance with fewer iterations.
As a result,
in all cases,
\textit{NSCaching} has better anytime performance  than both \textit{Bernoulli} and \textit{KBGAN}.

\subsubsection{Results on Semantic Matching Models}

The performance on semantic matching models is shown in the bottom rows of Table~\ref{tb-perf}. 
Same as that on translational distance models,
\textit{NSCaching} outperforms baseline scheme \textit{Bernoulli} significantly
as indicated by the bold and underline numbers. 
However, \textit{KBGAN} does not show a consistent performance.
In most settings, 
\textit{KBGAN from scratch} performs even worse than \textit{Bernoulli}.
This observation further verifies the fact that GAN-based methods usually suffer from instability and degeneracy.
They need careful balance between the generator and the discriminator, 
i.e., the target KG embedding model.
However, \textit{NSCaching} works consistently and performs the best 
both with pretrain or from scratch.

The learning curve of the testing performance for various algorithms 
is shown in Figure~\ref{fig-perf-complex}.
SimplE is used in this part.
As can be seen,
both \textit{Bernoulli} and \textit{NSCaching} will converge to some stable points.
In the contrast, 
\textit{KBGAN} will turn down and overfit on FB15K237 data set.
However,
\textit{NSCaching}, either with pretrain or from scratch, leads the performance and is well adopted on the semantic matching models without further tuning.
Besides, as for \textit{NSCaching (auto)},
we find that even though the sampling cost is higher,
the performance improvement is obvious and consistent on all these data sets.

\subsubsection{Results on Triplets Classification}

We do triplets classification in the same way as \cite{wang2014knowledge}.
This task is to confirm whether a given triplet $(h, r, t)$ is correct or not, i.e., do binary classification on the triplet. 
Compared with link prediction, triplets classification is more convenient in answering yes-or-no questions.
The decision rule of classification is learned as follows: 
for each $(h, r, t)$, if its score is no less than the relation-specific threshold $\sigma_r$, then we predict it to be positive. 
Otherwise, negative. 
The threshold $\sigma_r$ is determined by
maximizing the classification accuracy on the valid set.
We test this task on WN18RR and FB15K237 data sets based on TransD and SimplE.
As shown in Table~\ref{tb:classification}, \textit{NSCaching} still outperforms various baselines. 
This task further justifies that \textit{NSCaching} can help learn a better embedding of the KG.

\begin{table}[ht]
	\centering
	\caption{Comparison of various algorithms on triplet classification task.
		Bold number indicates the best performance.} 
	\label{tb:classification}
%	\begin{tabular}{C{30px}|cc|c|c}
\begin{tabular}{c|cc|c|c}
		\hline
		model     &  \multicolumn{2}{c|}{Dataset}  &     {WN18RR}      &    {FB15K237}     \\ \hline
		\multirow{5}{*}{TransD} & \multicolumn{2}{c|}{Bernoulli} &       86.81       &       78.24       \\
		&   KBGAN   &     pretrained     &       85.93       &       79.03       \\
		&           &      scratch       &       86.01       &       79.05       \\
		& NSCaching &     pretrained     &  \textbf{87.84}   & \underline{80.63} \\
		&           &      scratch       & \underline{87.64} &  \textbf{80.69}   \\ \hline
		\multirow{5}{*}{SimplE} & \multicolumn{2}{c|}{Bernoulli} &       84.48       &       77.64       \\
		&   KBGAN   &     pretrained     &       79.87       &       74.11       \\
		&           &      scratch       &       71.73       &       72.61       \\
		& NSCaching &     pretrained     &  \textbf{84.96}   & \underline{79.88} \\
		&           &      scratch       & \underline{84.83} &  \textbf{80.21}   \\ \hline
	\end{tabular}
\end{table}

\begin{figure*}[ht]
\centering
\subfigure[Impact of $\alpha_1$ (given $\alpha_3$).]
{\includegraphics[width=0.25\textwidth]{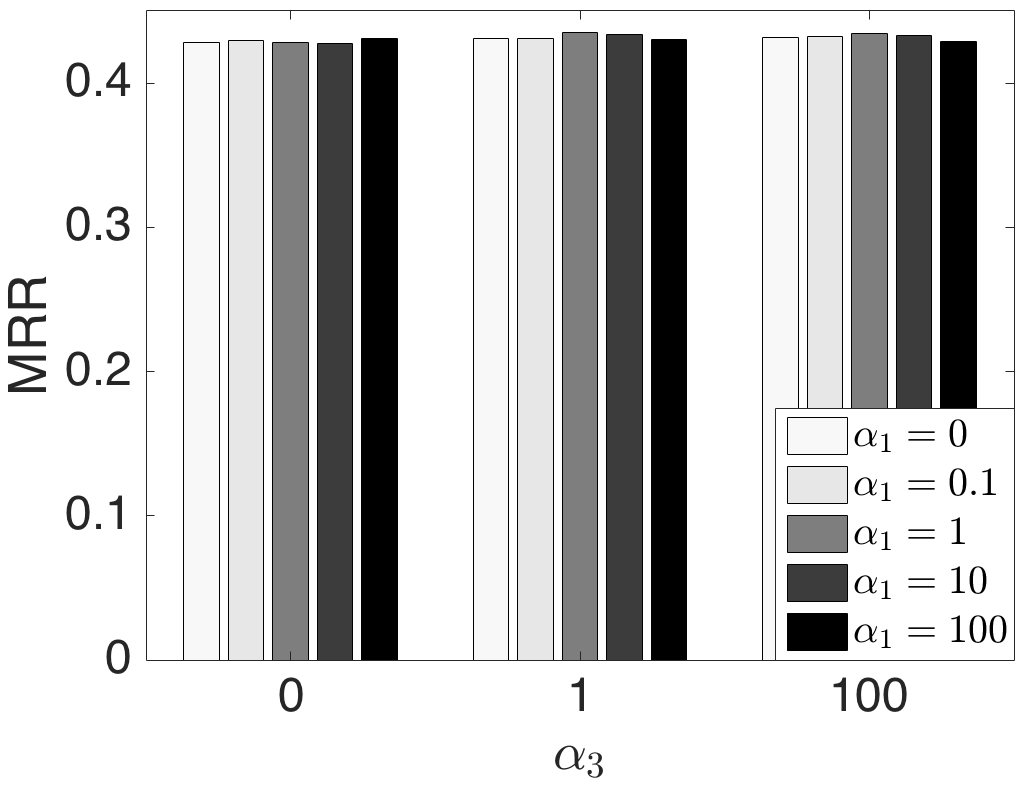}
	\label{fig:sample:alpha1}}\qquad
\subfigure[Impact of $\alpha_2$ (given $\alpha_3$).]
{\includegraphics[width=0.25\textwidth]{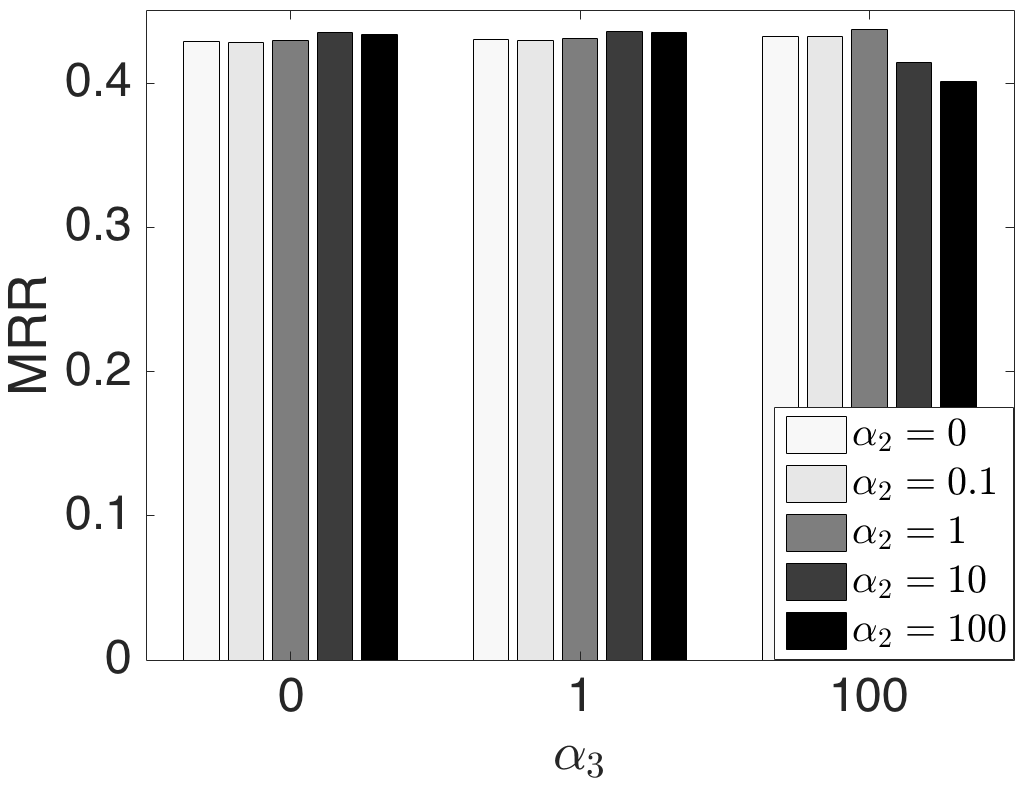}
	\label{fig:sample:alpha2}}\qquad
\subfigure[Impact of $\alpha_3$ (given $\alpha_2$).]
{\includegraphics[width=0.25\textwidth]{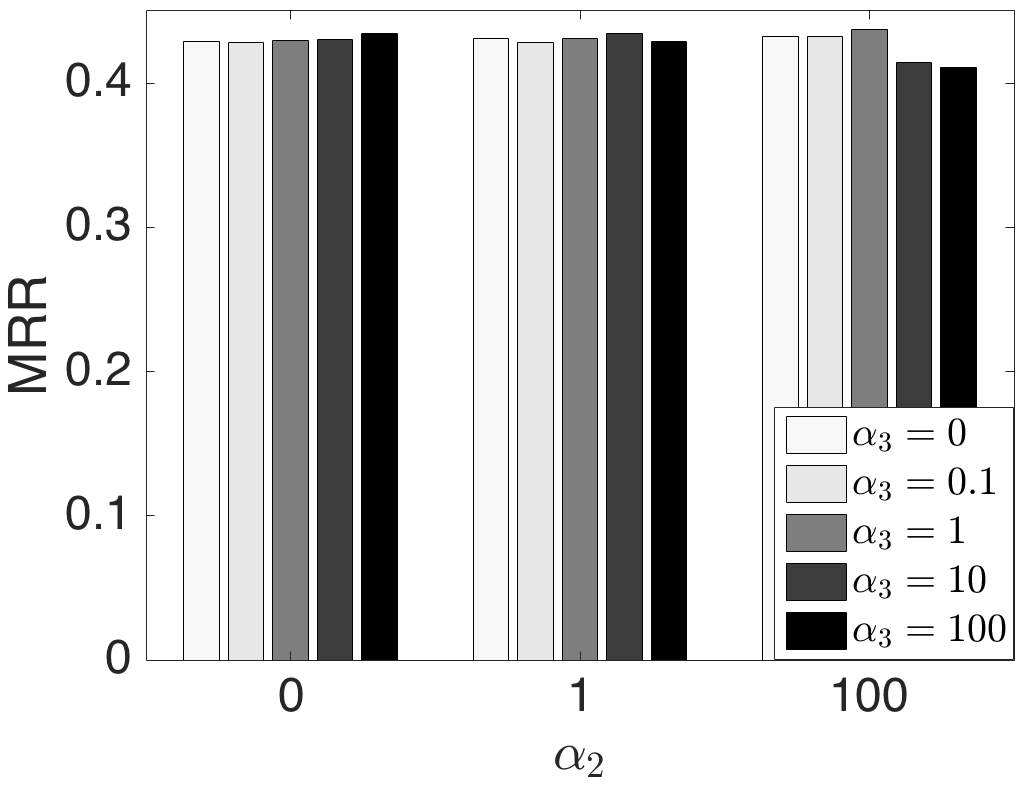}
	\label{fig:sample:alpha3}}
\caption{Balancing on exploration and exploitation with different values of $\alpha_1$, $\alpha_2$ and $\alpha_3$ 
	with SimplE on WN18RR.}
\end{figure*}

\subsection{Balancing E\&E}
\label{sec:ablation}

In this part,
we analyze the designing concerns on the hyper-parameters regarding ``exploration and exploitation''
 in Table~\ref{tab:eepara}.
SimplE and WN18RR are used as the scoring function and data set respectively.

\subsubsection{$\alpha_1$: Sampling positive triplet}
\label{ssec:exp:sample-cache}
Given a set of training triplets and the cache, how to sample the positive triplet is the first question we care about.
In Algorithm \ref{alg-cache-sample},
the most related hyper-parameters are $\alpha_1$,
which controls the distribution of positive samples in cache $\mathcal C$,
and $\alpha_3$,
which controls the content in the indexed cache $\mathcal C_i$.
The testing performance with different values of $\alpha_1$ is compared in Figure~\ref{fig:sample:alpha1}.
Since the content in cache is influenced by $\alpha_3$,
we use $0$ (low), $1$ (middle), and $100$ (high) as values of $\alpha_3$ for the testing.
As can be seen,
when $\alpha_3$ is small, 
different choices of $\alpha_1$ perform relatively bad 
and does not have regular influence on embedding performance.
As $\alpha_3$ becomes larger,
we see that a larger value of $\alpha_1$ performs better,
which verifies the better convergence property in Theorem~\ref{thm:conv}.
However, it will decrease with too large $\alpha_1$.
Take the distribution in Figure~\ref{fig:alphainf} as an example,
some positive triplets will not be selected when $\alpha_1$ is too large,
leading to a problematic training process.

\begin{table*}[ht]
	\centering
	\caption{Searching range of hyper-parameters and searched best value for different data sets.}
	\label{tab:search}
	\renewcommand{\arraystretch}{1.2}
	\begin{tabular}{c|c|c|c|c|c|c}
		\hline
		\multirow{2}{*}{hyper-parameters} & 
		\multirow{2}{*}{ranges} & \multicolumn{5}{c}{best searched} \\ \cline{3-7} 
		&  &{ WN18} & { WN18RR}& { FB15K} & { FB15K237} & { YAGO3-10} \\ \hline
		$\alpha_1$  &  [0, 1]     & 0.0668 &  0.013 & 0.0069 &  0  & 6.7e-4 \\ \hline
		$\alpha_2$	&   [0, 100]   & 3.463& 0  &0 &  3.122 & 16.62   \\ \hline
		$\alpha_3$	&   [0, 100]    & 24.25  & 2.759 & 1.848 & 2.579  &  0.2228 \\ \hline
		$N_1$       &    $\{10, 30, \dots, 90\}$   & 70 & 90 & 30 & 70 & 70 \\ \hline
		$N_2$       &    $\{10, 30, \dots, 90\}$  & 70  & 50  & 10 &  70 &  70   \\ \hline
	\end{tabular}
\end{table*}

\subsubsection{$\alpha_2$ and $\alpha_3$: Sampling and updating cache}
\label{ssec:exp:update-cache}

Once a positive triplet $(h_i, r_i, t_i)$ is sampled, we can get its corresponding cache $\mathcal C_i$ 
which stores the negative triplets.
The main parameters influencing the choice of negative triples are $\alpha_2$ and $\alpha_3$,
namely the temperature for sampling from cache and updating the cache.
To show how $\alpha_2$ and $\alpha_3$ balance E\&E, we fix $\alpha_3$ in certain ranges 
(low: $0$, middle: $1$ and high: $100$) and change $\alpha_2$ in Figure~\ref{fig:sample:alpha2},
and then do it alternatively in Figure~\ref{fig:sample:alpha3}.

From both Figure~\ref{fig:sample:alpha2} and \ref{fig:sample:alpha3}, 
we can see that balancing $\alpha_2$ and $\alpha_3$ are of vital importance.
When $\alpha_2$ or $\alpha_3$ has small value,
increasing the other one will improve the performance 
since exploitation is limited at this stage.
However,
when $\alpha_2$ or $\alpha_3$ becomes larger,
the other one should choose an appropriate value in order to avoid exploiting too much.
The performance goes up at initial stage and turns down as the break up of balance.
The reason is that
too large values of $\alpha_2$ and $\alpha_3$ will limit exploration
such that a few negative triplets will be selected too many times.
Besides,
false negative triplets will be more frequently sampled.
Fortunately, E\&E is well balanced  under a wide range of  $\alpha$'s values 
where
\textit{NSCaching} performs well without much effort in tuning $\alpha$'s.

Besides, when the cache is updated without referring to the gradient norms, i.e. $\alpha_3=0$ and $\alpha_2>0$, 
NSCaching works as an alternative version of KBGAN and outperforms Bernoulli baseline approach.
It works stabler than KBGAN which suffers from the instable training of the generator.
However,
the performance is still not the best 
since balance of E\&E is not in the best state when cache is updated without considering the negative triplets' qualities.

\subsubsection{$N_1$ and $N_2$: Cache size}
\label{ssec:exp:size}

\begin{figure}[ht]
	\centering
	\subfigure[Diff. $N_1$.]
	{\includegraphics[width = 0.49\columnwidth]{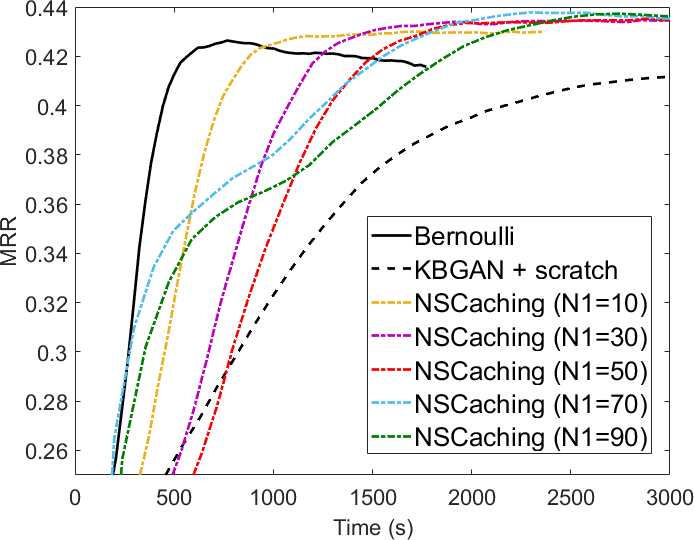}}
	\subfigure[Diff. $N_2$.]
	{\includegraphics[width = 0.49\columnwidth]{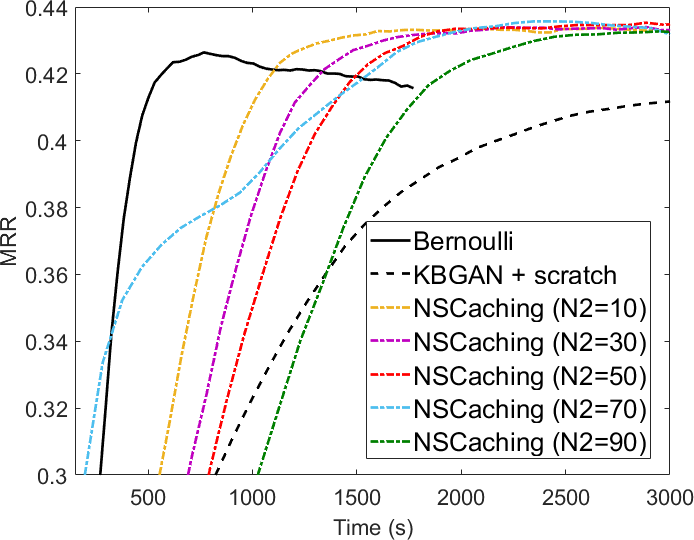}}
	
	\caption{Comparison of different $N_1$ when random subset size $N_2$ is fixed to 50, 
		and different $N_2$ when cache size $N_1$ is fixed to 50. Evaluated by SimplE model on WN18RR (best viewed in color).}
	\label{fig-CS-wn18-transd}
\end{figure}

Basically,
$N_1$ is the size of cache $\mathcal C_i$.
Then,
$N_2$ is the size of randomly sampled subset $\mathcal R_i$ of negative triplets from $\bar{\mathcal{S}}_i$,
which will later be  used to update the cache.
In this part,
we show their impact on the performance of \textit{NSCaching}.
The three temperature values are set as $\alpha_1=\alpha_2=0, \alpha_3=1$.
Figure~\ref{fig-CS-wn18-transd}(a) shows how performance changes by varying the cache size $N_1$ among $\{10, 30, 50, 70, 90\}$ with fixed $N_2=50$. 
When the cache size is small, average quality of triplets stored in the cache should be larger than those in a cache with larger size. 
As a result, 
false negative triplets will be more likely to be sampled, which will influence the embedding quality.
With the others values of $N_1$, \textit{NSCaching} performs quite stable. 
The convergence speed is similar, as well as the values in converged state.
Thus, when we need to set appropriate cache size, the value of $N_1$ can be searched from smaller values to larger ones until the performance is stable.

Different performance of the random candidate subset size $N_2$ is shown in Figure~\ref{fig-CS-wn18-transd}(b).
The entities in cache will be updated more frequently when $N_2$ gets larger, 
which lead to better exploration. But the trade-off is that larger value of $N_2$ is more expensive. 
As shown by the colored lines in Figure~\ref{fig-CS-wn18-transd}(b), 
NSCaching performs consistently when $N_2$ is larger than 10.
However, if the random subset is small, 
the content in cache will be harder to be updated, 
thus leading to poor performance as the yellow dashed line ($N_2=10$).

By combining together the influence of cache size $N_1$
and the random subset size $N_2$  
in Figure~\ref{fig-CS-wn18-transd},
we find that 
(i) NSCaching is not sensitive to the two sizes; 
(ii) both sizes can not be too small; 
(iii) $N_1=N_2$ is a good balance.

\subsubsection{Automatically Balancing E\&E}

In previous part, we have shown the importance of balancing between E\&E.
Here,
we use AutoML techniques \cite{yao2018taking} to automatically balance E\&E 
and further improve the performance on five benchmarks.

For each dataset, we use the same value of learning rate, batch size, embedding dimension and regularizer penalty as the Bernoulli baseline 
to make a fair comparison. 
The other hyper-parameters related to the negative sampling algorithm are searched within the ranges given
in Table~\ref{tab:search} by SMAC \cite{hutter2011sequential},
a well-known AutoML algorithm for hyper-parameter optimization. 
The initial hyper-parameter setting is $\alpha_1=\alpha_2=\alpha_3=0$ and $N_1=N_2=50$, 
namely a setting similar to Bernoulli sampling. 
Besides,
we take 
random search \cite{bergstra2012random} as a baseline
rather than the 
grid search,
as random search is generally more effective \cite{bergstra2012random}.

\begin{figure}[ht]
	\centering
	\subfigure[WN18RR.]{\includegraphics[width = 0.49\columnwidth]{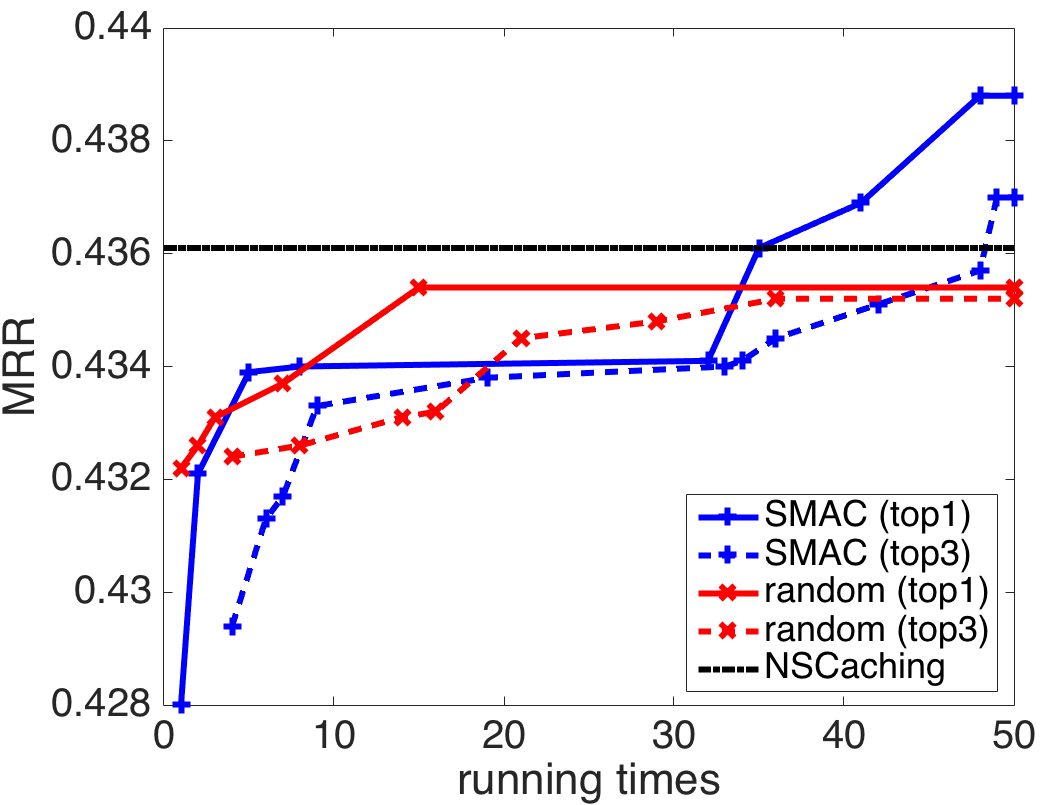}} 
	\subfigure[FB15K237.]{\includegraphics[width = 0.49\columnwidth]{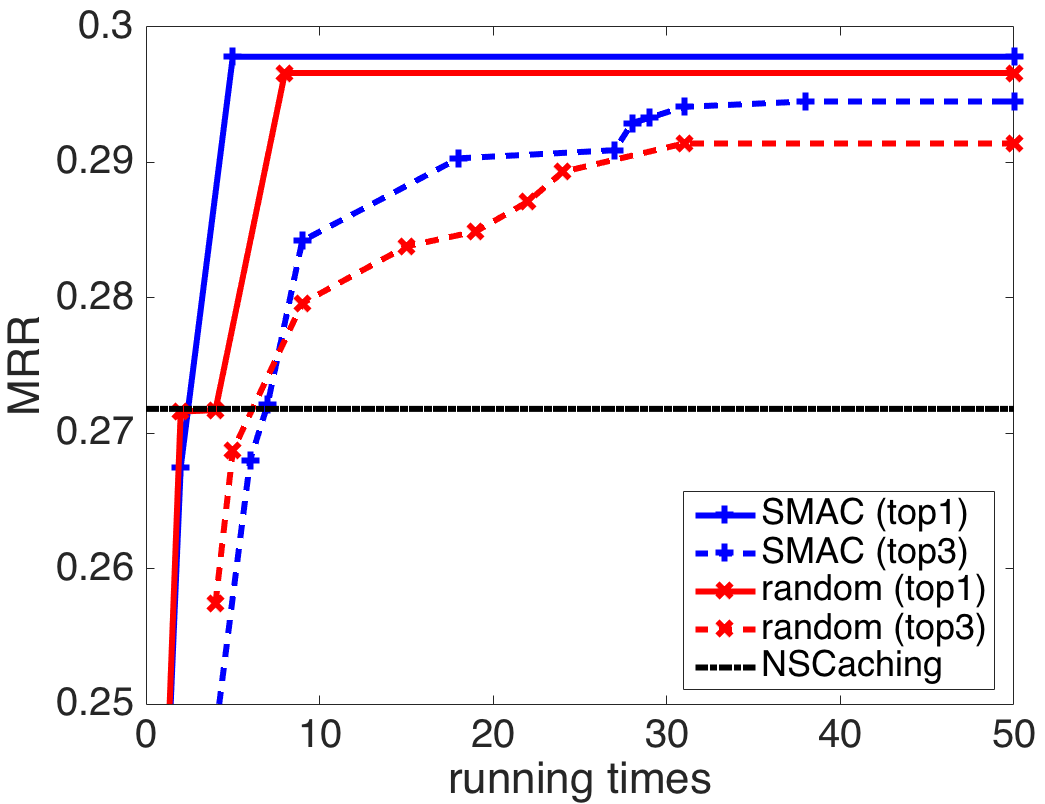}}
	\caption{Performance comparison of
		SMAC and random search (top1 and top3 average). 
		$x$-axis is number of running times for different hyper-parameter. 
		$y$-axis is the mean MRR on validation set. The MRR peformance of NSCaching is given as the
		black-dashed line for a reference (best viewed in color).}
	\label{fig:auto}
\end{figure}

Figure~\ref{fig:auto} shows the 
MRR from the best (denoted by ``top1'') and top three (denoted as ``top3'') models obtained during the search procedure.
Once the searching starts, the top performance will soon be boosted for both SMAC and random
by exploring hyper-parameters with better balance of E\&E.
Both of the two search algorithms find hyper-parameters that outperform the original \textit{NSCaching+scratch}.
However, with the help of Bayesian optimization,
SMAC is more efficient and effective than random.
This verifies the importance of introducing AutoML into the framework of \textit{NSCaching}.
After searching and running for 50 hyper-parameter settings,
we show the performance of the best hyper-parameter on testing data in
 Table~\ref{tb-perf}.
The hyper-parameter settings with best performance are given in Table~\ref{tab:search}.

\subsection{Ablation study}
\label{ssec:ablation}
	
\subsubsection{Lazy update}
\label{ssec:lazy}
\begin{figure}[ht]
	\centering
	{\includegraphics[width=0.495\columnwidth]{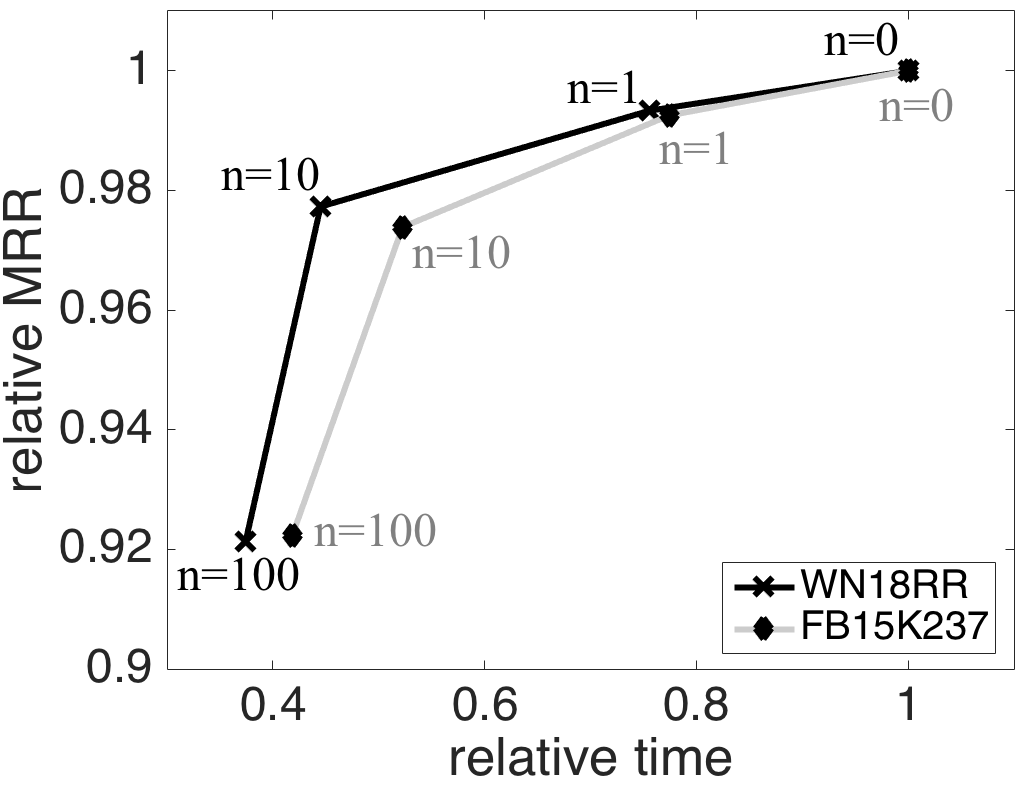}}
	\caption{Influence of different lazy update value $n$ of \textit{NSCaching}
		(SimplE is used as the testbed).}
	\label{fig:lazy}
\end{figure}

In Section~\ref{sec:proposed},
we have introduced the lazy update parameter $n$ to reduce the computation cost of \textit{NSCaching}.
In this part,
we analyze the influence of $n$ on the learning curve in Figure~\ref{fig:lazy}.
We run each model for 1,000 epochs,
and report the best performance and running time.
The relative time and MRR are divided by the corresponding values of $n=0$ respectively.
When $n$ increases,
the computation cost is reduced since less update operations are conducted.
However, the performance is gradually decreasing 
since the cache will be less frequently updated,
reducing the exploration.
Fortunately,
the decrease of performance is not obvious (less than 3\%) when $n\leq 10$.
So the value of $n$ can be regarded as a trade-off for time and performance,
adapting to different application requirements.

\subsubsection{Comparing with Self-adv}

In this part, we compare \textit{NSCaching} with the concurrent work \textit{Self-Adv}
by using the RotatE scoring function \cite{sun2018rotate}:
$f(h, r, t) =  - \|\mathbf h\circ \mathbf r - \mathbf t\|_1$,
where $\mathbf h, \mathbf r, \mathbf t$ are complex embeddings and 
$\circ$ is the Hadmard product in the complex space.

\begin{figure}[ht]
	\centering
	\subfigure[WN18RR.]
	{\includegraphics[width = 0.49\columnwidth]{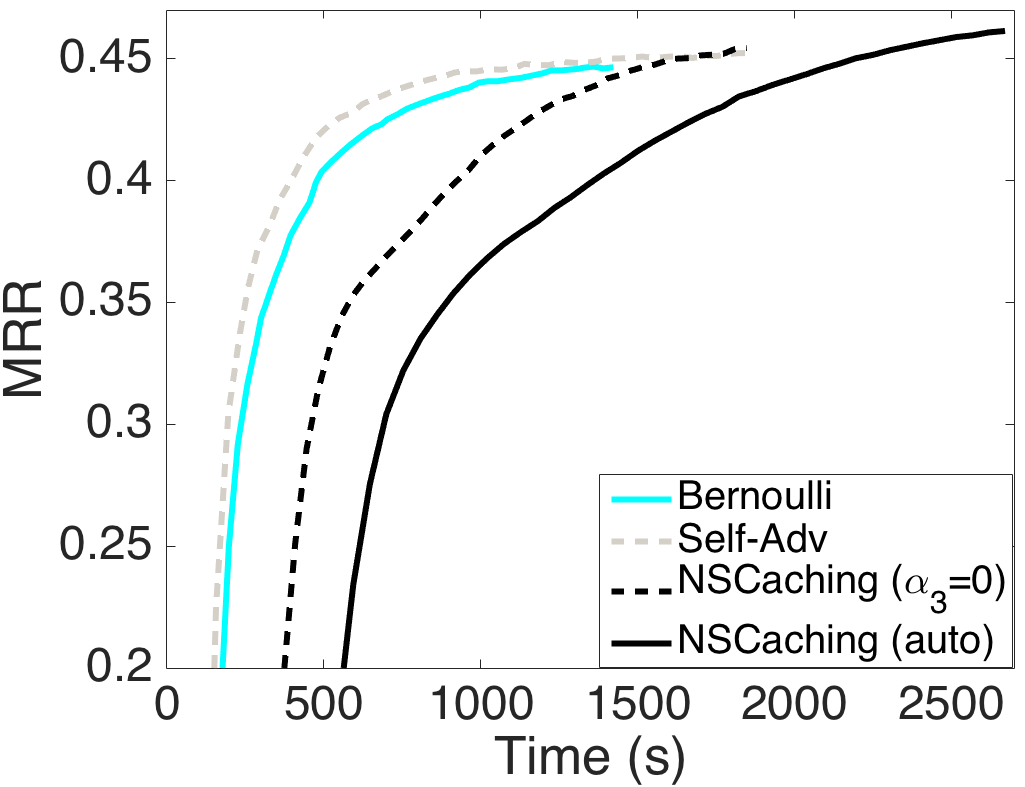}}
	\subfigure[FB15K237.]
	{\includegraphics[width = 0.49\columnwidth]{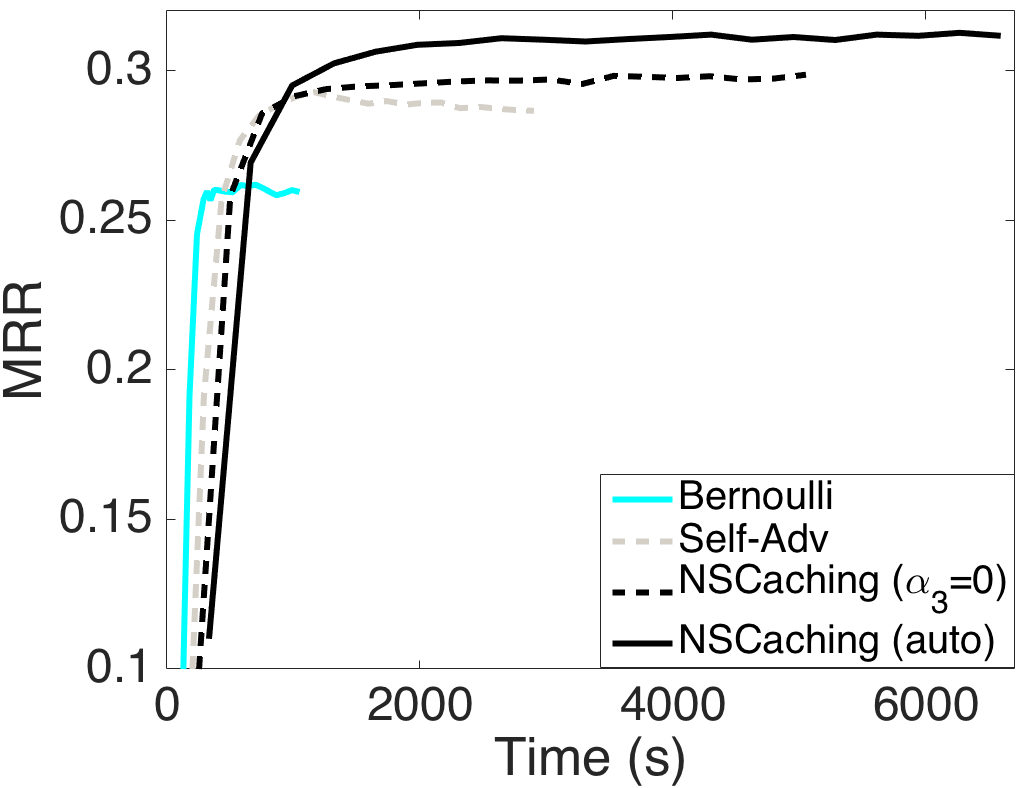}}
	\caption{Learning curve of different negative sampling method on RotatE.}
	\label{fig:rotate}
\end{figure}

As discussed in Section~\ref{sssec:methods},
\textit{Self-Adv} relies on sampling a small subset $\mathcal N$ from $\mathcal E_i$
and then sampling from $\mathcal N$.
A similar setting in \textit{NSCaching} is when $\alpha_3=0$,
where the cache is updated without depending on the scores 
so that the cache can have the same distribution with $\mathcal R_m$.
As shown in Figure~\ref{fig:rotate},
both \textit{Self-Adv} and \textit{NSCaching} outperform \textit{Bernoulli} sampling.
This again demonstrates the universality of using large-gradient negative samples to improve the performance.
Then,
\textit{Self-Adv} and \textit{NSCaching ($\alpha_3=0$)} have the similar best performance,
but \textit{NSCaching ($\alpha_3=0$)} is a bit slower due to the updating procedure.
Besides,
\textit{NSCaching (auto)} achieves the best performance by using the cache $\mathcal C_i$,
which has more large-gradient samples
than $\mathcal N$ in \textit{Self-Adv}.

\subsubsection{Influence of false negatives}
\label{ssec:falseexp}

In Section~\ref{ssec:falseneg},
we show that variance can be used as a standard to detect the false negatives.
In this part,
we empirically analyze the potential influence of introducing variance into the sampling method
with SimplE model and WN18RR data set.
We use the valid and test set here as the set of false negative samples.

\begin{figure}[ht]
	\centering
	\subfigure[Ratio of FN in the cache.]
	{\includegraphics[width=0.48\columnwidth]{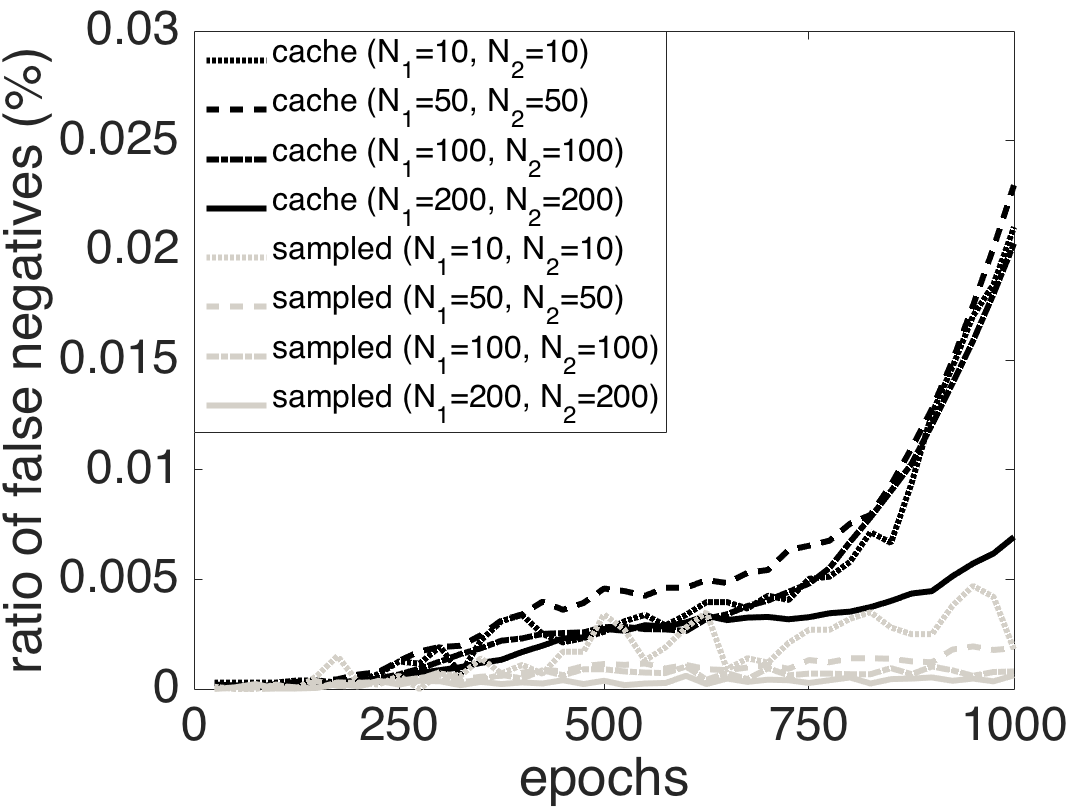}\label{fig:ratio_fn}}
	\subfigure[Average variance.]{\includegraphics[width=0.48\columnwidth]{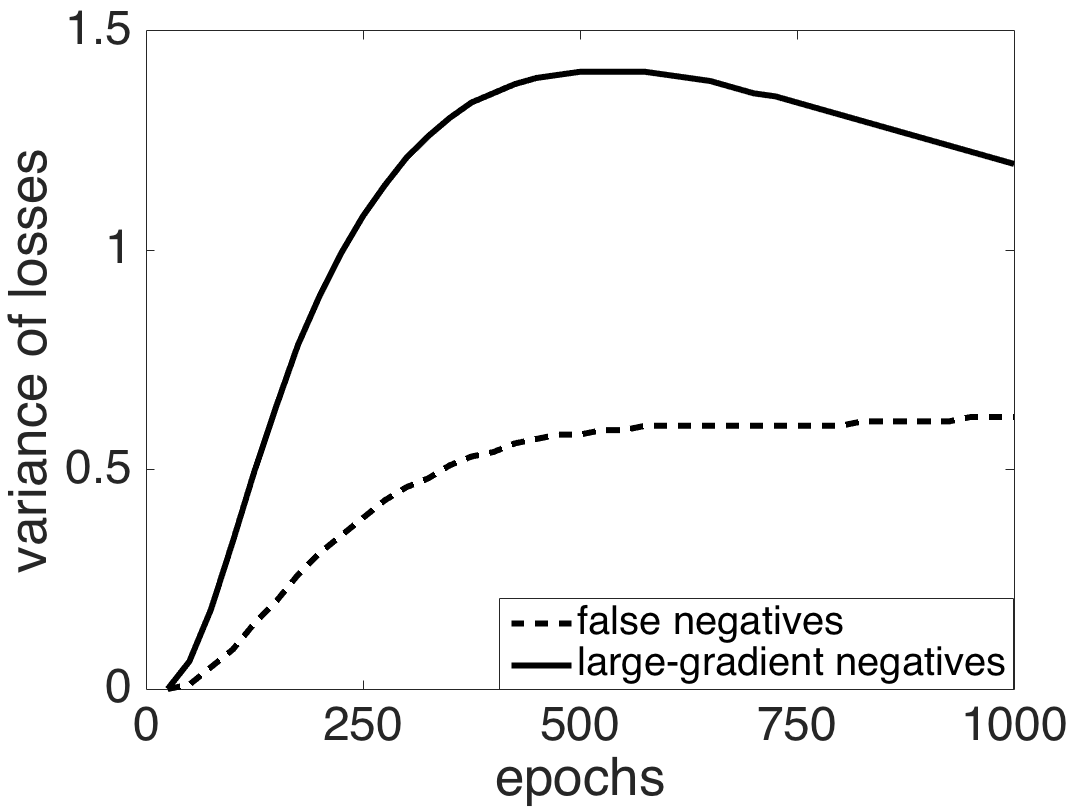}
		\label{fig:var_fn}}
	\caption{Understanding false negative (FN) triplets with SimplE on WN18RR.}
	\label{fig:falseneg}
\end{figure}

First, we analyze the possibility of sampling the false negative triplets from the cache.
In Figure~\ref{fig:ratio_fn},
we show the ratio of false negative triplets in the cache (denoted as \textit{cache})
and the percentage of false negatives in the sampled negative triplets (denoted as \textit{sampled}) during training.
The cache does contain a few false negative triplets,
but less than 0.03\% contents are false negative.
The ratio of false negatives among the sampled triplets is even smaller.
Besides,
we set $N_1=N_2$ and use different values to show how the cache size influences the ratio of false negative triplets. 
When the cache size $N_1$ increases,
the possibility of sampling the false negative triplets decreases.
Since in each iteration,
only one negative triplet is generated and the cache keeps updating,
we can avoid frequently picking up the false negative samples.

Second, we show that variance can be used as a standard to detect false negatives.
In Figure~\ref{fig:var_fn},
we plot the average variance of 5,000 false negative triplets and 5,000 large-gradient negative triplets.
The false negative triplets are sampled from the valid and test set.
For the large-gradient negative triplets,
we run \textit{NSCaching} for 1,000 epochs first and then sample 5,000 triplets,
which are not false negative,
from the final cache.
The scores of triplets are estimated per epoch,
and the variances are computed based on the recorded scores.
As shown in Figure~\ref{fig:var_fn},
the variance of large-gradient negative samples is larger than that of false negative ones.
This indicates that
the false negative triplets may be detected by tracing the variance during training.

\begin{figure}[ht]
	\centering
	\includegraphics[width=0.495\columnwidth]{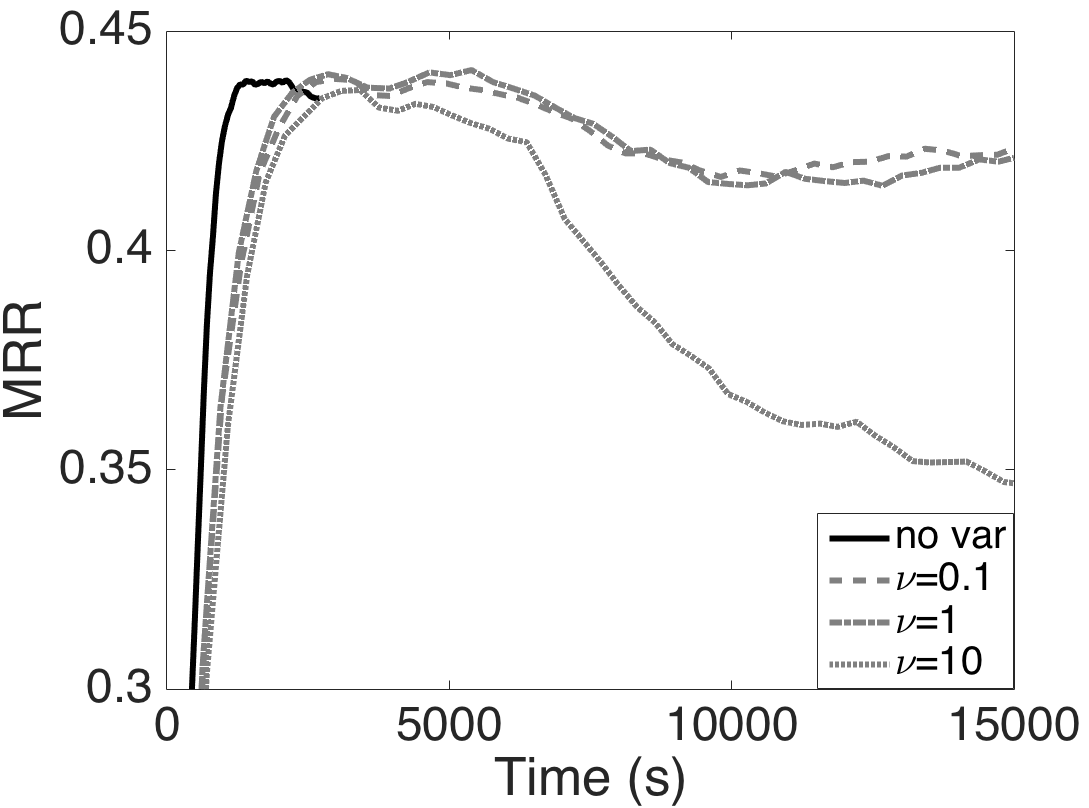}
	\caption{Adding variance with different $\gamma$ with SimplE on WN18RR.}
	\label{fig:falsecurve}
\end{figure}

Finally, we show the influence of adding variance into the sampling method.
We record the \textit{score} measured by SimplE and \textit{std} of the scores for the negative samples.
To save the memory cost, 
we use the Welford's online algorithm \cite{welford1962note} to estimate the ${std}$ of negative triplets.
Then,
we use $score + \nu std$,
where $\nu>0$ is a weighting parameter, 
as the metric here to sample the large-gradient negative triplets.
As shown in Figure~\ref{fig:falsecurve},
when $\nu$ is given an appropriate value
like $\nu\leq 1$, 
the best performance is almost the same,
but the computation cost increases a lot.
The training will be instable after adding the ${std}$ term,
especially when the value of $\nu$ is large.
Therefore,
it is not necessary to consider the variance to reduce the problem of false negative triplets in this work.
Instead, we control $\alpha$'s to avoid frequently sampling the false negative triplets.

\subsection{Theoretical Explanation}
\label{sec:visself}

Here,
we study the theoretical perspective of the proposed approach,
which gives more insights and helps us understand NSCaching better. 

\subsubsection{Illustration of Vanishing Gradients}

\begin{figure}[ht]
	\centering
	\subfigure[TransD.]
	{\includegraphics[width = 0.49\columnwidth]{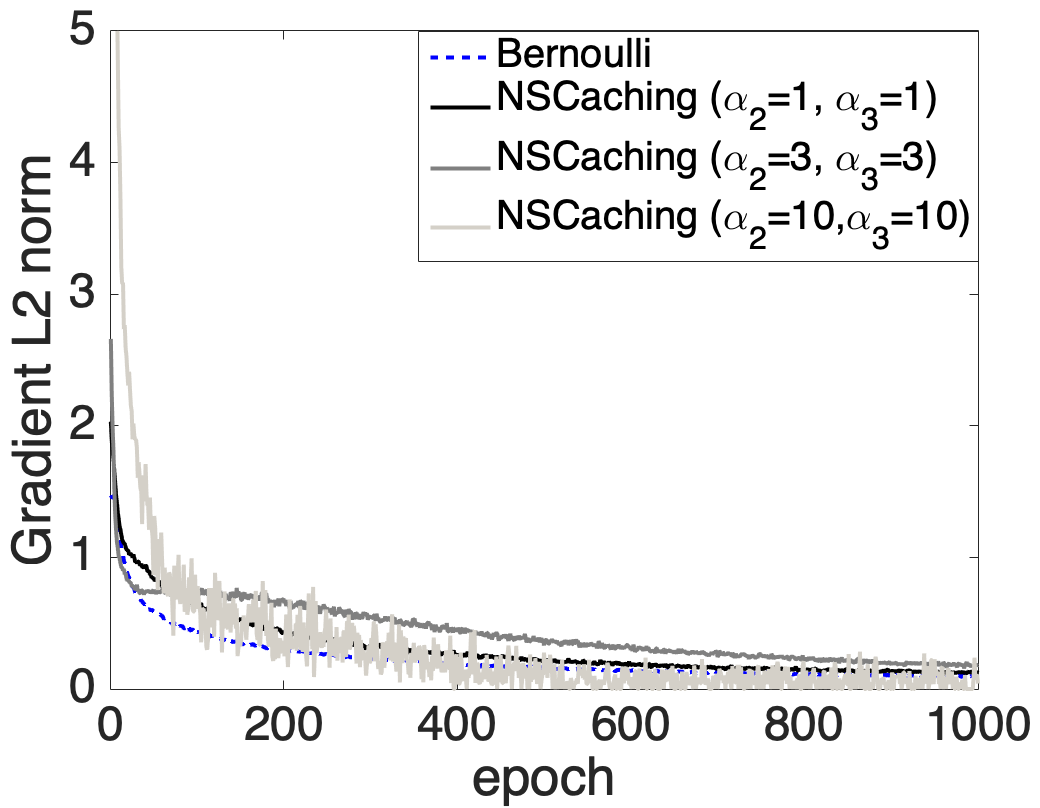}}
	\subfigure[SimplE.]
	{\includegraphics[width = 0.49\columnwidth]{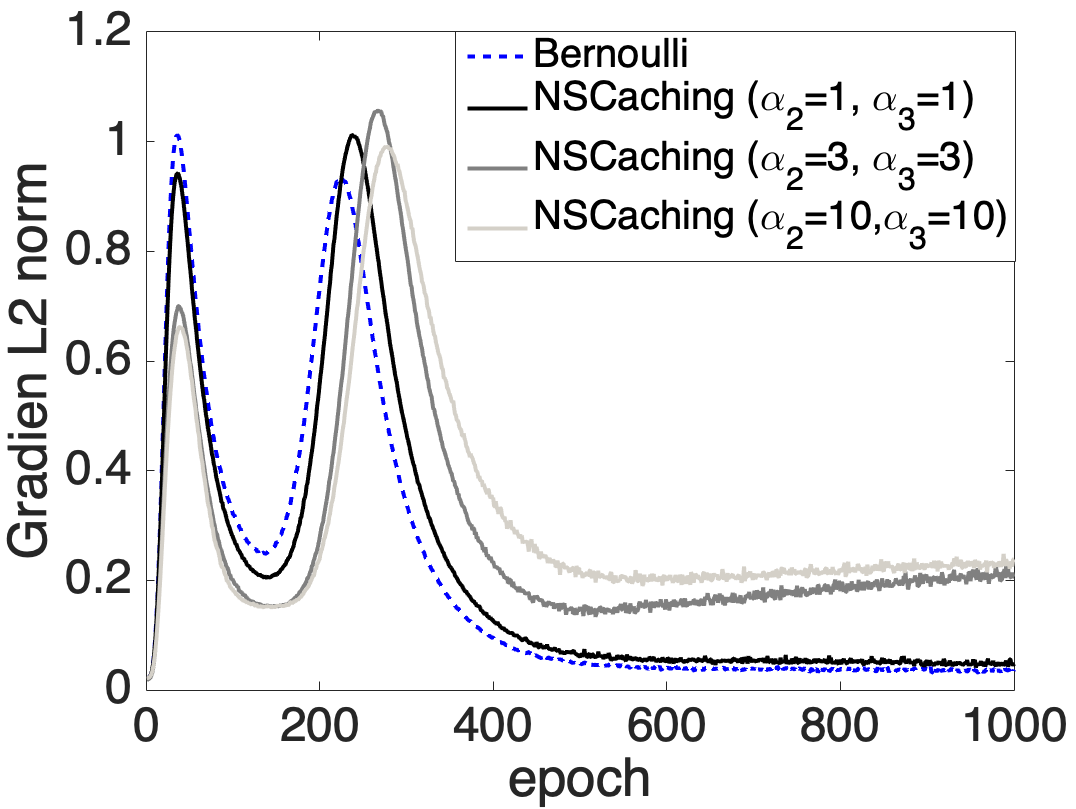}}

	\caption{Average $\ell_2$-norm of gradients within a mini-batch v.s. number of epochs for Bernoulli and NSCaching on WN18RR.}
	\label{fig-GN-wn18rr}
\end{figure}

To further clarity the vanishing gradient problem, we plot the average $\ell_2$-norm of gradients v.s. number of epochs in Figure~\ref{fig-GN-wn18rr}. 
Adam \cite{kingma2014adam}, a stochastic gradient descent  algorithm, is used as the optimizer.
First, we can see that while the norms of gradients for both \textit{NSCaching} and \textit{Bernoulli} become smaller, 
they will not decrease to zero since the sampling process of the mini-batch will introduce noise into gradients.
However, the norm from \textit{NSCaching} is larger than that from \textit{Bernoulli} 
due to the usage of cache-based negative sampling scheme.
Thus, we can see \textit{NSCaching} can successfully avoid the problem of vanishing gradient.
We also show the changes with different $\alpha$'s.
When the value of $\alpha$'s increases, the gradient norm will become larger, 
especially after the warm-up procedure, i.e., after 400 epochs.
For the TransD model, when $\alpha_2\!=\!\alpha_3\!=\!10$, the training will become unstable.
This also verifies that we should control the value of $\alpha$ through the AutoML technique.

\begin{figure}[ht]
	\centering
	\subfigure[WN18RR.]
	{\includegraphics[width = 0.49\columnwidth]{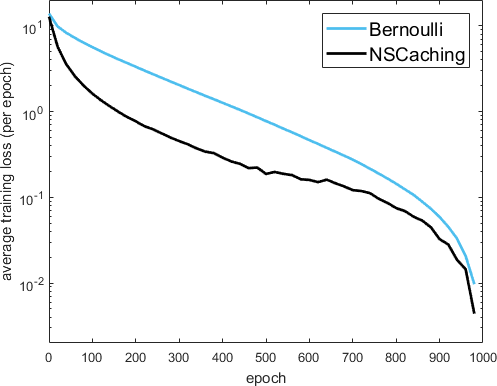}}
	\subfigure[FB15K237.]
	{\includegraphics[width = 0.49\columnwidth]{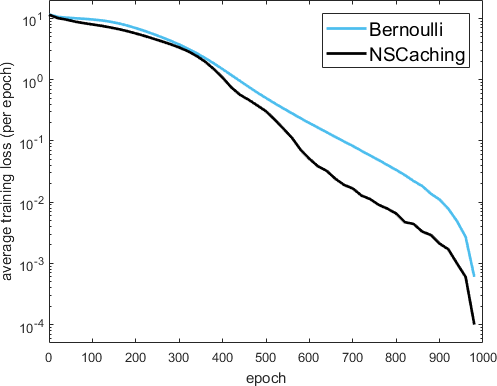}}
	
	\caption{Average training loss in one epoch v.s. number of epochs for Bernoulli and NSCaching.}
	\label{fig:conv}
\end{figure}

\subsubsection{Convex Case: Faster convergence}
\label{ssec:conv}

To demonstrate the faster convergence illustrated in Theorem~\ref{thm:conv},
we use TransE as the scoring function and use classification loss in \eqref{eq:sematic} for KG embedding.
As mentioned in Section~\ref{ssec:convex},
this does not fall into any existing KG embedding models,
but it satisfies assumptions in Theorem~\ref{thm:conv}.
Thus,
it is a good synthetic model to be studied.
We use the same hyper-parameters identified for \textit{NSCaching} in Section~\ref{ssec:compstate}, 
and compare it with \textit{Bernoulli} scheme.
FB15K237 and WN18RR are considered here.
The comparison of average training loss per epoch v.s. epoch 
between \textit{NSCaching} and \textit{Bernoulli}
is shown in Figure~\ref{fig:conv}.
As we can see,
\textit{NSCaching} indeed leads to faster convergence than \textit{Bernoulli},
which is explained by Theorem~\ref{thm:conv}.

\subsubsection{Nonconvex Case: Self-paced learning}

Finally,
we visualize the changes of content in the cache,
which verifies the effects of self-paced learning introduced in Section~\ref{sec:self-pace}. 
Following \cite{wang2018incorporating},
we also use 
FB13 here,
which has 75,043 entities, 13 relations and 316,232 training triplets,
 since triplets in this dataset are more interpretable than the five evaluated datasets.
We pick up 
$(manorama$, $profession$, $actor)$ as the positive triplet, 
and the changes in its tail-cache $\mathcal T_{r,t}$ are shown in Table~\ref{tab:cache:example}.
As can be seen,
negative tails are firstly meaningless,
e.g., \textit{ostrava} and \textit{ben\_lilly},
then they gradually changes to human jobs,
e.g., \textit{artist} and \textit{sex\_worker}.

\begin{table}[ht]
	\centering
	\caption{Examples of negative entities in cache on FB13. Each line displays 5 random sampled negative entities from tail-cache of a positive fact \textit{(manorama, profession, actor)} in different epochs.}
	\label{tab:cache:example}
	\renewcommand{\arraystretch}{1.1}
	\begin{tabular}{c|c}
		\hline
		epoch &                            entities in cache                            \\ \hline
		0   & \textit{allen\_clarke,  jose\_gola, ostrava, ben\_lilly, hans\_zinsser} \\ \hline
		20   &  \textit{accountant, frank\_pais, laura\_marx, como, domitia\_lepida}   \\ \hline
		100  &        \textit{artist, , aviator, hans\_zinsse, john\_h\_cough}         \\ \hline
		200  &     \textit{physician, artist, raich\_carter, coach, mark\_shivas}      \\ \hline
		500  &    \textit{artist, physician, cavan, sex\_worker, attorney\_at\_law}    \\ \hline
	\end{tabular}
\end{table}

\subsection{Graph Embedding}
\label{sec:exp:graph}

\begin{table*}[ht]
	\centering
	\caption{Node classification results.}
	\renewcommand{\arraystretch}{1.1}
	\setlength\tabcolsep{5pt}
	\begin{tabular}{c|c|c|c|c|c|c|c}
		\hline
		\multirow{2}{*}{datasets} & \multirow{2}{*}{ methods} &              \multicolumn{2}{c|}{30\%}              &              \multicolumn{2}{c|}{50\%}              &              \multicolumn{2}{c}{70\%}               \\ \cline{3-8}
		&                           &         Micro-F1         &         Macro-F1         &         Micro-F1         &         Macro-F1         &         Micro-F1         &         Macro-F1         \\ \hline\hline
		\multirow{4}{*}{Cora}   & LINE \cite{tang2015line}  &     0.8093$\pm$0.0027      &     0.8037$\pm$0.0021      &     0.8120$\pm$0.0025      &     0.8024$\pm$0.0029      &     0.8096$\pm$0.0043      &     0.8037$\pm$0.0057      \\ \cline{2-8}
		&         Node2vec          &     0.8130$\pm$0.0031      &     0.8061$\pm$0.0031      &     0.8142$\pm$0.0024      &      0.8064$\pm$0035      &     0.8376$\pm$0.0038      &     0.8323$\pm$0.0025      \\ \cline{2-8}
		& SeedNE \cite{gao2018self} &     0.8142$\pm$0.0033      &     0.8089$\pm$0.0023      &     0.8195$\pm$0.0024      &     0.8081$\pm$0.0030      &     0.8364$\pm$0.0047      &     0.8315$\pm$0.0024      \\ \cline{2-8}
		&         \textbf{NSCaching}         & \textbf{0.8185}$\pm$0.0031 & \textbf{0.8108$\pm$0.0028} & \textbf{0.8273$\pm$0.0039} & \textbf{0.8184$\pm$0.0043} & \textbf{0.8413$\pm$0.0019} & \textbf{0.8334$\pm$0.0027} \\ \hline
		\multirow{4}{*}{Citeseer} & LINE \cite{tang2015line}  &     0.5465$\pm$0.0027      &     0.4984$\pm$0.0044      &     0.5488$\pm$0.0035      &     0.5074$\pm$0.0032      &     0.5754$\pm$0.0027      &     0.5175$\pm$0.0039      \\ \cline{2-8}
		&         Node2vec          &     0.5755$\pm$0.0019      &     0.5253$\pm$0.0017      &     0.5793$\pm$0.0013      &     0.5357$\pm$0.0016      &     0.5936$\pm$0.0044      &     0.5410$\pm$0.0013      \\ \cline{2-8}
		& SeedNE \cite{gao2018self} &     0.5762$\pm$0.0028      &     0.5279$\pm$0.0030      &     0.5918$\pm$0.0025      &    {0.5369$\pm$0.0031}     &     0.6059$\pm$0.0015      &     0.5552$\pm$0.0038      \\ \cline{2-8}
		&         \textbf{NSCaching}         & \textbf{0.5864$\pm$0.0018} & \textbf{0.5297$\pm$0.0047} & \textbf{0.5982$\pm$0.0020} & \textbf{0.5531$\pm$0.0029} & \textbf{0.6120$\pm$0.0015} & \textbf{0.5712$\pm$0.0018} \\ \hline
	\end{tabular}
	\label{tab:graphemb}
\end{table*}

In this part,
we perform experiments with the skip-gram model
on the task of graph embedding.

\subsubsection{Experiment Setup}

Two famous graph data sets are used here:
Cora and Citeseer, both of which are academic citation networks introduced in \cite{mccallum2000automating}.
Cora contains 2,708 papers with 5,429 connections in machine learning area.
These papers belong to 7 different classes.
Citeseer is formed by 3,312 papers in 6 areas. The total number of connections is 4,660.
In this part
we compare NSCaching (in Algorithm~\ref{alg:graphemb}) with the following methods.
\begin{itemize}[leftmargin=12pt]
\item \textit{LINE} \cite{tang2015line}: 
In this method, the embeddings are trained to preserve the first order,
i.e., the direct connections,
and second order, 
i.e., the neighborhood similarities,
in graphs.

\item \textit{Node2vec} \cite{Grover2016}:
Different from LINE,
\textit{Node2vec} uses biased random walk to preserve the topology information on graphs.
The generated walks are regarded as sentences in the language model.
In this way, skip-gram model is used to learn the embeddings.
The distribution used to sample negative nodes is proportional to $\nicefrac{3}{4}$ of the nodes' frequency.

\item \textit{SeedNE} \cite{gao2018self}:
To improve the negative sampling in skip-gram model,
SeedNE selects the negative nodes 
whose similarities with the positive node are higher than an increasing threshold.
Either the self-embedding or a learned generator can be used to indicate the similarities.
To guarantee stability, we use self-embedding in this part.
This sampling method also has some connection with the self-paced learning,
but the problem of E\&E is not well addressed.

\end{itemize}

In order to make a fair comparison with the baselines,
we set the embedding dimension to be 100 for all the datasets
as in \cite{gao2018self}.
Similar as \cite{Grover2016}, 10 random walks are generated for each node with $p=0.25, q=0.25$,
both of which are hyper-parameters controlling the biased random walk. 
The window size $|\mathcal W_{u}|$ is set to be 10 for each node in the walks.
The walks are fixed when comparing different models. 
Finally, 
we use Adam \cite{kingma2014adam} as the optimizer.
The learning rate is set as the default value $10^{-3}$ and we use $\lambda=10^{-7}$ as the weight decay value.

To measure the quality of the learned embeddings of different methods,
we use node classification task as the testbed.
After the embeddings are learned,
a logistic regression model is learned as the classifier.
Specifically,
we use the cache-based skip-gram model to train the node embeddings.
Then the embeddings and their corresponding labels are fed into the classifier.
Following \cite{tang2015line,Grover2016,gao2018self},
we use
F1-score, 
which considers both the precision and recall,
to measure the test accuracy.
It is a widely used metric in binary classification 
and is computed by 
$F_1 = 2\cdot \nicefrac{\text{precision}\cdot \text{recall}}{\text{precision} + \text{recall}}$.
Considering that the problem here is a multi-class task,
we use two variants
\begin{itemize}
	\item Micro-F1: the precision and recall are computed by ignoring the type, 
	and then computing the F1-score.
	\item Macro-F1: the precision and recall are computed separately on each class, 
	and return the average F1-score.
\end{itemize}
The larger values of the F1-scores indicate better performance.

\subsubsection{Empirical Results}

We randomly select $\{30\%$, $50\%$, $70\%\}$ labeled nodes to train the classifier five times
and evaluate the performance on the remaining nodes respectively.
We report the average and the standard deviation 
on the node classification task in Table~\ref{tab:graphemb}.
Comparing with the frequency-based negative sampling method,
i.e., \textit{Node2vec} \cite{Grover2016},
the \textit{NSCaching} based skip-gram method and \textit{SeedNE} 
gain significant improvement
since they are able to capture the dynamic  distribution of negative samples.
By comparing \textit{Node2vec} (skip-gram) with \textit{LINE}, we can see that
the random walk based model is better than \textit{LINE},
which only uses direct connection and neighbors to measure the similarity.
Besides,
\textit{NSCaching}
outperforms another self-paced negative sampling method \textit{SeedNE}
with better control of E\&E.

\section{Conclusion}
\label{sec:conclude}

In this paper,
we propose NSCaching, a novel negative sampling method for knowledge graph embedding learning. 
The negative samples are sampled from a cache that can dynamically hold large-gradient negative samples. 
We theoretically understand the convergence and effectiveness from both convex and non-convex case.
In order to balance exploration and exploitation during the sampling procedure,
we use AutoML to automatically balance E\&E for NSCaching
in regard of sampling and updating the cache.
In addition,
we extend NSCaching to skip-gram model, which is widely used in word embedding and graph embedding.
Experimentally, we test NSCaching on benchmark datasets and various scoring functions. 
Empirical results show that the method can generalize well under various settings and achieves state-of-the-arts performance.
In future work,
we will use AutoML technique to search for scoring functions \cite{zhang2020autosf}
and explore recurrent neural architectures to learn from relational paths in knowledge graphs \cite{zhang2019neural}.
%we will further see how to improve efficiency on extremely large graph or knowledge graph
%with advanced data structure like hash tables.

\noindent
\textbf{Acknowledgements}

\appendix
\section{Appendix: Proof of Theorem~\ref{thm:conv}}
\label{prf:thm:conv}

\begin{proof}
	Following \cite{zhao2015stochastic},
	we consider a more general optimization formulation as
	$	\min_{\mathbf{w}}
	\bar{F}(\mathbf{w}) \equiv
	\frac{1}{n} \sum\nolimits_{i = 1}^n \phi_{i}(\mathbf{w}) + \lambda r(\mathbf{w})$,
	which covers \eqref{eq:genobj} as a special case with $\lambda = 0$.
	Then, 
	the stochastic training gives $\mathbf{w}^{t + 1}$ as
	\begin{align}	
	\mathbf{w}^{t + 1}
	\!=\! 
	\arg\min_{\mathbf{w}}
	\left[
	(np_{i_t}^t)^{-1}\mathbf{w}^{\top} \nabla \phi_{i_t}(\mathbf{w}^t)
	\! + \! \lambda r(\mathbf{w})
	\! + \! \frac{1}{\eta_t} \mathcal{B}_{\psi}(\mathbf{w}, \mathbf{w}^t)
	\right]\!,\!
	\label{eq:updates}
	\end{align}
	where $\mathcal{B}_{\psi}$ is a Bregman distance function measuring 
	the difference between $\mathbf{w}$ and $\mathbf{w}^t$.
	Based on \eqref{eq:updates},
	we state Theorem~3 in \cite{zhao2015stochastic} as
	\begin{theorem}[\cite{zhao2015stochastic}] \label{thm:gen}
		Let $\mathbf{w}^t$ be generated from \eqref{eq:updates}.
		Assume $\mathcal{B}_{\psi}$ is $\sigma$-strongly convex w.r.t. a norm $\| \cdot \|$,
		$\bar{F}$ and $r$ are convex,
		if $\eta_t = \eta$,
		the following inequality holds for any $T \ge 1$
		\begin{align*}
		& \frac{1}{T}\sum\nolimits_{t =1}^T 
		\mathbb{E}[\bar{F}(\mathbf{w}^t)] - \mathbb E[\bar{F}(\mathbf{w}^*)]  \\
		& \le \frac{1}{T}
		\left[
		\frac{1}{\eta} \mathcal{B}_{\psi}(\mathbf{w}^*, \mathbf{w}^1)
		+ \frac{\eta}{2\sigma} \sum\nolimits_{t = 1}^T 
		\mathbb{E}
		\|
		\nicefrac{\nabla \phi_{i_t}(\mathbf{w}^t)}{ n \mathbf{p}_{i_t}^t }
		\|^2
		\right] 
		\end{align*}
		where the expectation is take with distribution $\mathbf{p}^t$.
	\end{theorem}
	
	In our Algorithm~\ref{alg:prototype},
	we do not have $r$ and thus $\lambda = 0$ in \eqref{eq:genobj}.
	Besides,
	$\mathcal{B}_{\psi}(\mathbf{w}, \mathbf{w}^t) = \frac{1}{2} \| \mathbf{w} - \mathbf{w}^t \|^2$ in our case,
	thus $\sigma = 1$.
	Above all, 
	we have Theorem~\ref{thm:conv} which is derived from Theorem~\ref{thm:gen}.
\end{proof}

\bibliographystyle{plain}
\bibliography{bib}

\end{document}